\documentclass{article}
\usepackage{graphicx}
\usepackage{booktabs}
\usepackage{multirow}
\usepackage{makecell}
\usepackage{graphicx}
\usepackage{wrapfig}
\usepackage{booktabs}
\usepackage{multirow}
\usepackage{graphicx}
\usepackage{xcolor}
\usepackage{caption}
\usepackage{amsmath}
\usepackage{amsfonts}
\usepackage{subcaption}

\usepackage{xcolor}

\newcommand{\sdev}[1]{%
  {\raisebox{0.45ex}{\fontsize{4.6}{4.8}\selectfont\textcolor{gray}{\,$\pm$#1}}}%
}

\usepackage[final]{corl_2026} 

\title{Diffusion ReRoll: Revisable Denoising \\for Robotic Sequential Prediction}

\author{
  Seonsoo Kim \quad
  Seongil Hong \quad
  Jun-Gill Kang  \\
  Agency for Defense Development \\
  \texttt{\{sunsoo3165,science4729,jungillkang\}@gmail.com} \\
  \vspace{1.5em}
  \href{https://seonsoo-p1.github.io/DiffusionReRoll}
     {seonsoo-p1.github.io/DiffusionReRoll}
  \vspace{-2.0em}
}

\begin{document}
\maketitle

\vspace{-1.0em}

\begin{abstract}
        We propose Diffusion ReRoll, a diffusion-based framework for robotic sequential prediction that enables revisable denoising over horizons. Existing diffusion-based sequence predictors typically perform a single monotonic denoising process. In contrast, Diffusion ReRoll selectively re-noises regions that reach a predefined noise threshold while the remaining regions continue denoising, so the re-noised regions can be refined again using context from the rest of the horizon. This structured re-noising enables iterative cross-horizon revision, allowing earlier and later segments to revise one another, while maintaining local consistency. We evaluate Diffusion ReRoll against full-sequence diffusion and causal denoising based on Diffusion Forcing across long-horizon planning, policy learning, unified video-action modeling, and real-world navigation. On OGBench PointMaze and AntMaze, Diffusion ReRoll achieves relative gains in average success rate of 21\% over Diffusion Forcing in matched guidance-based planning and 23\% over Diffuser in goal-inpainting. In diffusion-policy-style action prediction, Diffusion ReRoll improves average success by 56.5\% relative to Diffusion Policy across different prediction horizons and history lengths on LIBERO-10 under the official reset protocol. In unified video-action prediction, Diffusion ReRoll improves policy and inverse dynamics performance, especially under out-of-distribution evaluation, and achieves the best action-video consistency. Real-world navigation further demonstrates improved success under visual distribution shifts with onboard inference. These results show that structured re-noising enables cross-horizon
        revision and improves performance across robotic sequential prediction tasks.
\end{abstract}


\begin{figure}[h!]
    \centering
    \includegraphics[width=0.91\linewidth]{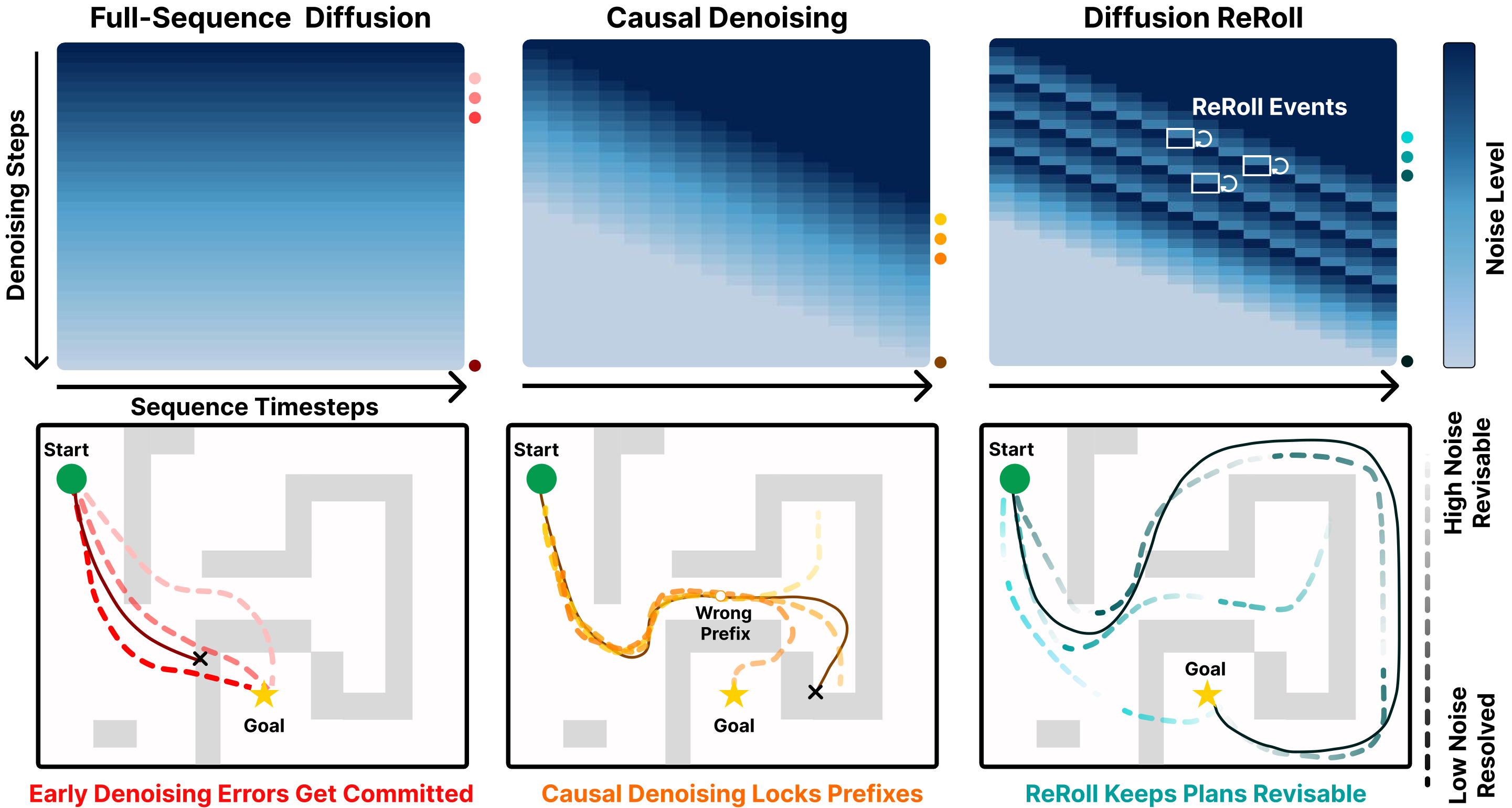}
    \caption{
Comparison of full-sequence diffusion, causal denoising, and Diffusion ReRoll.
ReRoll events enable cross-horizon revision, preventing premature commitment to erroneous predictions.
}
    \label{fig:one}
\end{figure}


\section{Introduction}
\vspace{-0.2em}

Sequential prediction is central to robotics and embodied intelligence, including video prediction, planning, policy learning, and behavior synthesis.
Diffusion models~\cite{DiffusionThermo,DDPM,conditionalGenerative} are widely used for these tasks because they capture high-dimensional, temporally correlated, multimodal outputs~\cite{diffusion_policy,videoDiffusion,diffuser,UVA,chen2025large,generativeSKillChaining}.
However, standard full-sequence diffusion denoises with a shared noise level across the entire horizon monotonically, giving limited control over temporal dependencies across the horizon.

\vspace{-0.1em}
  
Prior work has introduced temporal structure for diffusion-based sequential generation.
Some methods compose smaller generative components for hierarchical or stitched long-horizon planning, as in CompDiffuser~\cite{stitch,stitching2}; others assign timestep-dependent noise levels, as in Diffusion Forcing, enabling flexible schedules such as causal denoising~\cite{AR,diffusion_rolling,diffusion_forcing}. 
These approaches make sequence organization more explicit, but how distant horizon segments influence one another during denoising remains less directly addressed.
For example, an early segment may enter a low-noise state before terminal constraints are incorporated, making an erroneous prefix difficult to correct later.
 
  \vspace{-0.1em}

This motivates incorporating revisability into the generative process without discarding sequential structure. 
Prior revision methods revisit intermediate outputs through noise injection for editing~\cite{sgedit}, diffusion resampling for inpainting~\cite{repaint}, masked re-prediction for iterative token generation~\cite{maskGen,masked_generative}, and repair-based refinement for infeasible diffusion plans~\cite{refiningDiffusion}. 
These works show the value of revision, but leave open how to decide which parts of a sequence should be corrected.

\vspace{-0.1em}

Rather than adding a separate repair module, we ask whether the diffusion denoising process itself can provide this mechanism.
In robotic control, stochasticity and supervised iterative denoising provide beneficial inductive biases for generative policies~\cite{much_ado,bias}.
This suggests that stochasticity during training may shape the model's learned inductive bias, rather than serving only as noise to be removed.
In video diffusion, different denoising steps have been associated with exploration, correction, and consolidation~\cite{demystifing_video}, suggesting that noise level can influence the role a prediction plays during generation.
Together, these observations suggest that structured stochastic noise-level patterns, rather than independent noise sampling, can provide a useful inductive bias for revisable denoising. 
These patterns assign sequence regions to noise levels with different functional roles, while the noise level scheduling controls when and where those roles are used across the horizon.

\vspace{-0.1em}



We propose \emph{Diffusion ReRoll}, a schedule matrix-based framework for revisable robotic sequence generation. 
Building on per-token noise conditioning, Diffusion ReRoll organizes a single trajectory with virtual piecewise-linear noise-level patterns, called linear chunks, without explicitly splitting the trajectory. 
To support the desired schedule, we train with randomized linear chunks, exposing the model to structured token-wise stochasticity.
During denoising, regions that reach the predefined reset trigger level are re-noised and refined again under updated context from the rest of the sequence, preventing premature commitment across the full horizon.
This enables iterative cross-horizon revision, where earlier and later segments mutually revise one another before the final sequence is produced.
Moreover, by changing the schedule matrix, the same model can adjust how information propagates across the horizon, emphasizing forward, backward, or endpoint-conditioned influence.

\vspace{-0.1em}

We evaluate Diffusion ReRoll across long-horizon planning, policy learning,
unified video-action modeling, and real-world navigation.
On OGBench PointMaze and AntMaze~\cite{OgbenchBench}, ReRoll raises average
success from 72.8\% to 88.1\% over Diffusion Forcing~\cite{diffusion_forcing}
in guidance-based replanning and from 73.8\% to 90.8\% over
Diffuser~\cite{diffuser} in goal inpainting.
On LIBERO-10 Official setup~\cite{libero}, average success rises from 40.1\% to 62.7\%
over Diffusion Policy~\cite{diffusion_policy} across observation histories and horizons.
In unified video-action prediction~\cite{uwm}, ReRoll raises joint-policy
success under out-of-distribution (OOD) evaluation from UWM's 58.1\% to 69.7\%.
Beyond policy prediction, it also improves inverse dynamics and reduces
zero-action diagnostic error.
Finally, in real-world navigation with onboard inference,
ReRoll achieves the highest mean success under OOD conditions.

\vspace{-0.1em}

Our contributions are threefold.
First, we introduce a revisable denoising process that enables cross-horizon revision through structured noise-level scheduling.
Second, we demonstrate the benefits of repeated cross-horizon revision across diverse robotic prediction tasks.
Third, we show that varying the schedule matrix enables programmable control of information flow across sequences.

\section{Preliminaries}

\subsection{Generative Sequential Prediction}

We consider conditional sequence generation, where a model predicts
$x_{1:T}=(x_1,\ldots,x_T)$ given context $c$. 
Depending on the task, $x_t$ may denote an action, state, or video latent, while
$c$ may include observations, goals, or rewards. Throughout the paper, subscripts denote sequence timesteps and superscripts denote diffusion noise levels. 
Thus, $x_t$ is the token at sequence timestep $t$, and $x_t^k$ is the same token corrupted to noise level $k$. 
Here, $k\in\{0,\ldots,K\}$ is a discrete diffusion index. $k=0$ denotes a clean sample and $k=K$ denotes the Gaussian noise state. 
We use ``noise level'' to refer to this diffusion index, not the exact noise variance or signal-to-noise ratio (SNR) value.

A standard diffusion model corrupts a clean sequence $x_{1:T}^0$ as
\begin{equation}
    x_{1:T}^{k}
    =
    \sqrt{\bar{\alpha}_k}x_{1:T}^{0}
    +
    \sqrt{1-\bar{\alpha}_k}\epsilon,
    \qquad
    \epsilon \sim \mathcal{N}(0,I),
    \quad
    \bar{\alpha}_k=\prod_{i=1}^{k}(1-\beta_i),
\end{equation}
where $\{\beta_i\}_{i=1}^{K}$ is the noise variance schedule, and trains a denoising network $\epsilon_\theta(x_{1:T}^{k}, k, c)$ to predict the injected noise.
At inference, the model starts from noise and iteratively denoises the sequence.


\subsection{Per-Token Noise Levels and Diffusion Forcing}
\label{subsec:PerToken}
Standard full-sequence diffusion assigns one noise level to the entire sequence. 
Diffusion Forcing (DF) generalizes this by allowing each token to have its own noise level~\cite{diffusion_forcing}. 
We denote the per-token noise levels, noisy sequence, and schedule matrix as
\begin{equation}
    \mathbf{k}=(k_1,\ldots,k_T), \qquad
    x_{1:T}^{\mathbf{k}}=(x_1^{k_1},\ldots,x_T^{k_T}), \qquad
    \mathbf{S}\in\{0,\ldots,K\}^{M\times T}.
\end{equation}
Here, $k_t$ is the diffusion noise level assigned to token $x_t$, and
$\mathbf{S}_{m,t}$ specifies the noise level of token $t$ at denoising step row $m$.

\begin{wrapfigure}{r}{0.42\textwidth}
  \vspace{-1.2em}
  \centering
  \includegraphics[width=0.41\textwidth]{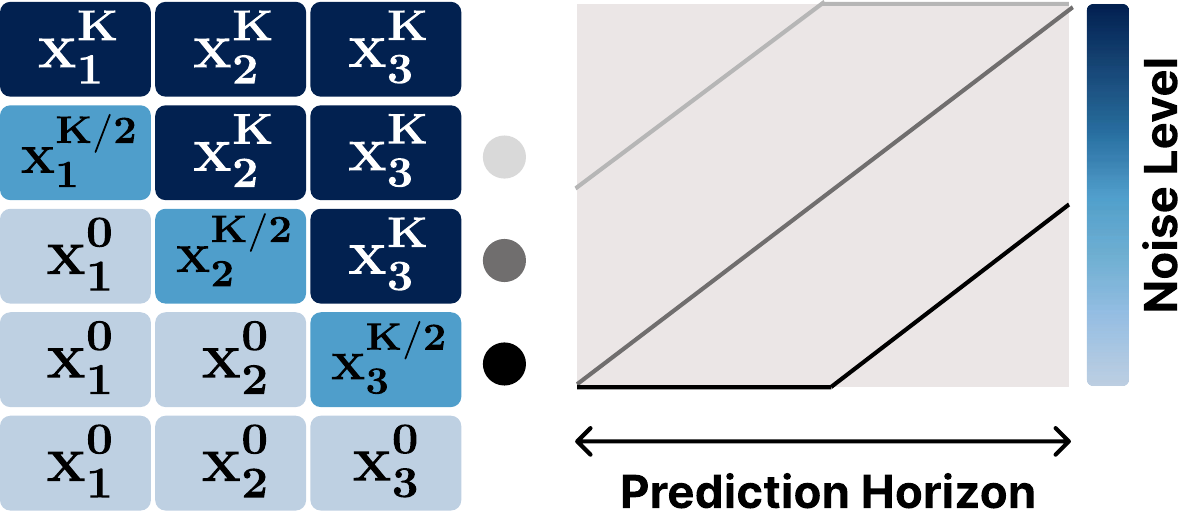}
  \caption{Causal Denoising by Per-Token Noise Levels }
  \label{fig:Causal}
  \vspace{-1.0em}
\end{wrapfigure}

This formulation can be viewed as noise-as-masking. $x_t^0$ is fully visible, $x_t^K$ is effectively masked, and intermediate levels are partially visible. 
Per-token noise levels therefore place different sequence tokens at different denoising states. 
Diffusion Forcing trains with randomly assigned token-wise noise levels, which enables the same model to be sampled with structured schedule matrices, including causal denoising schedules that resolve earlier tokens before later ones. Fig.~\ref{fig:Causal} visualizes causal denoising. Diffusion ReRoll adopts this per-token noise-level view, but uses a different training distribution and denoising schedule.

\section{Diffusion ReRoll}
\label{sec:diffusion_reroll}

\begin{wrapfigure}{r}{0.33\textwidth}
  \vspace{-1.2em}
  \centering
  \includegraphics[width=0.32\textwidth]{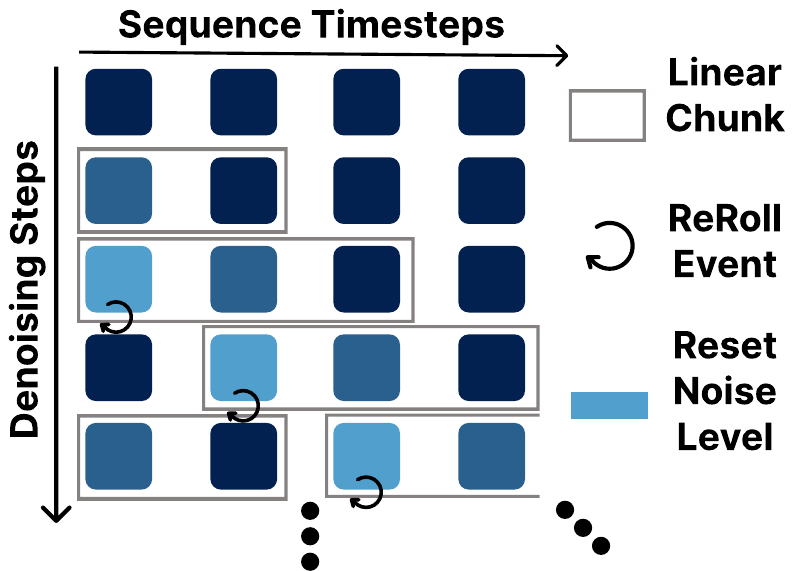}
  \caption{Part of a simplified Diffusion ReRoll schedule.}
  \label{fig:DRBrief}
  \vspace{-1.5em}
\end{wrapfigure}

Diffusion ReRoll (DR) is a diffusion-based sequential prediction framework that actively manages token-wise noise levels through a schedule matrix. 
The schedule matrix is defined over sequence timesteps and denoising steps, specifying the noise level assigned to each token at each denoising step. 
Unlike full-sequence diffusion, DR imposes piecewise-linear noise-level patterns over the sequence. 
Each linear region is called a \emph{linear chunk}. 
These chunks are not explicit trajectory partitions; they are virtual patterns in the schedule matrix that control how different horizon regions are denoised and re-noised.

During denoising, when a chunk edge reaches the reset noise level, a ReRoll event re-noises the corresponding token region and attaches it to an adjacent chunk. 
The schedule therefore does not monotonically denoise the whole horizon. 
Instead, some regions continue denoising while selected regions return to higher noise levels for further refinement under updated context. Fig.~\ref{fig:DRBrief} visualizes this procedure.
This keeps the horizon revisable while preserving local chunk-wise structure.

DR does not require a new backbone architecture.
It only requires per-token noise-level conditioning, as in Diffusion Forcing~\cite{diffusion_forcing,forcing2,DFFine}.
We use full-attention Transformer denoisers.
Noise levels are assigned per sequence token, with image patches sharing
the level of their latent frame.

\subsection{Schedule Matrix for Diffusion ReRoll}
\label{subsec:denoising_matrix}


\begin{wrapfigure}{r}{0.65\textwidth}
  \vspace{-1.2em}
  \centering
  \includegraphics[width=0.64\textwidth]{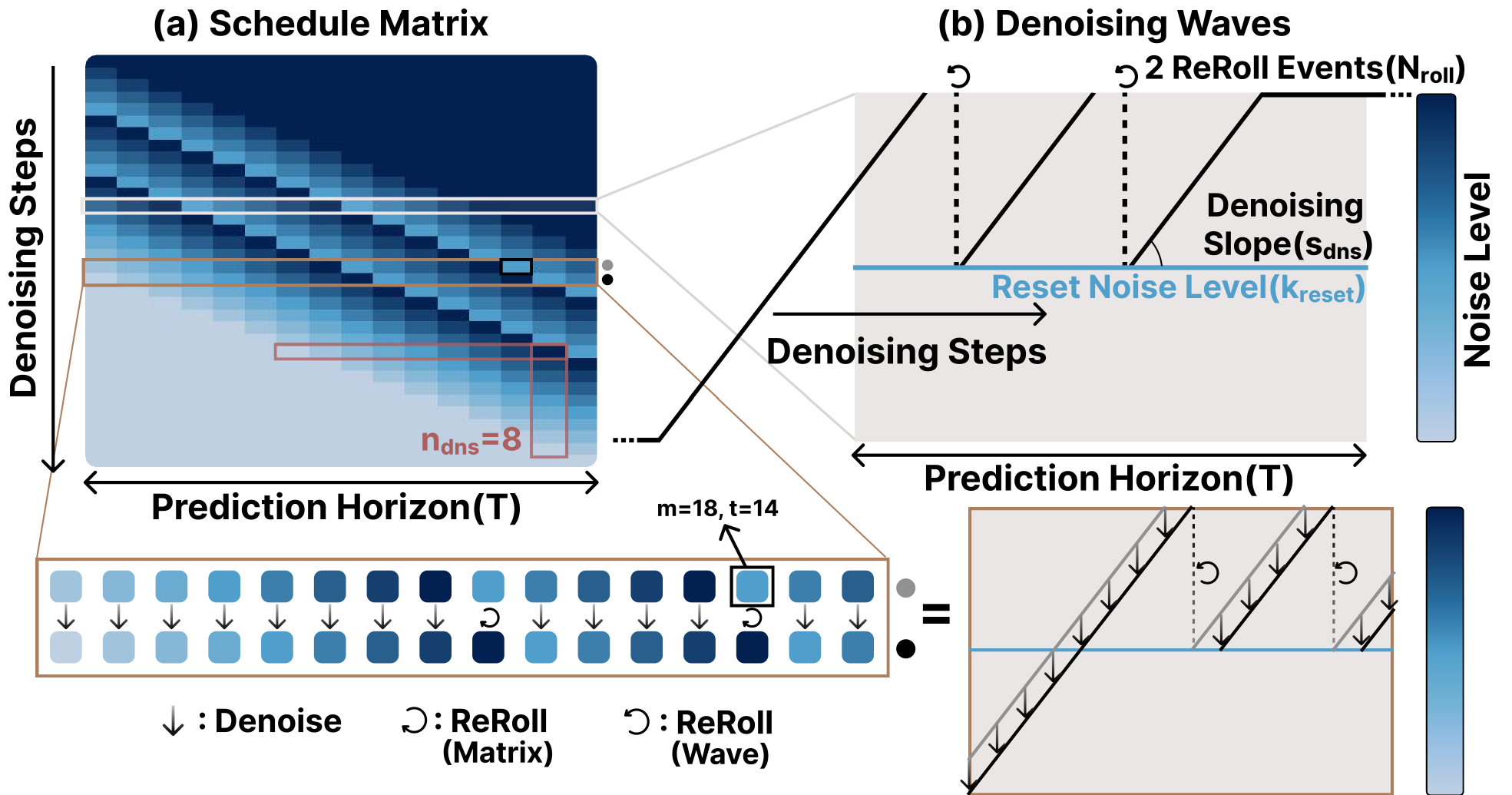}
  \caption{Representation of Diffusion ReRoll with (a) Schedule Matrix and (b) Denoising Waves.}
  \label{fig:RepReRoll}
  \vspace{-0.7em}
\end{wrapfigure}

We describe denoising with a schedule matrix $\mathbf{S}$, where
$\mathbf{S}_{m,t}$ denotes the noise level of token $t$ at row $m$.
Between consecutive rows, tokens are denoised with Denoising Diffusion Implicit Models (DDIM)~\cite{DDIM},
kept unchanged, or re-noised according to their noise-level transition.
While full-sequence diffusion assigns a shared noise level to all tokens,
Diffusion ReRoll constructs $\mathbf{S}$ from virtual piecewise-linear chunks,
which can be viewed as denoising waves across the prediction horizon.
Fig.~\ref{fig:RepReRoll} shows the same denoising schedule in matrix and wave representations.
When a wave reaches the reset noise level, the corresponding region is re-noised
and joins an adjacent wave.
The schedule is controlled by three parameters:
(i) denoising slope, (ii) reset noise level, and (iii) number of ReRoll events.

\paragraph{Denoising slope.}
Let $T$ be the prediction horizon, $K$ the maximum noise level, and
$n_{\mathrm{dns}}$ the number of denoising updates required for one token
to move from full noise $K$ to clean value $0$ within one uninterrupted
denoising wave.
We define the denoising slope as
$s_{\mathrm{dns}}=T/n_{\mathrm{dns}}$.
For example, $T=16$ and $n_{\mathrm{dns}}=8$ gives
$s_{\mathrm{dns}}=2$ in Fig.~\ref{fig:RepReRoll}.
A larger $s_{\mathrm{dns}}$ creates shorter chunks for local refinement,
while a smaller $s_{\mathrm{dns}}$ creates longer chunks with broader
temporal context.

\paragraph{Reset noise level.}
Let $k_{\mathrm{reset}}$ denote the reset noise level that triggers a
ReRoll event.
When a chunk edge reaches $k_{\mathrm{reset}}$, the corresponding region
is re-noised to full noise $K$.
If $k_{\mathrm{reset}}$ is too high, the region is reset before sufficient
denoising; if too low, it becomes nearly deterministic before revision.
In our experiments, intermediate values around the middle of the
noise-level range work robustly.

\paragraph{Number of ReRoll events.}
\begin{wrapfigure}{r}{0.35\textwidth}
  \vspace{-1.2em}
  \centering
  \includegraphics[width=0.34\textwidth]{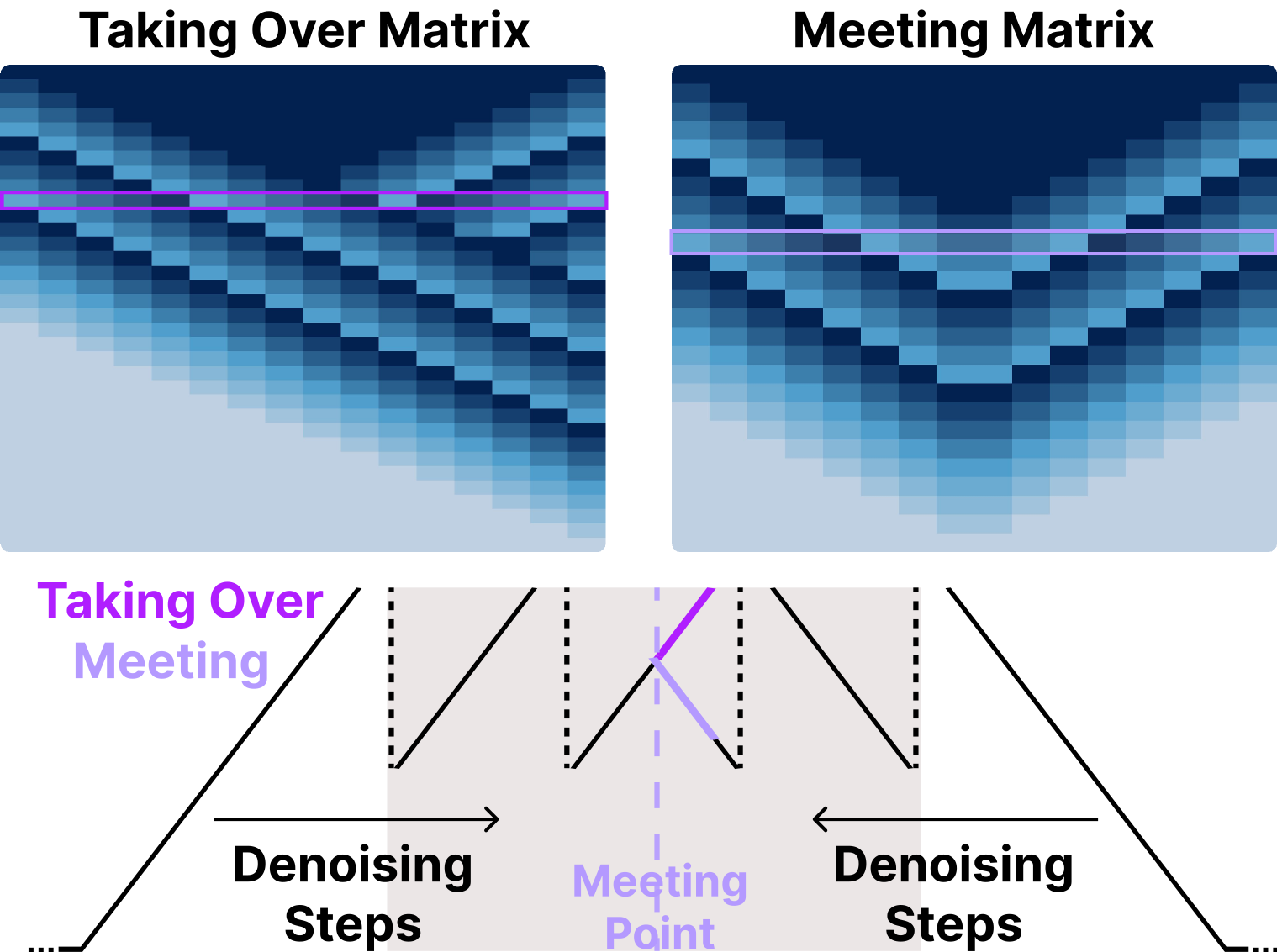}
  \caption{Two bidirectional schedule matrices are shown with denoising wave representation. }
  \label{fig:bidir}
  \vspace{-1.5em}
\end{wrapfigure}
Let $N_{\mathrm{roll}}$ be the number of ReRoll events. This parameter denotes the number of restart waves, equivalently
the maximum number of resets experienced by a token.
Too few events limit horizon-wide revision, while too many add redundant
re-noising and increase the number of denoising updates.
We choose $N_{\mathrm{roll}}$ to distribute structured revision across the horizon while keeping the denoising budget comparable to the baseline.

\paragraph{Bidirectional schedule matrix.}

The default \emph{forward schedule matrix} propagates chunks causally toward the future. 
When both initial and terminal constraints are available, we use a \emph{bidirectional schedule matrix}, where waves propagate from both ends of the horizon.
The two sides may use different $s_{\mathrm{dns}}$, $k_{\mathrm{reset}}$, and $N_{\mathrm{roll}}$. Their meeting point can also be shifted to control information flow between start and terminal constraints.
Since unresolved regions remain noisy and revisable, DR naturally supports non-causal and asymmetric schedule matrices.

As shown in Fig.~\ref{fig:bidir}, we consider two variants: a \emph{taking over matrix}, where the forward wave takes over after
overlap, and a \emph{meeting matrix}, where forward and backward waves meet at the meeting point. 
Both variants preserve the ReRoll principle, while using endpoint information for sequence revision differently. 
Details are provided in Appendix~\ref{subsec:Reverse}.


\subsection{Training with Linear Chunks}
\label{subsec:training_linear_chunks}

\begin{figure}[h!]
    \centering
    \includegraphics[width=1.0\linewidth]{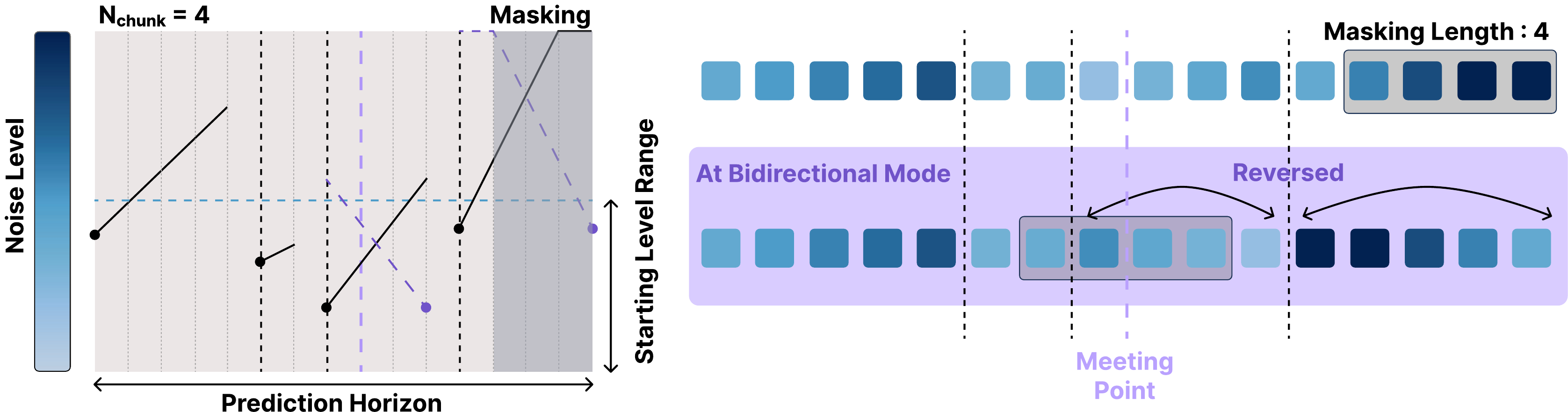}
    \caption{
    Training with randomized linear noise-level chunks.
    We sample piecewise-linear noise profiles with random chunk numbers, slopes, starting noise levels, and masking lengths.
    Bidirectional mode requires additional modifications, highlighted in purple.
    }
    \label{fig:Training_procedure}
\end{figure}

DR uses piecewise-linear noise-level chunks at deployment. 
Training only on the fixed schedule matrix would expose the model to narrow noise configurations, so we use randomized linear chunks that preserve local linear structure while covering a broader noise-level distribution.

Fig.~\ref{fig:Training_procedure} illustrates randomized linear-chunk noise assignment.
For each training sequence, we sample the number of chunks $N_{\mathrm{chunk}}$ and split the horizon into $N_{\mathrm{chunk}}$ contiguous segments.
Each segment receives a starting noise level and chunk slope, defining a linear noise profile over its tokens.
Thus, token-wise noise levels follow structured local patterns rather than being sampled independently.

Let $s_{\mathrm{dns}}$ be the deployment denoising slope.
By default, we sample $N_{\mathrm{chunk}}\in\{1,\ldots,N_{\max}\}$, with
$N_{\max}\approx 2s_{\mathrm{dns}}$, sample chunk slopes from
$[0.5,2.5]s_{\mathrm{dns}}$, and sample starting levels from a
zero-to-middle noise range except for the first chunk.
$N_{\max}$, the chunk slope range, and the starting level range are
training hyperparameters that can be adjusted for different tasks or deployment matrices.

We also apply masking to match unresolved regions at deployment. 
Forward training masks a random suffix with full noise. 
Bidirectional training masks a symmetric region around the meeting point and reverses right-side chunks, exposing the model to revision from both directions.

Given token-wise noise levels $k_{1:T}$, we train with the standard diffusion denoising loss, changing only the token-wise noise sampling distribution. 
Fig.~\ref{fig:TrainingAlignment} shows that this linear-chunk training is important for effective ReRoll deployment. 
Details for training are explained in Appendix~\ref{sec:TrainingDetails}.



\section{Experiments}
\label{sec:experiments}

We evaluate DR on long-horizon planning, policy learning, unified
video-action modeling, and real-world navigation.
We follow prior codebases and protocols closely~\cite{diffusion_forcing,mctd,diffusion_policy,OAT,uwm}.
Within each task family, DF and DR use matched architectures and differ
only in training noise distribution and deployment schedule.
DF uses causal denoising in forward settings and the corresponding
bidirectional schedule without ReRoll when terminal constraints are available.
Full-sequence baselines follow their task-specific implementations.
DR uses near-shared schedule parameters across the four task families to
avoid task-specific tuning.
Full details are in Appendix~\ref{sec:ExperimentSettings}.

At inference, full-sequence baselines use their standard samplers,
while DF and DR use schedule-matrix sampling.
DR uses no more denoising updates than the corresponding full-sequence
baseline in the main evaluations.

\subsection{Effect of ReRoll and Training Alignment}
\label{subsec:reroll_events}

  \vspace{-0.2em}

\begin{figure}[h!]
    \centering
    \includegraphics[width=1.0\linewidth]{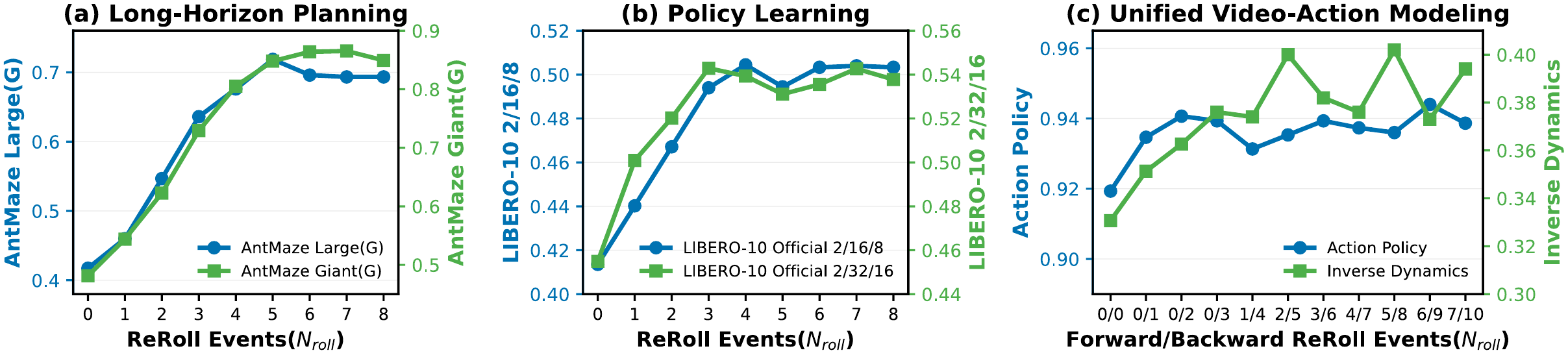}
    \caption{
    Effect of ReRoll events in (a) long-horizon planning, (b) policy learning,
    and (c) unified video-action modeling.
    Values are average success rates.
    In (a), (G) denotes guidance-based planning; in (b), labels follow
    Obs./Hor./Act. 
    }
    \label{fig:ReRollEvents}
\end{figure}

\begin{wrapfigure}{r}{0.295\textwidth}
  \vspace{-1.25em}
  \centering
  \includegraphics[width=0.29\textwidth]{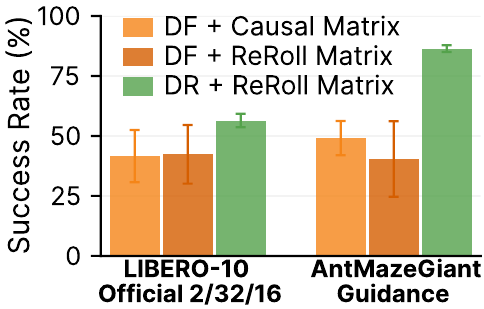}
    \caption{
    Effect of linear-chunk training on ReRoll performance. 
    }
  \label{fig:TrainingAlignment}
  \vspace{-1.0em}
\end{wrapfigure}

Before comparing DR with other methods, we examine the effects of ReRoll events
and linear-chunk training.
With zero ReRoll events, forward schedules reduce to causal denoising.
Fig.~\ref{fig:ReRollEvents} shows that adding ReRoll events generally improves
performance across the three simulation task families before saturation.

Fig.~\ref{fig:TrainingAlignment} further shows that applying the ReRoll matrix
to independently-noised DF does not consistently improve performance, whereas
DR with linear-chunk training benefits substantially.
Thus, effective revisable denoising requires both repeated cross-horizon revision through ReRoll events and training on aligned structured noise patterns.

Fig.~\ref{fig:ReRollEvents} varies the number of ReRoll events; the denoising slope and reset
noise level provide two additional controls over revision and are studied
in Appendix~\ref{sec:ChooseDenoising}.
Based on the observed saturation, we use six ReRoll events in the main
simulation experiments (six forward events for bidirectional schedules)
and three for real-world navigation to reduce deployment latency.

  \vspace{-0.34em}

\subsection{Long-Horizon Planning}
\label{subsec:planning}

  \vspace{-0.2em}


We evaluate Diffusion ReRoll on OGBench PointMaze and AntMaze~\cite{OgbenchBench},
which test consistent long-horizon planning toward distant goals.
Baselines are Diffusion Forcing (DF)~\cite{diffusion_forcing} with causal or
bidirectional denoising and Diffuser~\cite{diffuser} with full-sequence diffusion.
DF and DR share a Transformer backbone; Diffuser retains its standard U-Net,
as a full-sequence version of the DF/DR Transformer performed worse.
We largely follow the MCTD setup~\cite{mctd}, with details in
Appendix~\ref{subsec:Maze}.

We consider two protocols, training separate models for each method and protocol.
In \emph{guidance-based replanning}, trajectories are sampled with distance-based
goal guidance~\cite{classiferGuidance,guidance2} and replanning.
In \emph{goal inpainting}, the start and goal states are fixed and the model
generates the intermediate trajectory.
For controlled comparison, we train an inpainting-compatible DF variant using
the same masking scheme and corresponding bidirectional schedule as DR, but
without ReRoll events.

\begin{table}[h!]
\centering

\scriptsize
\setlength{\tabcolsep}{4.2pt}
\renewcommand{\arraystretch}{1.03}
\resizebox{0.87\linewidth}{!}{
\begin{tabular}{@{}llccc ccc@{}}
\toprule
\multirow{2}{*}{Mode}
& \multirow{2}{*}{Method}
& \multicolumn{3}{c}{AntMaze}
& \multicolumn{3}{c}{PointMaze} \\
\cmidrule(lr){3-5}
\cmidrule(lr){6-8}
& & Giant & Large & Medium
& Giant & Large & Medium \\
\midrule

\multirow{3}{*}{\shortstack[c]{Guidance\\Replan}}
& Diffuser
& 16.0\sdev{1.1}
& 42.1\sdev{4.2}
& 69.1\sdev{1.0}
& 26.5\sdev{4.9}
& 30.0\sdev{1.8}
& 46.4\sdev{0.8} \\

& Forcing
& \underline{49.1}\sdev{9.1}
& \underline{55.3}\sdev{5.7}
& \underline{93.9}\sdev{2.4}
& \underline{88.9}\sdev{4.0}
& \underline{60.8}\sdev{1.7}
& \underline{88.7}\sdev{7.6} \\

& ReRoll
& \textbf{86.4}\sdev{1.4}
& \textbf{69.6}\sdev{1.4}
& \textbf{98.0}\sdev{0.4}
& \textbf{99.3}\sdev{0.8}
& \textbf{76.3}\sdev{1.8}
& \textbf{98.8}\sdev{0.4} \\

\midrule

\multirow{3}{*}{Inpainting}
& Diffuser
& 33.2\sdev{1.4}
& \underline{68.1}\sdev{4.3}
& 88.3\sdev{1.6}
& 64.9\sdev{1.6}
& \underline{96.1}\sdev{0.5}
& \underline{92.0}\sdev{3.9} \\

& Forcing
& \underline{71.9}\sdev{4.0}
& 67.5\sdev{0.6}
& \underline{90.5}\sdev{0.6}
& \underline{94.5}\sdev{1.6}
& 95.7\sdev{1.8}
& 91.2\sdev{2.8} \\

& ReRoll
& \textbf{80.9}\sdev{3.6}
& \textbf{74.4}\sdev{0.4}
& \textbf{93.7}\sdev{0.6}
& \textbf{99.1}\sdev{0.2}
& \textbf{99.5}\sdev{0.6}
& \textbf{96.9}\sdev{2.7} \\

\bottomrule
\end{tabular}
}
\caption{
Long-horizon planning success rates (\%) on OGBench PointMaze and AntMaze.
Values are mean $\pm$ std over three training seeds.
Best means are bolded; second-best are underlined.
}
\label{tab:planning_results}
\end{table}

\begin{figure}[h!]
    \centering
    \includegraphics[width=1.0\linewidth]{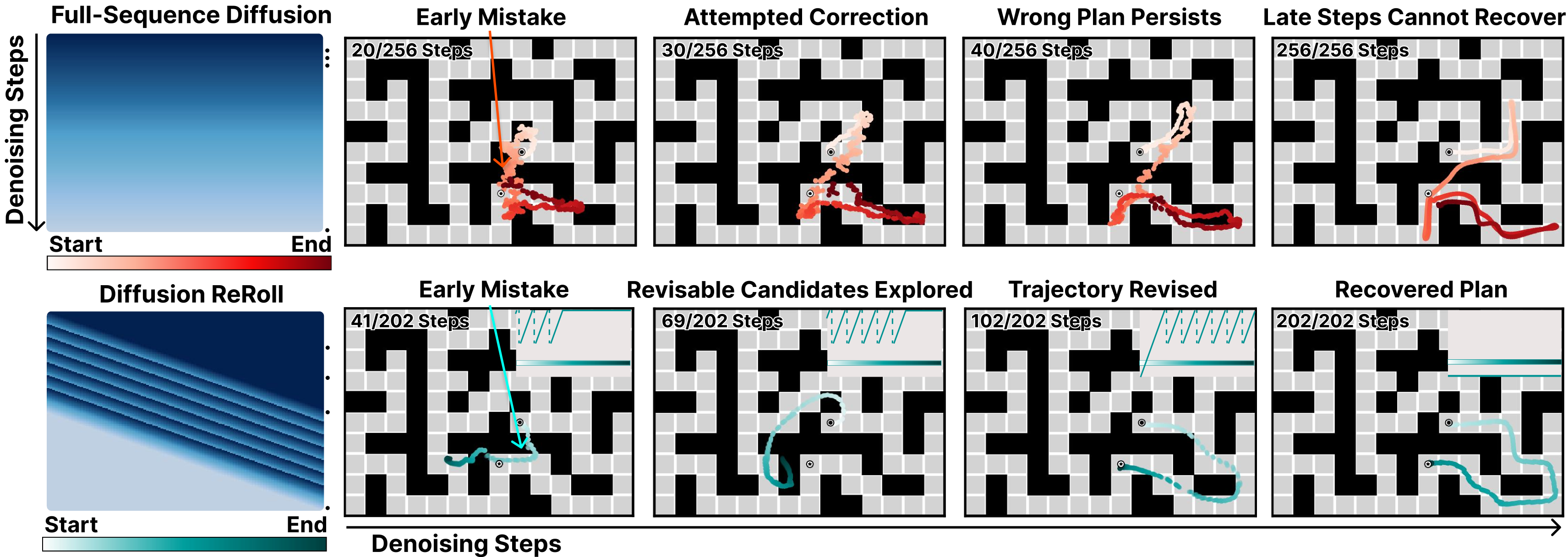}
    \caption{
    Denoising trajectories of Diffuser and Diffusion ReRoll are shown.
    Diffusion ReRoll keeps trajectories revisable, enabling recovery from early planning errors. Goal-directed guidance is used for planning on AntMaze Giant. Predicted clean trajectories $\hat{x}_0$ are visualized.
    }
    \label{fig:Explore}
\end{figure}

\vspace{0.3em}

Table~\ref{tab:planning_results} shows that DR achieves the best performance under both protocols across all maze variants. 
The gains are pronounced on AntMaze Giant and AntMaze Large, where long-horizon consistency is most critical. 
Fig.~\ref{fig:Explore} illustrates the mechanism. Full-sequence diffusion can commit to an early incorrect plan, while ReRoll keeps parts of the trajectory revisable and recovers a valid path.

\subsection{Policy Learning}
\label{subsec:policy_learning}
\vspace{0.3em}

We evaluate DR in diffusion-policy-style action prediction on LIBERO-10 multi-task~\cite{libero} and RoboCasa~\cite{robocasa} single-task learning.
This tests whether DR improves action-chunk prediction across observation histories and horizons.
For DF and DR, we modify the Chi-Transformer~\cite{diffusion_policy} so each action token receives its own noise level, while using each method's action representation.
LIBERO-10 policies are trained from scratch, with scalar \texttt{task\_uid} as the task identifier, while RoboCasa uses a pretrained vision encoder.
Other settings are shared when possible. Details are in Appendix~\ref{subsubsec:EvaluationDP}.

\vspace{0.1em}

We found that LIBERO-10 evaluation is sensitive to the reset behavior during object spawning, particularly whether the gripper remains fixed or is opened while objects settle. 
Opening the gripper makes the initial state closer to the training demonstration distribution. 
Because existing codebases differ in this initialization protocol, we report results under both the official initialization and an open-gripper initialization. 
Details and evaluation concerns are provided in Appendix~\ref{subsubsec:liberoIssue}.

\begin{table}[h!]

\vspace{0.6em}

\centering

\scriptsize
\setlength{\tabcolsep}{2.9pt}
\renewcommand{\arraystretch}{1.08}
\resizebox{\linewidth}{!}{
\begin{tabular}{lccccc ccccc}
\toprule
\multirow{2}{*}{Method}
& \multicolumn{5}{c}{LIBERO-10 Official \;{\scriptsize (Obs./Hor./Act.)}}
& \multicolumn{5}{c}{LIBERO-10 Open \;{\scriptsize (Obs./Hor./Act.)}} \\
\cmidrule(lr){2-6}
\cmidrule(lr){7-11}
& 2/16/8 & 2/32/16 & 2/64/16 & 4/32/16 & 8/32/16
& 2/16/8 & 2/32/16 & 2/64/16 & 4/32/16 & 8/32/16 \\
\midrule

Diff. Policy
& 28.1\sdev{3.8}
& 41.1\sdev{8.2}
& 35.1\sdev{2.7}
& 42.2\sdev{8.8}
& 53.8\sdev{0.9}
& \underline{76.2}\sdev{0.5}
& \underline{65.5}\sdev{1.9}
& 57.9\sdev{2.0}
& 69.1\sdev{1.4}
& 64.7\sdev{1.9} \\

Forcing
& \underline{37.3}\sdev{12.0}
& \underline{41.6}\sdev{10.9}
& \underline{41.5}\sdev{4.1}
& \underline{63.0}\sdev{2.7}
& \underline{62.2}\sdev{4.0}
& 68.3\sdev{1.7}
& 60.1\sdev{1.1}
& \underline{59.0}\sdev{2.9}
& \underline{70.7}\sdev{0.8}
& \underline{75.1}\sdev{2.4} \\

ReRoll
& \textbf{51.0}\sdev{24.1}
& \textbf{56.4}\sdev{2.8}
& \textbf{63.4}\sdev{6.4}
& \textbf{74.5}\sdev{2.6}
& \textbf{68.1}\sdev{7.7}
& \textbf{82.5}\sdev{3.6}
& \textbf{75.6}\sdev{3.1}
& \textbf{70.7}\sdev{1.0}
& \textbf{81.5}\sdev{2.9}
& \textbf{82.9}\sdev{0.8} \\

\bottomrule
\end{tabular}
}
\caption{
LIBERO-10 multi-task policy learning results under two initialization protocols.
We report success rates (\%) as mean $\pm$ standard deviation over three training seeds, using the best checkpoint from each run.
Best mean results are bolded and second-best mean results are underlined.
}

\vspace{-0.7em}

\label{tab:libero_policy}
\end{table}

\begin{table}[h!]
\centering

\scriptsize
\setlength{\tabcolsep}{3.2pt}
\renewcommand{\arraystretch}{1.06}
\resizebox{\linewidth}{!}{
\begin{tabular}{lccc ccc ccc ccc}
\toprule
\multirow{2}{*}{Method}
& \multicolumn{3}{c}{CloseSingleDoor}
& \multicolumn{3}{c}{CoffeePressButton}
& \multicolumn{3}{c}{CoffeeServeMug}
& \multicolumn{3}{c}{PnPCounterToSink (300 Demos)} \\
\cmidrule(lr){2-4}
\cmidrule(lr){5-7}
\cmidrule(lr){8-10}
\cmidrule(lr){11-13}
& 2/32/16 & 2/64/16 & 4/32/16
& 2/32/16 & 2/64/16 & 4/32/16
& 2/32/16 & 2/64/16 & 4/32/16
& 2/32/16 & 2/64/16 & 4/32/16 \\
\midrule

Diff. Policy
& 52.0\sdev{2.0}
& 52.7\sdev{2.3}
& \underline{56.7}\sdev{8.3}
& 76.0\sdev{5.3}
& 54.7\sdev{8.3}
& 80.0\sdev{4.0}
& \textbf{43.3}\sdev{3.1}
& 32.7\sdev{4.2}
& \textbf{39.3}\sdev{5.8}
& \textbf{30.7}\sdev{3.1}
& \underline{22.0}\sdev{4.0}
& 24.7\sdev{2.3} \\

Forcing
& \underline{58.7}\sdev{8.1}
& \textbf{68.0}\sdev{3.5}
& \textbf{65.3}\sdev{4.6}
& \underline{79.3}\sdev{8.1}
& \underline{70.7}\sdev{1.2}
& \textbf{88.7}\sdev{5.0}
& 34.0\sdev{7.2}
& \textbf{41.3}\sdev{4.2}
& \underline{36.0}\sdev{5.3}
& \underline{28.7}\sdev{5.0}
& \textbf{29.3}\sdev{1.2}
& \textbf{31.3}\sdev{6.1} \\

ReRoll
& \textbf{67.3}\sdev{5.8}
& \underline{62.0}\sdev{5.3}
& \textbf{65.3}\sdev{6.4}
& \textbf{84.0}\sdev{5.3}
& \textbf{80.7}\sdev{7.0}
& \underline{86.0}\sdev{4.0}
& \underline{40.0}\sdev{5.3}
& \underline{34.7}\sdev{4.6}
& \underline{36.0}\sdev{3.5}
& \underline{28.7}\sdev{3.1}
& \textbf{29.3}\sdev{6.1}
& \underline{29.3}\sdev{2.3} \\

\bottomrule
\end{tabular}
}
\caption{
RoboCasa single-task policy learning results.
We report success rates (\%) as mean $\pm$ standard deviation over three training seeds, using the best checkpoint from each run.
Best mean results are bolded and second-best mean results are underlined.
}
\vspace{-0.7em}

\label{tab:robocasa_policy}
\end{table}


Table~\ref{tab:libero_policy} shows that DR achieves the highest mean performance
across LIBERO-10 settings under both initialization protocols.
These long-horizon, multi-stage tasks can provide informative cross-horizon
context for revising action predictions.
In contrast, Table~\ref{tab:robocasa_policy} shows less consistent gains on
the selected RoboCasa tasks, which target atomic skills.
We hypothesize that ReRoll is most beneficial when strong temporal dependencies
make cross-horizon revision informative.

\paragraph{Reducing inference latency.}
\begin{wrapfigure}{r}{0.47\textwidth}
  \vspace{-1.0em}
  \centering
  \includegraphics[width=0.469\textwidth]{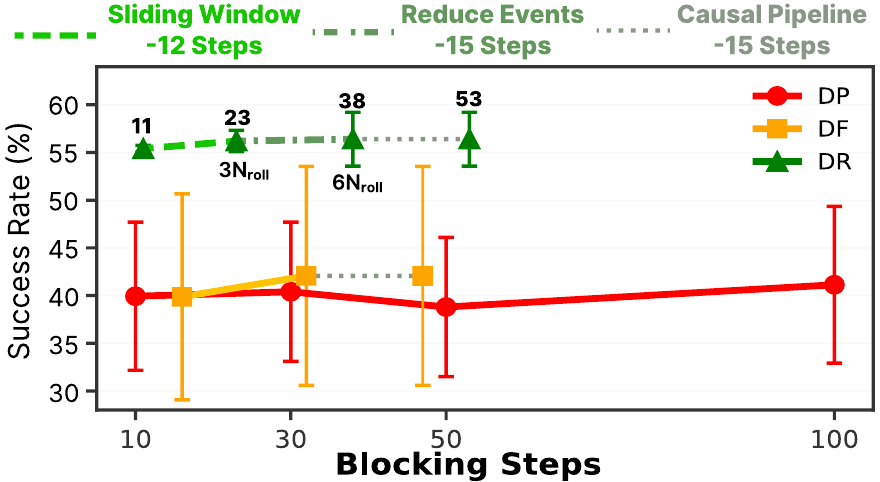}
  \caption{
  Success versus blocking denoising updates on LIBERO-10 Official
  with \textsc{Obs./Hor./Act.}=2/32/16. Error bars show $\pm$ standard deviation across three training seeds.
  }
  \label{fig:inference_latency}
  \vspace{-0.5em}
\end{wrapfigure}

To support real-time deployment, we present practical inference-time
optimizations for ReRoll.
These include reducing ReRoll events, carrying the unexecuted suffix into
the next prediction with a sliding window, and pipelining during the final
causal denoising phase to overlap computation with action execution.
Pipelining reduces blocking latency rather than total computation.
As an example, Fig.~\ref{fig:inference_latency} illustrates these
optimizations on LIBERO-10 Official.
After the initial prediction, DR requires only 11 blocking updates instead of 53,
while retaining higher mean success than the tested DP and DF configurations.
Implementation details are provided in
Appendix~\ref{sec:InferenceOptimization}.

\subsection{Unified Video-Action Modeling}
\label{subsec:uwm}

\begin{table}[h!]
\centering
\scriptsize
\setlength{\tabcolsep}{3.2pt}
\renewcommand{\arraystretch}{1.06}
\resizebox{\linewidth}{!}{
\begin{tabular}{lcccccc}
\toprule
Method
& \makecell{Act. Pol.$\uparrow$ (\%)}
& \makecell{Joint Pol.$\uparrow$ (\%)}
& \makecell{Inv. Dyn.$\uparrow$ (\%)}
& \makecell{OOD Act.$\uparrow$ (\%)}
& \makecell{OOD Joint$\uparrow$ (\%)}
& \makecell{Zero-Act. Err.$\downarrow$ (m)} \\
\midrule

UWM
& 86.5\sdev{1.9}
& 86.7\sdev{1.2}
& 26.7\sdev{3.6}
& 56.6\sdev{3.3}
& 58.1\sdev{0.8}
& 0.328\sdev{0.004} \\

UWM + DF
& \underline{90.8}\sdev{0.7}
& \underline{90.6}\sdev{3.7}
& \underline{32.0}\sdev{1.3}
& \underline{61.1}\sdev{2.2}
& \underline{60.7}\sdev{2.7}
& \underline{0.317}\sdev{0.012} \\

UWM + DR
& \textbf{94.4}\sdev{0.9}
& \textbf{93.9}\sdev{0.8}
& \textbf{37.3}\sdev{0.8}
& \textbf{67.9}\sdev{2.8}
& \textbf{69.7}\sdev{2.0}
& \textbf{0.301}\sdev{0.003} \\

\bottomrule
\end{tabular}
}
\caption{
Unified world model prediction results on LIBERO-10.
We report mean $\pm$ standard deviation over three fine-tuning seeds.
}
\vspace{-0.7em}
\label{tab:uwm_results}
\end{table}

\begin{wrapfigure}{r}{0.25\textwidth}
  \vspace{-1.5em}
  \centering
  \includegraphics[width=0.245\textwidth]{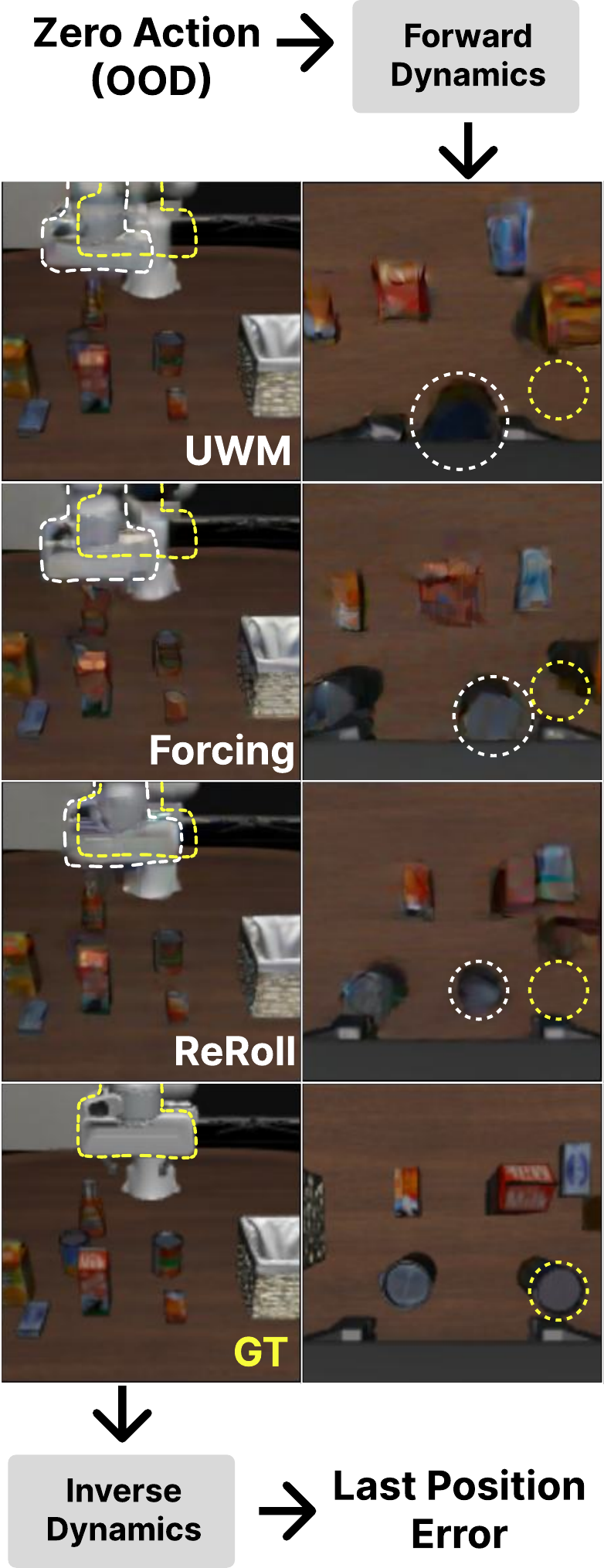}
  \caption{Zero-action consistency evaluation and generated final-frame examples.}
  \label{fig:ZeroAction}
  \vspace{-2.0em}
\end{wrapfigure}

\vspace{-0.2em}

We next evaluate DR in a unified video-action prediction setting. 
Unified World Models (UWM) support action prediction, future-visual prediction, and joint video-action generation within a single model~\citep{uwm}. 
Here, we test whether DR improves policy generation, inverse dynamics, OOD robustness, and action-video consistency. 
Although such models can exploit video-only data without action annotations~\cite{chen2025large,videoReasoner,videoRobotPolicies}, our focus is downstream control performance and consistency.


We apply DR to UWM by assigning structured linear noise levels over a joint sequence of 32 action tokens and two future latent frames, with one noise level per frame.
Models are pretrained on LIBERO-90 and fine-tuned on each LIBERO-10 task. 
We compare UWM, UWM + DF, and UWM + DR under action-only policy prediction, joint video-action policy prediction, inverse dynamics, and OOD evaluation. 
Policy evaluation uses \textsc{Obs./Hor./Act.}=2/32/32.
For OOD evaluation, the object initialization range is enlarged by
$0.03\,\mathrm{m}$ and task-irrelevant background objects are removed.
Details are in Appendix~\ref{subsubsec:uwmeval}.


Table~\ref{tab:uwm_results} shows that DR improves policy prediction and inverse dynamics, with the largest gains under visual OOD evaluation. 
Additional analyses of schedule matrix applications are provided in Appendix~\ref{subsubsec:UWMReverse}.


We further test action-video consistency with a zero-action intervention.
Since task-fine-tuned models are trained on successful behaviors, this condition is OOD.
We fix future action tokens to zero, generate future images, and apply inverse
dynamics to measure implied terminal end-effector motion
(Fig.~\ref{fig:ZeroAction}); details are in Appendix~\ref{subsec:zero_action_details}.
DR achieves the lowest zero-action error, indicating the strongest consistency
under this diagnostic.

\subsection{Real-World Navigation}
\label{sec:real_navigation}

\begin{wrapfigure}{r}{0.68\textwidth}
  \vspace{-1.5em}
  \centering
  \includegraphics[width=0.678\textwidth]{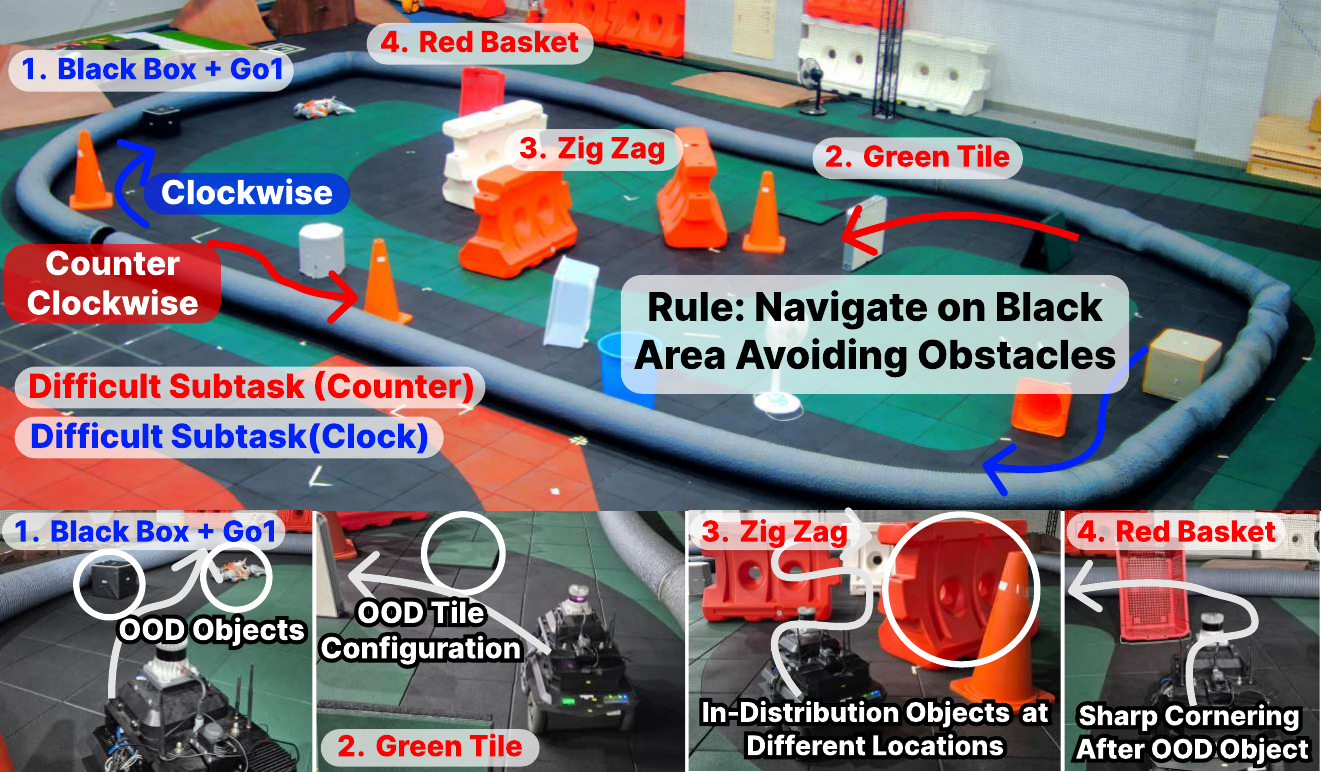}
  \caption{
  Real-world navigation course with four evaluated subtasks.
  The robot must remain in black drivable regions while avoiding
  green restricted regions and obstacles.
  }
  \label{fig:Navigation}
  \vspace{-1.0em}
\end{wrapfigure}

  \vspace{-0.25em}

We evaluate real-world visual navigation on a Clearpath Jackal using
approximately 40 minutes of demonstrations with randomized obstacle
positions and lighting.
Evaluation uses held-out configurations of the same course with unseen
objects, altered floor appearances, and reduced illumination
(Fig.~\ref{fig:Navigation}).
This tests closed-loop deployment under real sensing, 
and visual distribution shifts.

DP, DF, and DR share a modified Chi-Transformer architecture with
onboard inference and 5\,Hz closed-loop control.
Each prediction contains 16 waypoints; the first eight condition a shared
inverse-dynamics controller that outputs one velocity command per update
(Obs./Hor./Act.=1/16/8).
DP, DF, and DR use 20, 23, and 20 sampler updates, respectively.
DR uses the forward matrix with $N_{\mathrm{roll}}=3$, while other
ReRoll parameters follow the policy-learning setting.
Shared navigation training settings are detailed in
Appendix~\ref{subsec:RealNavigation}.

\begin{figure}[h!]
  \centering
  \includegraphics[width=\linewidth]{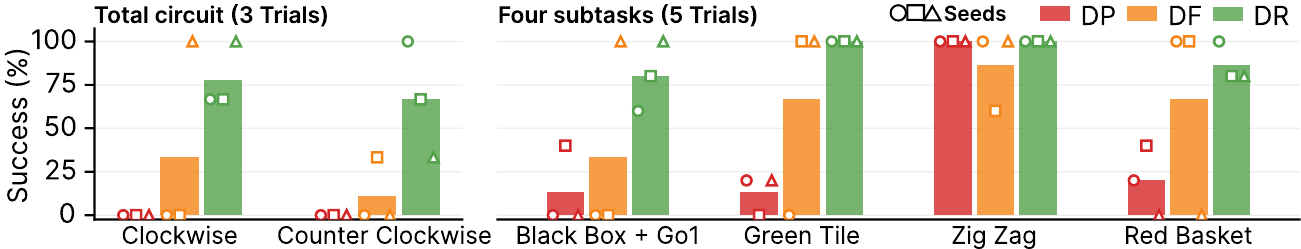}
   \vspace{-0.7em}

  \caption{
  Navigation success across three training seeds per method, with
  three trials per circuit direction and five per subtask.
  Bars show means, markers show individual seeds.
  }
  \label{fig:OnlyNav}
  \vspace{-0.4em}

\end{figure}

We evaluate clockwise and counterclockwise full-circuit completion under
OOD conditions and four challenging subtasks.
The four subtasks are constituent sections of the full circuit and are
also evaluated separately.
Fig.~\ref{fig:OnlyNav} shows strong performance on the in-distribution
Zig-Zag task for all methods, while DR achieves the highest mean success
in both directions and all three OOD subtasks.

  \vspace{-0.4em}

\section{Limitations}
\label{sec:limitations}
This work has three limitations.
First, although near-shared parameters are effective across the four
task families, new tasks may require tuning of the training noise
distribution and deployment schedule to balance performance and
inference cost.
Second, our evidence for cross-horizon revision is empirical; we do not
provide a theoretical characterization of when revision improves
predictions or guarantees of convergence.
Third, evaluation is limited to Transformer denoisers and benchmark-scale
datasets. Validation in large general-purpose robot models
remains future work.
\vspace{-0.3em}

\section{Conclusion}
\vspace{-0.3em}

We introduced Diffusion ReRoll, a schedule matrix-based framework for
revisable robotic sequence generation.
By combining linear-chunk training with selective re-noising, ReRoll
allows horizon regions to revise one another using updated context.
Experiments demonstrate benefits in long-horizon planning, multi-task
action prediction, and unified video-action modeling, with further validation
on real-world navigation.
Together, these results support structured noise scheduling as a
mechanism for cross-horizon revision in robotic prediction.


	




\clearpage
\acknowledgments{
This work was supported by the Agency for Defense Development grant funded by the Korean Government in 2026.
We thank Seongjun Bae and Junyoung Kim for their assistance with the real-world navigation deployment setup, Eungjoe Kang for helpful feedback on the manuscript, and the anonymous reviewers for their constructive comments.
}

\bibliography{corl_2026_template_cameraready/ref}  

\clearpage

\appendix
\part*{Appendix}

\section{Choosing the Schedule Matrix}
\label{sec:ChooseDenoising}

\subsection{Schedule-Matrix Sampling and DDIM Updates}
\label{subsec:DDIM}

This section describes the schedule-matrix samplers used by Diffusion
Forcing and Diffusion ReRoll.
The schedule matrix stores discrete noise levels rather than raw training
diffusion timesteps.
Let $\mathbf{S}_{m,t}\in\{0,\ldots,K\}$ denote the noise level of token
$t$ at schedule row $m$, where $0$ is clean and $K$ is the maximum-noise
level used by the sampler.
Each schedule level is mapped to a training diffusion timestep by
\[
    \tau:\{0,\ldots,K\}
    \rightarrow
    \{-1,0,\ldots,K_{\mathrm{diff}}-1\},
\]
where $K_{\mathrm{diff}}$ is the number of training diffusion steps.
We set $\tau(0)=-1$, with $\bar{\alpha}_{\tau(0)}=1$, and
$\tau(K)=K_{\mathrm{diff}}-1$.

For a transition from row $m$ to $m+1$, let
\[
    k=\mathbf{S}_{m,t},
    \qquad
    \ell=\mathbf{S}_{m+1,t},
\]
and define
\[
    a_k=\bar{\alpha}_{\tau(k)},
    \qquad
    a_{\ell}=\bar{\alpha}_{\tau(\ell)}.
\]
For $k>0$, under our $x_0$-prediction parameterization, the denoising
network predicts a clean token $\hat{x}_t^0$ from $x_t^k$.
The corresponding predicted noise is
\begin{equation}
    \hat{\epsilon}_t
    =
    \frac{
        x_t^k-\sqrt{a_k}\hat{x}_t^0
    }{
        \sqrt{1-a_k}
    }.
\end{equation}

For a denoising transition $\ell<k$, we apply the DDIM update
\begin{equation}
x_t^\ell
=
\sqrt{a_{\ell}}\hat{x}_t^0
+
\sqrt{1-a_{\ell}-\sigma_{k\rightarrow\ell}^2}\,
\hat{\epsilon}_t
+
\sigma_{k\rightarrow\ell}z,
\qquad
z\sim\mathcal{N}(0,I),
\end{equation}
where
\begin{equation}
\sigma_{k\rightarrow\ell}
=
\eta
\sqrt{
\frac{1-a_k/a_{\ell}}{1-a_k}
(1-a_{\ell})
}.
\end{equation}
We use $\eta=0$ in all experiments, so downward DDIM transitions are
deterministic and introduce no additional noise.

A schedule transition may also preserve or increase a token's noise level.
If $\ell=k$, the token is left unchanged.
If $\ell>k$, the transition is a ReRoll reset rather than a reverse DDIM
update, and we apply forward noising:
\begin{equation}
x_t^\ell
=
\sqrt{\frac{a_{\ell}}{a_k}}x_t^k
+
\sqrt{1-\frac{a_{\ell}}{a_k}}\epsilon,
\qquad
\epsilon\sim\mathcal{N}(0,I).
\end{equation}
Thus, decreasing noise levels use DDIM denoising, equal levels leave the
token unchanged, and increasing levels use explicit forward noising.
Reset transitions remain stochastic even with $\eta=0$, since fresh
Gaussian noise is injected during re-noising.
This allows tokens to follow different noise-level trajectories according
to the schedule matrix.

For guidance-based sampling, the guidance term modifies the model-predicted
direction before the DDIM update; the schedule transition itself is unchanged.

\subsection{Bidirectional Schedule Matrices}
\label{subsec:Reverse}

\begin{figure}[h!]
    \centering
    \includegraphics[width=0.9\linewidth]{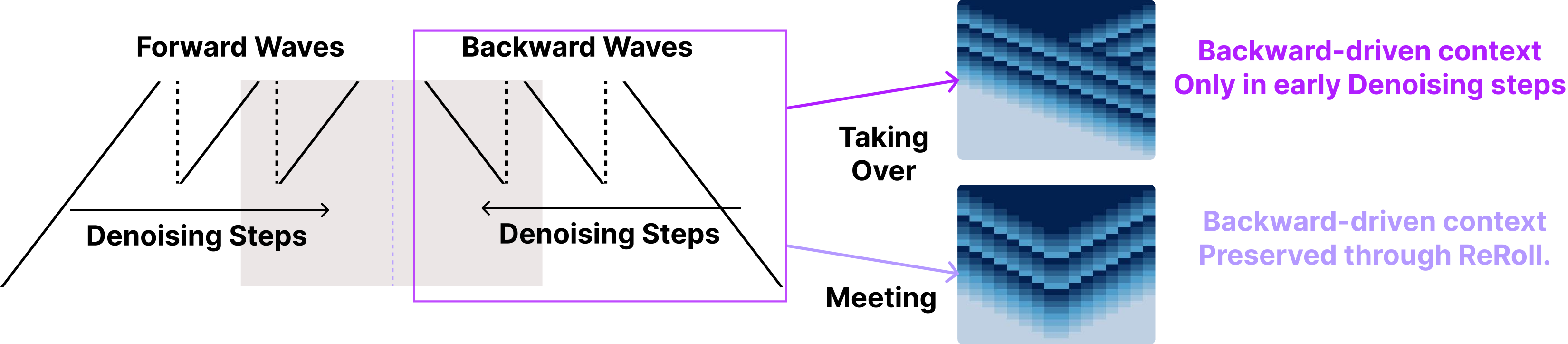}
    \caption{Bidirectional variants introduce backward waves from the terminal.
The taking over matrix uses goal-side derived context early and then returns to forward revision, while the meeting matrix preserves bidirectional revision. }
    \label{fig:Bidirectional_Comparison}
\end{figure}

Diffusion ReRoll can use different schedule matrices depending on the structure of the task. 
When both initial and terminal constraints are available, such as goal inpainting or future-image conditioning, it is useful to introduce denoising waves from both ends of the horizon. 
We consider two bidirectional variants: the \emph{meeting matrix} and the \emph{taking over matrix}.

The \emph{meeting matrix} introduces forward and backward waves that repeatedly revise the sequence from both directions. 
This creates a strongly non-causal denoising process, where non-causal information from the terminal side can continuously influence earlier segments. 
Such behavior might be beneficial for some tasks. However, performance can degrade on tasks where causal deployment structure is important.

The \emph{taking over matrix} uses terminal information more conservatively. 
Backward waves are used in the early part of denoising to provide goal-side context, but they are later overwritten by forward causal waves. 
This is useful for some tasks, where persistent non-causal influence from the future can introduce an undesirable deployment bias. 
Thus, the taking over matrix allows endpoint information to guide early revision while preserving a more causal final prediction structure.

These variants show that the ReRoll schedule matrix is not a fixed schedule, but a design interface for controlling how information flows across the horizon. 
We evaluate their use in two settings below.

\begin{wrapfigure}{r}{0.4\textwidth}
  \vspace{-1.0em}
  \centering
  \includegraphics[width=0.39\textwidth]{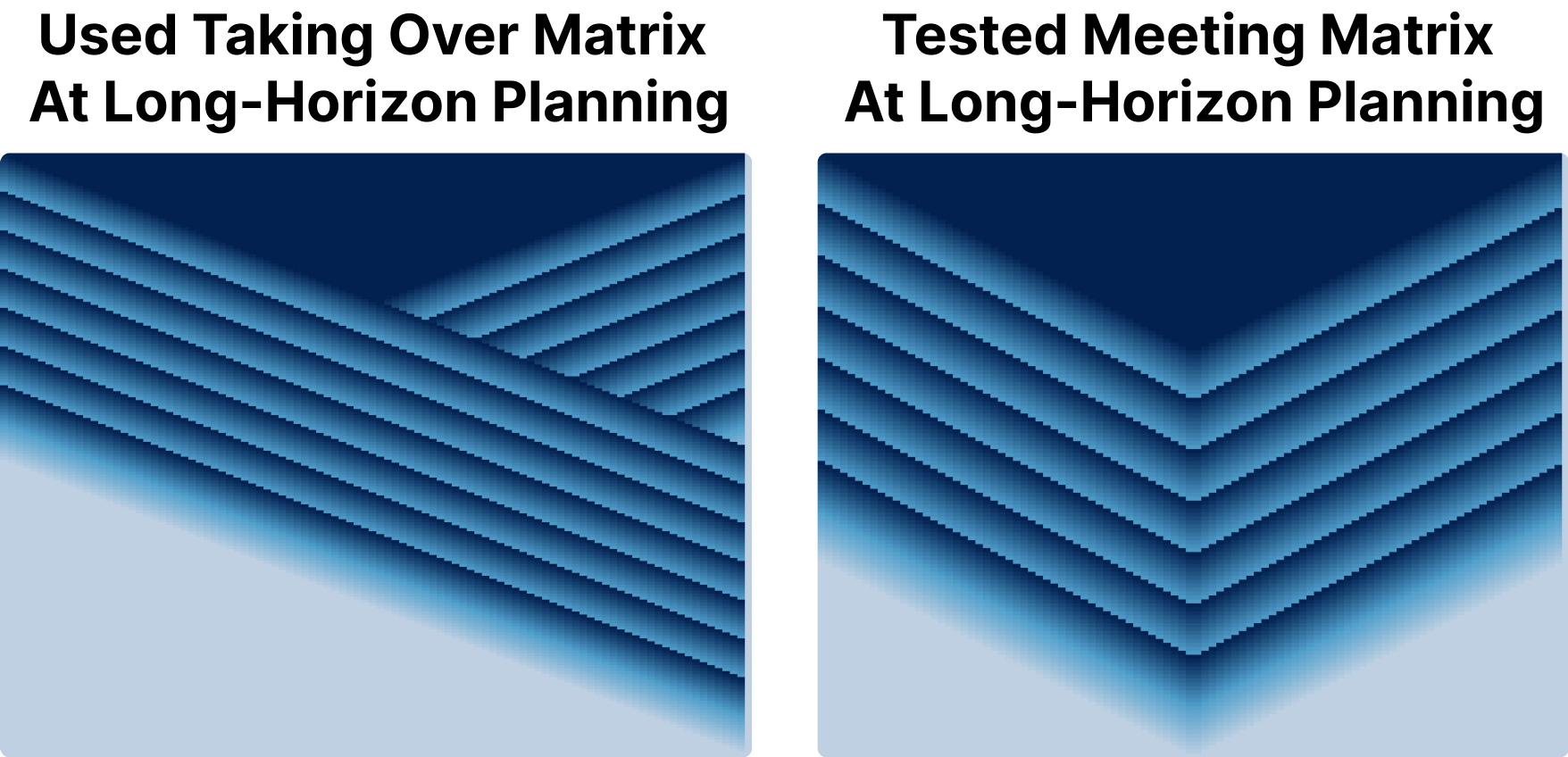}
\caption{
Bidirectional schedule matrices tested on OGBench maze planning.
The taking over matrix performs slightly better.
}
  \label{fig:Bidirectional_Comparison}
\end{wrapfigure}

\subsubsection{Applications to Maze Planning}
\label{subsubsec:MazeReverse}

For maze planning, the bidirectional schedule matrix is mainly used in the goal-inpainting setting, where both start and goal states are given. 
In the main experiments, we use the \emph{taking over matrix}. 
The meeting and taking over matrices show similar overall performance, but the taking over matrix is slightly better when execution is restricted to a planned prefix.

Although the main bidirectional maze experiments use goal inpainting, the same idea can also be applied to guidance-based replanning. 
Table~\ref{tab:maze_bidir_guidance} compares the default forward ReRoll matrix with a bidirectional taking over matrix under the guidance + replanning protocol. 
The result suggests that endpoint-aware revision can improve some long-horizon cases, especially AntMaze Large, while preserving competitive performance on other tasks. 
This illustrates that ReRoll schedule matrices can control information flow between causal execution and terminal constraints through selective re-noising.

\begin{table}[h!]
\centering
\scriptsize
\setlength{\tabcolsep}{5pt}
\renewcommand{\arraystretch}{1.05}
\resizebox{0.65\linewidth}{!}{
\begin{tabular}{lcc}
\toprule
Task & Forward matrix & Bidirectional taking over \\
\midrule
AntMaze Giant  
& 86.4\sdev{1.4} 
& \textbf{87.2} \\

AntMaze Large  
& 69.6\sdev{1.4} 
& \textbf{77.6} \\

AntMaze Medium 
& \textbf{98.0}\sdev{0.4} 
& 97.6 \\

PointMaze Giant  
& 99.3\sdev{0.8} 
& \textbf{100.0} \\

PointMaze Large  
& \textbf{76.3}\sdev{1.8} 
& 74.4 \\

PointMaze Medium 
& 98.8\sdev{0.4} 
& \textbf{100.0} \\
\bottomrule
\end{tabular}
}
\caption{
Forward and bidirectional ReRoll schedules under guidance-based replanning.
Forward-matrix results are mean $\pm$ standard deviation over three training seeds.
Bidirectional taking over results are single-seed and intended as exploratory evidence of schedule-matrix flexibility.
}
\label{tab:maze_bidir_guidance}
\end{table}

\subsubsection{Applications to Unified Video-Action Modeling}
\label{subsubsec:UWMReverse}

We also compare bidirectional schedule matrices in Unified Video-Action Modeling, where action tokens and future-image tokens are denoised in a joint sequence. 
Unlike maze planning, the terminal side contains future visual tokens, which can provide useful context for action generation. 
We find that the \emph{meeting matrix} consistently outperforms the \emph{taking over matrix}, suggesting that preserving bidirectional interaction between action and image tokens is beneficial for video-action modeling. 
The taking over matrix can let action-side context dominate later image generation, which may reduce the benefit of future visual context.

We place the meeting point at $0.765$ of the joint action-image horizon rather than at the midpoint. 
This biases the schedule toward image-to-action revision while limiting unnecessary non-causal action-to-action influence. 

\begin{figure}[h!]
    \centering
    \includegraphics[width=0.9\linewidth]{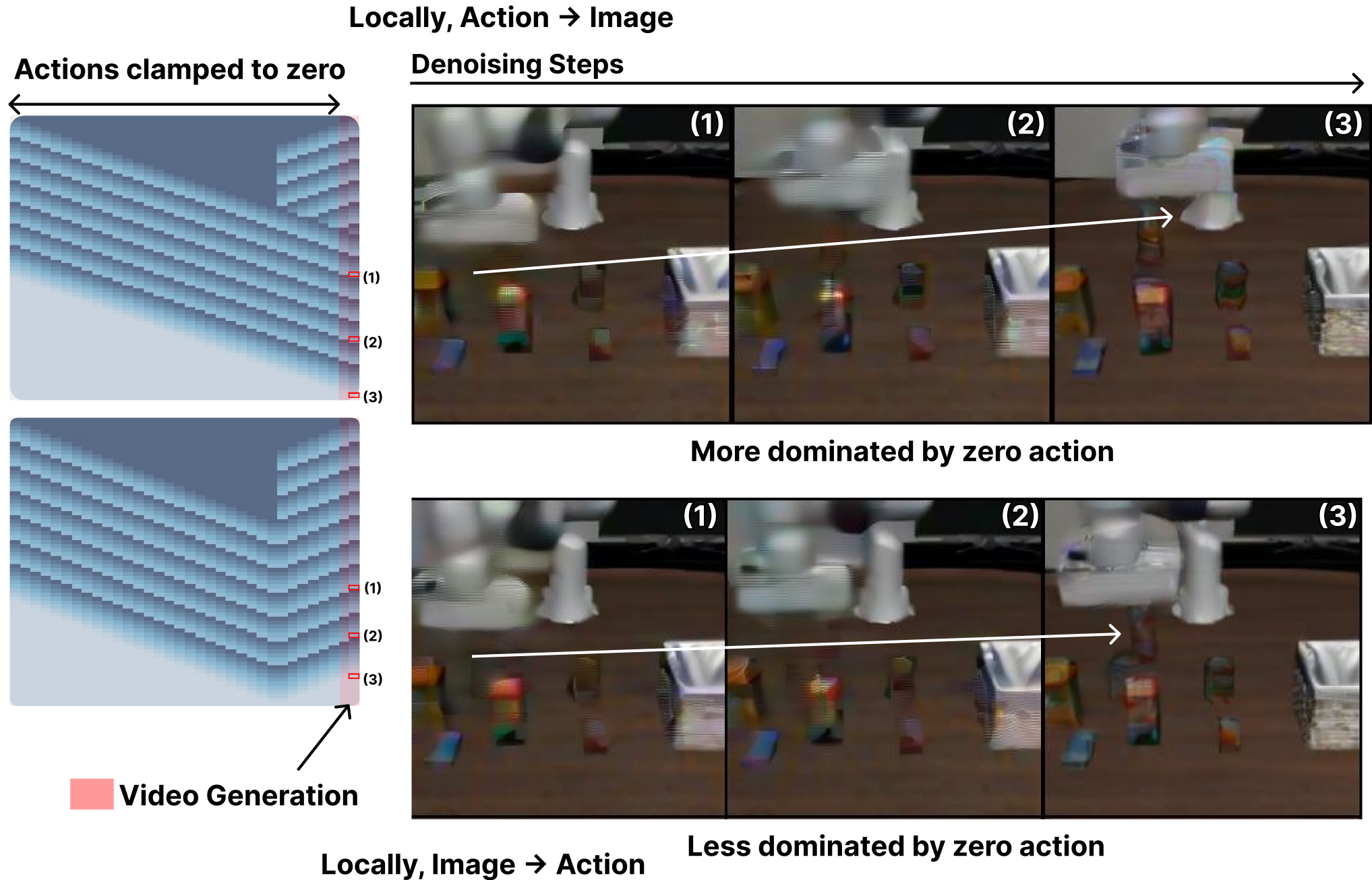}
    \caption{
    Zero-valued action-token probe of bidirectional schedules.
    Action-token values are clamped to zero while their noise-level
    labels follow the schedule. The taking over matrix is more strongly
    influenced by the clamped values than the meeting matrix.
    This probe differs from the quantitative zero-action intervention.
    Predicted clean latents $\hat{x}^{0}$ are visualized.
    }
    \label{fig:zero_actionvisualization}
\end{figure}

Fig.~\ref{fig:zero_actionvisualization} further visualizes how the two matrices induce different cross-modal information flow. 
This diagnostic is separate from the quantitative zero-action intervention in the main experiments. 
Here, we clamp the action-token values to zero but keep their noise levels following the schedule matrix, and then generate the future image tokens. 
With the taking over matrix, later denoising is more locally action-to-image, so the generated video is strongly pulled toward the zero-action behavior. 
With the meeting matrix, image-side context remains active through bidirectional revision, making the generated future images less dominated by the zero-action condition. 
This illustrates that ReRoll schedule matrices can control whether information flows primarily from actions to images or from images back to actions during denoising.

\subsection{ReRoll Schedule Ablations}
\label{subsec:physical}

The denoising slope controls the temporal span and propagation speed of each denoising wave. 
The reset noise level determines when a wave is restarted. 
The number of ReRoll events determines how broadly repeated revision is distributed across the horizon. 
Although the main experiments use nearly shared default settings for controlled comparison, the appropriate schedule can depend on horizon length, temporal information density, desired chunk granularity, and inference budget. 
The following experiments illustrate how these choices affect performance and guide schedule selection.

\subsubsection{Same ReRoll Parameters at Different Horizons}
\label{subsubsec:Horiozons}

\begin{wrapfigure}{r}{0.45\textwidth}
  \vspace{-1.0em}
  \centering
  \includegraphics[width=0.44\textwidth]{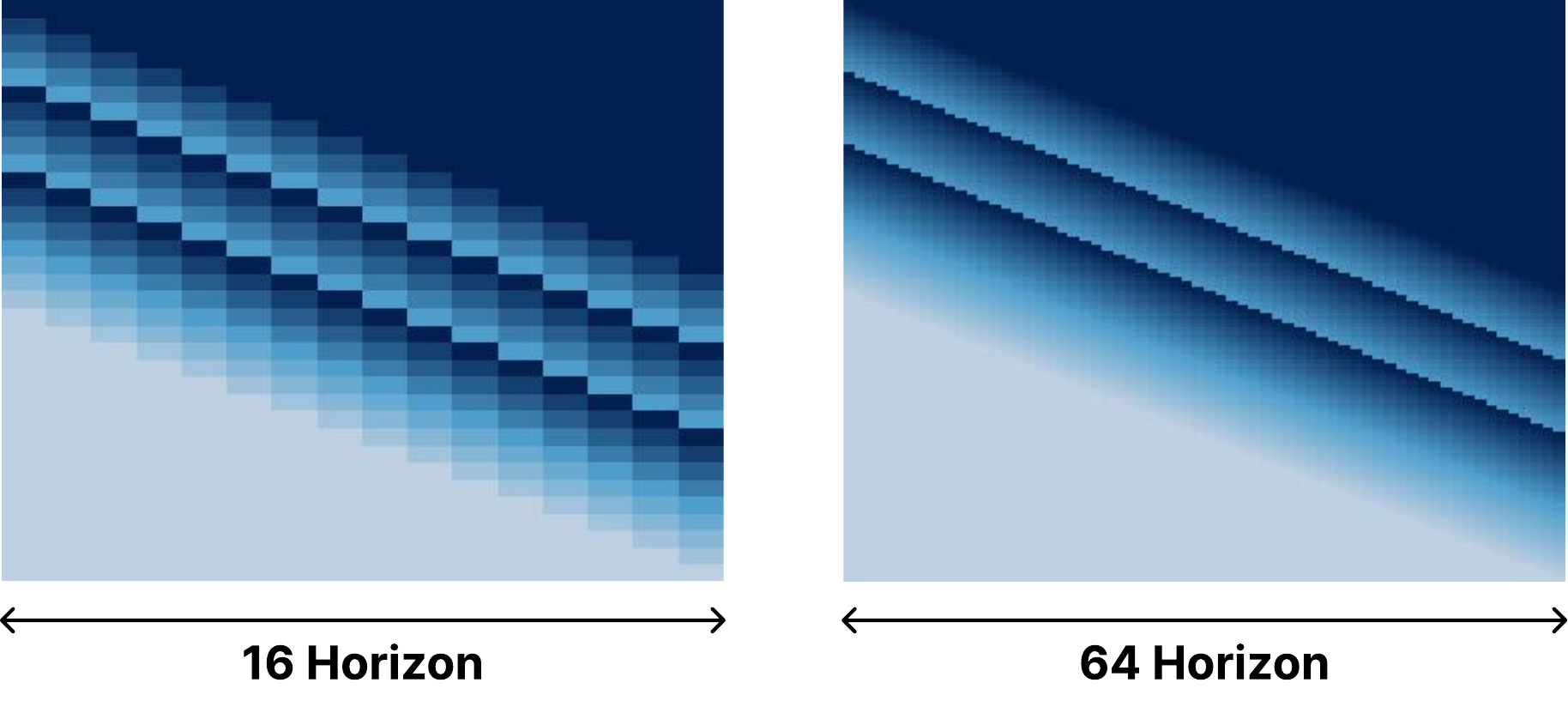}
  \caption{
Schedule matrices with same ReRoll parameters under different horizons.
  }

  \vspace{-1.0em}
  \label{fig:DifferentHorizon}
\end{wrapfigure}

The ReRoll schedule matrix is controlled by three coupled parameters: denoising slope, reset noise level, and number of ReRoll events. 
Because the schedule matrix is defined over both sequence timesteps and schedule rows, these parameters have direct meaning with respect to the prediction horizon. 
The same normalized parameter values at different horizon lengths can induce similar revision behavior across tasks, but the actual number of sampler updates and the available noise-level resolution depend on the horizon length.

For example, with a short horizon, the reset noise level can only be placed at coarse discrete positions in the matrix. 
With horizon $T=16$ and denoising slope $s_{\mathrm{dns}}=4$, only a small set of normalized reset levels, such as $0.75$, $0.5$, $0.25$, and $0$, can be represented. 
Longer horizons provide finer resolution and allow the same parameterization to realize smoother schedule variations.

\subsubsection{Effect of Reset Noise Level}
\label{subsubsec:LevelAndNumber}

The reset noise level and number of ReRoll events jointly control
revision in the schedule matrix.
The reset level determines when an active denoising wave restarts:
high levels reset regions before sufficient denoising, while low levels
allow predictions to become nearly deterministic before revision.

  \vspace{-0.7em}

\begin{figure}[h!]
    \centering
    \includegraphics[width=0.99\linewidth]{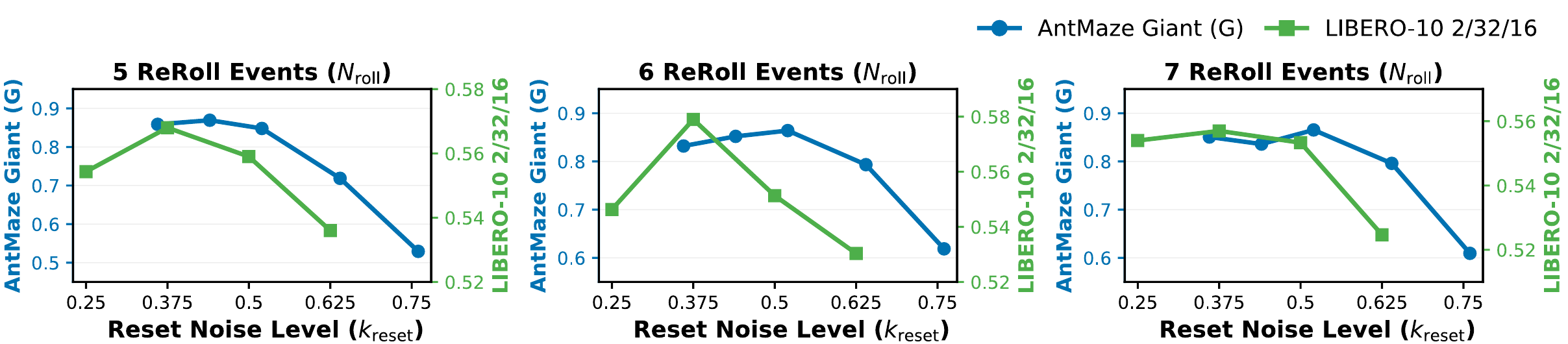}
    \caption{
    Reset noise-level ablation across ReRoll counts on AntMaze Giant
    guidance planning and LIBERO-10 Official 2/32/16.
    Intermediate reset levels generally perform best.
    }
    \label{fig:liberoreset}
\end{figure}

For the maze results in Figs.~\ref{fig:liberoreset}
and~\ref{fig:ReRollEvents}, we average the best checkpoint from each
of three training seeds.
For LIBERO, we first average each seed's best checkpoint and its
immediately preceding and following checkpoints, then average across
the three seeds.

Across ReRoll counts, both tasks generally perform best at intermediate
reset levels.
The preferred level differs across tasks: LIBERO-10 can benefit from
levels below the default $k_{\mathrm{reset}}=0.5K$, whereas AntMaze Giant
performs better at $k_{\mathrm{reset}}=0.52K$.
These differences may reflect task horizon, temporal density, and the
underlying noise schedule described in
Appendix~\ref{sec:ExperimentSettings}.

This trade-off can be interpreted as balancing confidence in
cross-horizon predictions. Low reset levels may encourage overcommitment,
while high levels leave insufficiently denoised context for other regions.
Estimating region-wise confidence to adaptively select reset levels or
trigger ReRoll events is a direction for future work.

\subsubsection{Results with Lower Denoising Slopes}
\label{subsubsec:TooShort}
\begin{table}[h!]
\centering

\scriptsize
\setlength{\tabcolsep}{3.2pt}
\renewcommand{\arraystretch}{1.08}
\resizebox{\linewidth}{!}{
\begin{tabular}{lccccccc cc}
\toprule
\multirow{2}{*}{Task}
& \multicolumn{3}{c}{Schedule matrix}
& \multicolumn{3}{c}{Training chunks}
& \multirow{2}{*}{Setting}
& \multicolumn{2}{c}{Success rate} \\
\cmidrule(lr){2-4}
\cmidrule(lr){5-7}
\cmidrule(lr){9-10}
& $s_{\mathrm{dns}}$
& $k_{\mathrm{reset}}/K$
& $N_{\mathrm{roll}}$
& Slope range
& Start range
& $N_{\max}$
&
& Main / Official
& Open \\
\midrule

\multicolumn{10}{l}{{Default schedule used in main experiments}} \\
\midrule

AntMaze Large
& 4 & 0.52 & 6
& $[2,10]$
& $[0,0.5K]$
& 8
& Guidance
& 69.6\sdev{1.4}
& -- \\

AntMaze Giant
& 4 & 0.52 & 6
& $[2,10]$
& $[0,0.5K]$
& 8
& Guidance
& 86.4\sdev{1.4}
& -- \\

LIBERO-10
& 4 & 0.50 & 6
& $[2,10]$
& $[0,0.5K]$
& 8
& 2/16/8
& 51.0\sdev{24.1}
& 82.5\sdev{3.6} \\

\midrule
\multicolumn{10}{l}{{Lower-slope example configurations}} \\
\midrule

AntMaze Large
& 1 & 0.52 & 2
& $[0.5,2]$
& $[0,0.5K]$
& 4
& Guidance
& 66.4
& -- \\

AntMaze Large
& 2 & 0.52 & 3
& $[1,4]$
& $[0,0.5K]$
& 6
& Guidance
& 69.2
& -- \\

AntMaze Giant
& 2 & 0.52 & 3
& $[1,4]$
& $[0,0.5K]$
& 6
& Guidance
& 74.0
& -- \\

LIBERO-10
& 2 & 0.50 & 2
& $[1,4.5]$
& $[0,0.5K]$
& 4
& 2/16/8
& 47.2\sdev{13.2}
& 77.0\sdev{2.8} \\

\bottomrule
\end{tabular}
}
\caption{
Example ReRoll schedule and linear-chunk training configurations.
Default settings are used in the main experiments, while lower-slope settings show that ReRoll can also be trained and deployed with different schedule-matrix parameters.
Success rates are reported in percent.
}
\label{tab:reroll_param_examples}
\end{table}

Several ReRoll training and schedule-matrix parameters are tied to the denoising slope $s_{\mathrm{dns}}$. 
While the main experiments mostly use $s_{\mathrm{dns}}=4$ or nearby values, lower slopes can also work well when the training chunk distribution is adjusted accordingly. 
For common prediction horizons used in our experiments, we recommend starting with $s_{\mathrm{dns}}\in[2,4]$ and then adjusting the chunk slope range, reset level, and number of ReRoll events according to the task horizon and inference budget.
We include representative lower-slope configurations in Table~\ref{tab:reroll_param_examples}. 
These results are not intended as a direct comparison, but they show that ReRoll can be trained and deployed with different schedule-matrix parameterizations across tasks.

\subsection{Sampler Update Counts and Inference Time}
\label{subsec:InferenceTime}

The inference cost of DR is determined by the schedule matrix.
Each row specifies token-wise noise levels, and one sampler update occurs
between consecutive rows.
Thus, the update count equals the number of schedule rows traversed minus one; policy sampling may stop once the executable prefix is clean.
Unlike full-sequence diffusion, where the sampling-step budget is usually
chosen directly, DR's update count is induced by the prediction horizon and schedule parameters, including denoising slope, reset noise level, and number of ReRoll events.

Longer prediction horizons generally require more schedule rows to preserve the same denoising-wave shape. 
Therefore, when the prediction horizon is a design choice, it should be selected together with the inference-time budget. 
After training with compatible linear-chunk noise patterns, the simplest way to reduce inference cost is often to use fewer ReRoll events at deployment. 
For example, in the policy-learning ablation in Fig.~\ref{fig:ReRollEvents}, performance already saturates after a moderate number of ReRoll events, so a smaller-event schedule can be used when faster inference is preferred. 
The reset noise level can also change the number of schedule rows, but we mainly report the effect of changing ReRoll events. For visualization of these matrices, see Appendix~\ref{sec:ExperimentSettings}.

\begin{table}[h!]
\centering
\tiny
\setlength{\tabcolsep}{3.0pt}
\renewcommand{\arraystretch}{1.02}
\resizebox{0.64\linewidth}{!}{
\begin{tabular}{lccccc}
\toprule
\multirow{2}{*}{$N_{\mathrm{roll}}$}
& \multirow{2}{*}{Planning}
& \multicolumn{3}{c}{Policy learning}
& \multirow{2}{*}{Unified Video-Action} \\
\cmidrule(lr){3-5}
& & 16/8 & 32/16 & 64/16 & \\
\midrule
0  & 124 & 11 & 23 & 31 & 34 \\
1 & 137 & 14 & 28 & 40 & 40 \\
2 & 150 & 17 & 33 & 49 & 46  \\
3 & 163 & 20 & 38 & 58 & 52 \\
4 & 176 & 23 & 43 & 67 & 58 \\
5 & 189 & 26 & 48 & 76 & 64 \\
6 & 202 & 29 & 53 & 85 & 70 \\
7 & 215 & 32 & 58 & 94 & 76  \\
8 & 228 & 35 & 63 & 103 & 82  \\
\bottomrule
\end{tabular}
}
\caption{
Sampler update counts induced by ReRoll schedule matrices.
For video-action generation, $N_{\mathrm{roll}}$ denotes the forward-side ReRoll count.
16/8 denotes prediction horizon/action steps.
}
\label{tab:ddim_steps_all}
\end{table}

These counts show that schedule design directly affects inference cost. 
Increasing ReRoll events generally improves revisability but adds DDIM updates, while fewer events can reduce latency when performance has already saturated. 
In practice, we choose the schedule by considering both task performance and the available inference budget.

\paragraph{Policy inference time.}
We additionally measure real full-policy inference latency on a single NVIDIA A5000 GPU with batch size 1. 
Each value is mean $\pm$ standard deviation over 30 timed iterations after 10 warmup iterations. 
The timed region includes image encoding and action generation, but excludes environment stepping, media writing, and final action unnormalization. 
Table~\ref{tab:policy_inference_compact} compares standard Diffusion Policy step budgets with DR schedules using different ReRoll counts.

\begin{table}[h!]
\centering

\tiny
\setlength{\tabcolsep}{3.2pt}
\renewcommand{\arraystretch}{1.02}
\resizebox{0.75\linewidth}{!}{
\begin{tabular}{lccc}
\toprule
Sampler 
& O2/H16/A8 
& O2/H32/A16 
& O2/H64/A16 \\
\midrule

DP, 10 DDIM
& 10 / 44.7\sdev{1.7}
& 10 / 44.9\sdev{1.6}
& 10 / 46.5\sdev{3.7} \\

DP, 30 DDIM
& 30 / 121.6\sdev{4.9}
& 30 / 123.6\sdev{2.7}
& 30 / 121.1\sdev{4.4} \\

DP, 50 DDIM
& 50 / 207.4\sdev{12.6}
& 50 / 203.9\sdev{9.0}
& 50 / 194.8\sdev{3.8} \\

DP, 100 DDPM
& 100 / 390.7\sdev{7.7}
& 100 / 421.1\sdev{34.6}
& 100 / 387.4\sdev{10.9} \\

\midrule

DR, 3 ReRoll
& 20 / 82.8\sdev{5.4}
& 38 / 155.9\sdev{7.3}
& 58 / 229.8\sdev{5.6} \\

DR, 6 ReRoll
& 29 / 123.1\sdev{2.4}
& 53 / 225.2\sdev{4.7}
& 85 / 346.3\sdev{4.6} \\

\bottomrule
\end{tabular}
}
\caption{
Policy inference latency on a single RTX A5000 GPU.
Each cell reports sampler updates / latency in milliseconds.
}
\label{tab:policy_inference_compact}
\vspace{0.1em}

\end{table}

\subsection{Inference Time Optimization}
\label{sec:InferenceOptimization}

Longer horizons and more ReRoll events increase the number of schedule
rows and sampler updates.
For closed-loop deployment, however, not every update needs to block
action execution.
We therefore use two training-free inference optimizations:
\emph{sliding-window reuse} and \emph{causal pipelining}.
Here, blocking updates denote denoising updates that must finish before
the next executable action becomes available.

\paragraph{Sliding-window reuse.}
Under receding-horizon control, only a prefix of each predicted action
sequence is executed.
Rather than discarding the remaining suffix, we convert the predictions
back to physical per-step actions, retain the unexecuted actions, and
re-encode them using the cumulative representation and horizon-dependent
normalization at their new window indices.
This avoids directly shifting normalized values whose scaling depends
on their original horizon positions.

\begin{figure}[h!]
    \centering
    \includegraphics[width=0.95\linewidth]{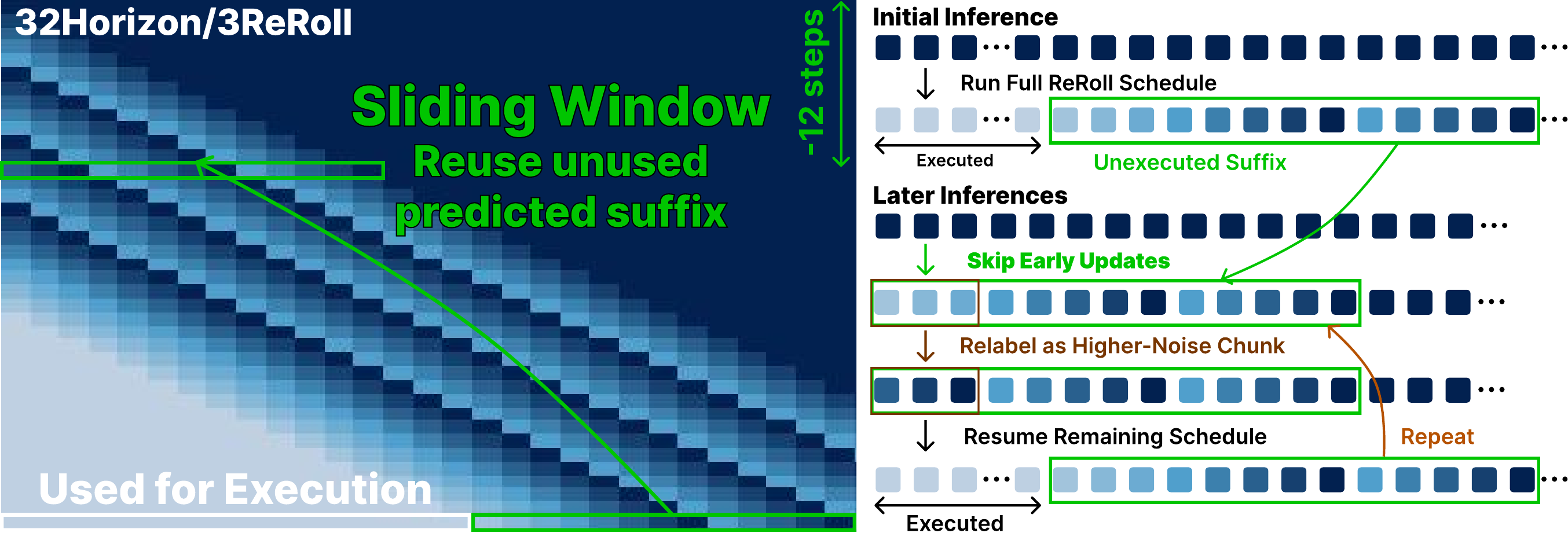}
    \caption{
    Sliding-window reuse.
    The unexecuted suffix from the previous prediction is shifted into the
    next window and reused with reassigned higher noise levels, allowing
    early denoising updates to be skipped.
    }
    \label{fig:sliding_window}
\end{figure}

When the resumed schedule requires higher noise levels than the reused
tokens previously carried, we assign the higher noise-level labels
without adding noise to the transformed values.
This relabeling is an inference heuristic rather than a forward-noising
transition.
The reused suffix therefore warm-starts the shifted schedule, allowing
early updates to be skipped.
We repeat this procedure at subsequent replanning steps without additional
training; the first prediction uses the regular schedule because no
previous suffix is available.

\paragraph{Causal pipelining.}
The final part of the ReRoll schedule is causal, so later actions need not
be fully resolved before the next action is executed.
We first run the schedule until the next action reaches zero noise, then
release that action and continue denoising later tokens while it is being
executed.
In our LIBERO setup, one action lasts 50\,ms while one denoising update takes
approximately 5\,ms, allowing several subsequent updates to overlap with
execution.
Thus, causal pipelining reduces blocking latency without reducing the
total denoising computation.

\begin{figure}[h!]
    \centering
    \includegraphics[width=0.89\linewidth]{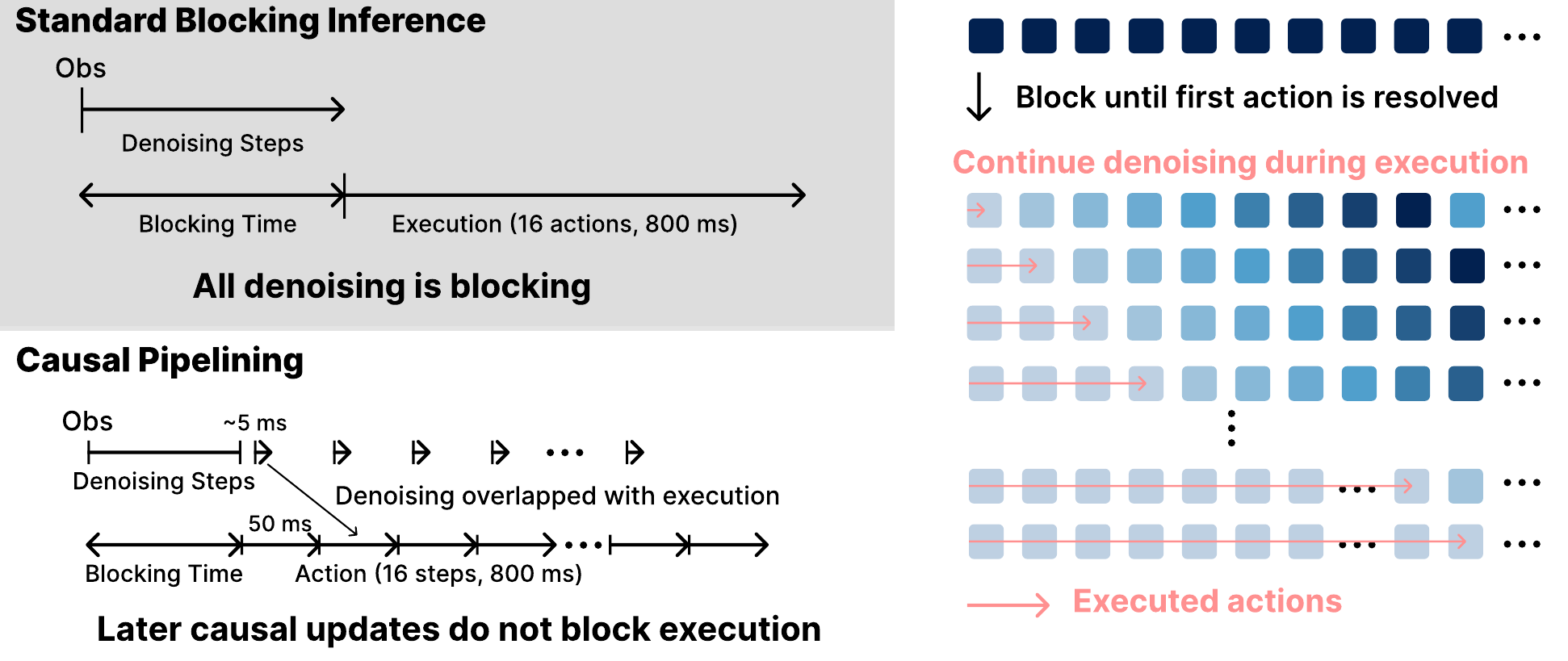}
    \caption{
  Causal pipelining.
  Standard inference completes denoising before execution, whereas
  pipelining blocks only until the next action is resolved and overlaps
  later denoising updates with action execution.
    }
  \label{fig:causalPipe}
\end{figure}


\paragraph{Latency reduction.}
Fig.~\ref{fig:inference_latency} shows the effect of these optimizations on
LIBERO-10 Official with \textsc{Obs./Hor./Act.}=2/32/16.
Starting from the six-ReRoll schedule with 53 blocking updates, causal
pipelining, reducing to three ReRoll events, and sliding-window reuse reduce
the blocking updates to 38, 23, and 11, respectively, while preserving similar
mean success.
For DF, the lower-latency configuration simply doubles the denoising slope,
reducing the number of blocking updates from 32 to 16.

The same optimizations scale to longer horizons.
For the longest \textsc{2/64/16} setting with six ReRoll events, causal
pipelining reduces 85 blocking updates to 70, and sliding-window reuse further
reduces them to 34, without changing the number of ReRoll events.
Reducing $N_{\mathrm{roll}}$ can further lower latency, but may degrade
performance when additional revision is useful.
In our LIBERO policy experiments, performance largely saturates by three
ReRoll events (Fig.~\ref{fig:ReRollEvents}), making later events mostly
redundant, whereas other tasks may benefit from more revision.

\section{Training Details for Diffusion ReRoll}
\label{sec:TrainingDetails}

\subsection{Noise Distribution for Training }
\label{subsec:NoiseDistribution}

Unlike Diffusion Forcing, which samples token-wise noise levels independently~\cite{diffusion_forcing}, Diffusion ReRoll trains with structured chunk-wise linear noise profiles. 
The goal is to expose the model to a broad range of noise levels while preserving the local linear patterns used by the deployment schedule matrix. For the ablations in this section, we mainly use the LIBERO-10 official protocol and AntMaze Giant guidance-based replanning.





\subsubsection{First-Chunk Noise Assignment}
\label{subsubsec:FirstChunk}

Most chunks are sampled to mimic denoising waves that appear after a ReRoll event. 
Their starting offsets are therefore sampled from the reset-level range. 
The first chunk is different because it has no preceding ReRoll event. 
It must cover both the initial high-noise wave and later deployment states where the beginning of the horizon has already been denoised.

\begin{figure}[h!]
    \centering
    \includegraphics[width=0.7\linewidth]{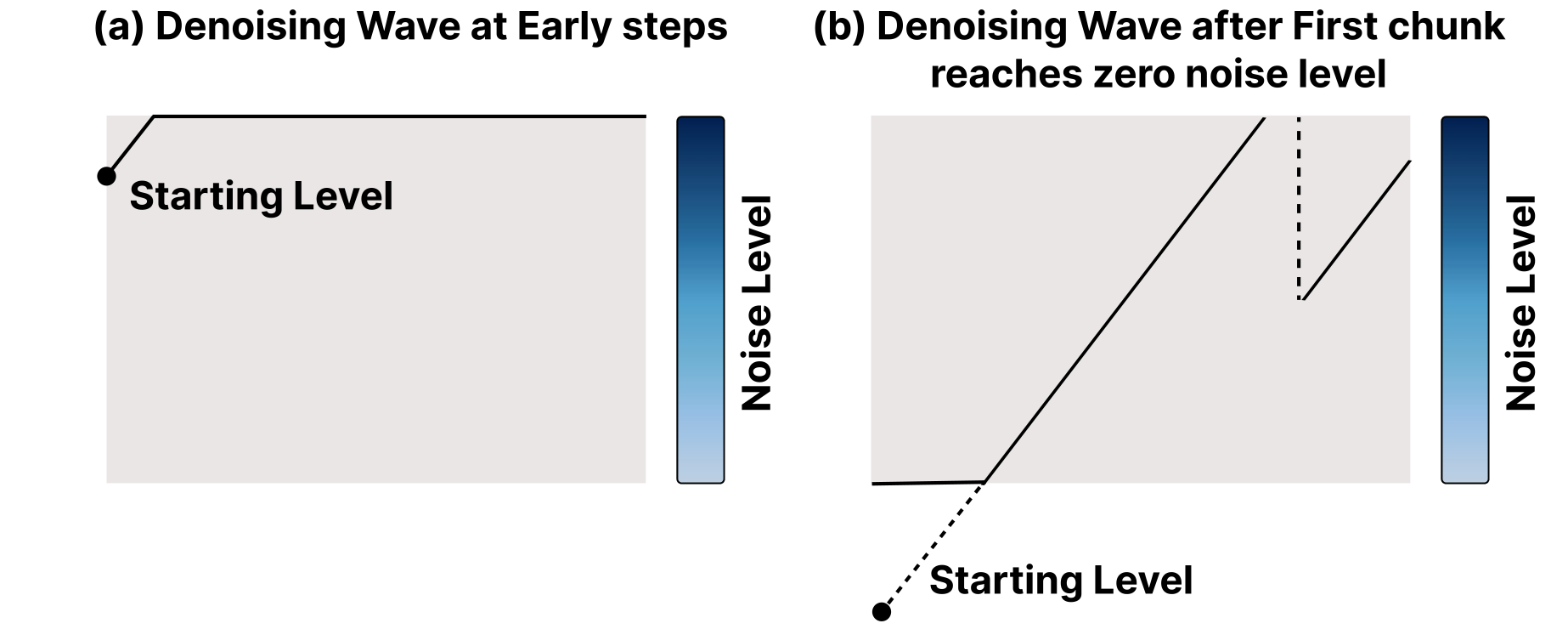}
    \caption{
    First-chunk noise assignment.
    Later chunks start from the reset-level range, while the first chunk uses a wider starting-offset range to cover both high-noise initial states and clean-prefix states.
    }
    \label{fig:firstchunk}
\end{figure}

Let $K$ be the maximum noise level and let $j$ denote the local token index within a chunk. 
Given chunk slope $\alpha_c$ and starting offset $s_c$, token-wise noise levels are assigned by
\begin{equation}
    k_j =
    \operatorname{clip}_{[0,K]}
    \left(
        \left\lfloor s_c + \alpha_c j \right\rfloor
    \right).
\end{equation}
For later chunks, we sample $s_c$ from the reset-level range, e.g., $s_c\sim \mathcal{U}(s_{\min},s_{\max})$. 
For the first chunk, we instead use a wider range:
\begin{equation}
    s_0 \sim \mathcal{U}(s_{\mathrm{first}}^{\min}, K),
    \qquad
    s_{\mathrm{first}}^{\min}
    =
    \max
    \left(
        -K,\;
        \left\lfloor -\frac{T}{N_{\mathrm{chunk}}}\alpha_{\min} \right\rfloor
    \right).
\end{equation}
where $T$ is the prediction horizon, $N_{\mathrm{chunk}}$ is the sampled number of chunks, and $\alpha_{\min}$ is the minimum sampled chunk slope.

Here, $s_0$ is the starting offset of the first linear profile before clipping. 
When $s_0<0$, clipping produces a clean prefix with noise level $0$, matching deployment states where the beginning of the horizon is already denoised. 
When $s_0$ is large, the first chunk starts in a high-noise regime, matching the initial denoising wave. 
In bidirectional training, the same rule is applied symmetrically to the last chunk when it is reversed, since it plays the role of the first denoising wave from the terminal side. 
This special first-chunk sampling aligns training with the range of first-wave states encountered by the deployment schedule matrix.

\subsubsection{Bidirectional Mode Noise Assignment}
\label{subsubsec:BidirNoiseAss}

Bidirectional training uses the same randomized linear-chunk sampler as forward training, but allows selected chunks on the right side of the meeting region to be reversed. 
For a token with local index $j$ in a chunk of length $\ell_c$, a reversed chunk uses
\begin{equation}
    j_{\mathrm{rev}} = \ell_c - 1 - j .
\end{equation}
Thus, the same linear noise rule can represent either a left-to-right or right-to-left denoising wave.

We empirically find that training only with bidirectional chunks is less effective. 
Instead, we mix forward and bidirectional noise patterns.
Each training sample is drawn from one of three branches: (i) forward branch, (ii) masked bidirectional branch, or (iii) full-horizon bidirectional branch.

In the \emph{forward branch}, we use the standard forward training pattern: a random suffix is assigned full noise, and no chunks are reversed.
Although bidirectional deployment uses waves from both ends, this branch remains useful because forward local revision still appears in left-origin waves and taking over schedules.

In the \emph{masked bidirectional branch}, we sample a masked interval around the meeting region and assign full noise to tokens inside it.
Chunks on the right side of this interval use the reversed coordinate $j_{\mathrm{rev}}$, exposing the model to backward revision from the terminal side.

In the \emph{full-horizon bidirectional branch}, the mask is removed, but a sampled subset of right-side chunks is still reversed.
This covers post-meeting deployment states, where the horizon is no longer separated by a masked region.

The masked bidirectional branch is sampled with probability $p_{\mathrm{bi}}$, and the full-horizon bidirectional branch is sampled with probability $p_{\mathrm{full}}$, with
\begin{equation}
p_{\mathrm{bi}} + p_{\mathrm{full}} \leq 1 .
\end{equation}
The remaining probability is assigned to the forward branch.

For tasks with inpainting-style conditioning, we additionally expose the model to near-clean conditioning regions.
Conditioned on sampling a bidirectional branch, we apply inpainting-style noise assignment with probability $p_{\mathrm{inpaint}}$.
The selected conditioning region is assigned a low noise level sampled from $[0,0.25K]$.
Thus, the overall probability of this case is $(p_{\mathrm{bi}}+p_{\mathrm{full}})p_{\mathrm{inpaint}}$. For example, if $p_{\mathrm{bi}}+p_{\mathrm{full}}=0.5$ and $p_{\mathrm{inpaint}}=0.1$, it is applied to $5\%$ of training samples.
In maze goal-inpainting, this low-noise region is assigned to the final token, which contains the goal. This has little effect on standard training behavior, but improves the stability of inpainting-style deployment. For the Diffusion Forcing variant used in inpainting, we apply the same masking mechanisms during training while omitting the linear noise-level assignment.

Overall, bidirectional training does not reverse the entire trajectory. 
It reverses only selected right-side or post-meeting chunks while preserving the randomized linear-chunk distribution used in forward training.

\subsubsection{Chunk Slope Range}
\label{subsubsec:SlopeRange}

The chunk-slope range is chosen around the denoising slope used by the deployment ReRoll schedule matrix. 
If this range is too narrow, the model mainly observes one local linear pattern and can overfit to a specific schedule. 
We therefore train with a wider slope range so that the model can handle a broader family of linear chunks, including possible schedule variations.

\begin{wrapfigure}{r}{0.45\textwidth}
  \centering
  \vspace{-1.0em}
  \includegraphics[width=0.44\textwidth]{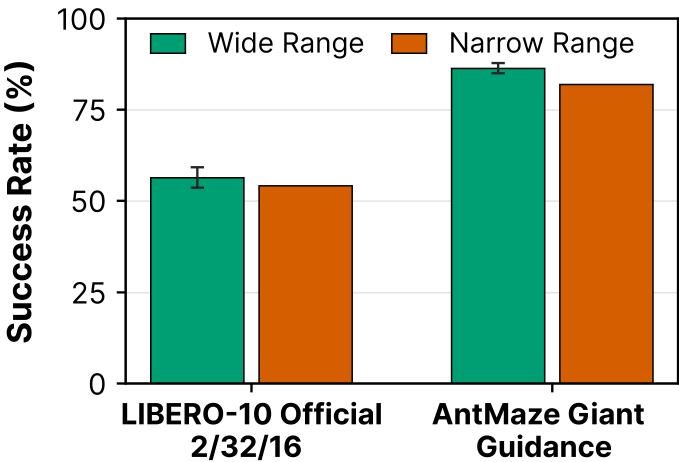}
    \caption{
    Effect of chunk-slope range.
    A wider range gives slightly better performance.
    }
    \label{fig:SlopeAblation}
      \vspace{1.5em}

\end{wrapfigure}

In the main setting, each chunk slope is sampled as
\begin{equation}
    \alpha_c \sim \mathcal{U}
    \left(
        0.5s_{\mathrm{dns}},\; 2.5s_{\mathrm{dns}}
    \right),
\end{equation}
where $s_{\mathrm{dns}}$ is the denoising slope of the deployment schedule. 
We compare this default wide range with a narrow range, 
\begin{equation}
\alpha_c \sim \mathcal{U}(0.875s_{\mathrm{dns}}, 1.125s_{\mathrm{dns}})
\end{equation}


Fig.~\ref{fig:SlopeAblation} shows a small performance decrease with the
narrower range on both LIBERO-10 and AntMaze Giant.
This suggests a modest benefit from covering diverse local linear noise
patterns rather than closely matching a single deployment schedule.

In our experiments, the deployment schedule uses a fixed denoising slope. 
However, more flexible schedules could vary the slope over denoising steps or sequence regions. For example, one could use steeper slopes at early denoising steps and shallower slopes later.
Training with a wide chunk-slope range may better support such variable-slope schedules.

\subsubsection{Starting-Level Range}
\label{subsubsec:StartingRange}

The starting-level range controls the initial noise offset of each sampled linear chunk. 
Although deployment chunks are often introduced near the reset noise level, training only around this narrow range can leave other noise levels underrepresented. 
For example, if later chunks start only from $[0.25K,0.75K]$, then noise levels below $0.25K$ are rarely assigned except through the first chunk.

We therefore sample later-chunk starting levels from a zero-to-middle noise range:
\begin{equation}
    s_c \sim \mathcal{U}(0,\; s_{\max}),
\end{equation}
where $s_{\max}$ is chosen around the middle or reset-level region of the diffusion schedule. 
This range still covers the reset-level behavior used by the deployment schedule matrix, while also exposing the model to low-noise and nearly clean chunk states.

\subsection{Important Components for Diffusion ReRoll Training}
\label{subsec:ImportantComponents}

Training Diffusion ReRoll requires more than sampling the right token-wise noise levels. 
Because ReRoll relies on cross-horizon revision, the model also needs compatible representations, masking choices, and architecture interfaces.
This section summarizes additional critical components that affect DR performance. 

\subsubsection{Low Sensitivity to Basic Learning Parameters}
\label{subsubsec:LowSense}
Diffusion ReRoll does not require a separate optimization recipe. 
Across tasks, we largely keep the learning rates, optimizers, batch sizes, and training schedules from the corresponding baseline codebases~\cite{diffusion_forcing,mctd,diffusion_policy,OAT,uwm}. 
Most changes are instead related to token-wise noise assignment, linear-chunk training, masking, and the deployment schedule matrix. 
When applying DR to new tasks, we therefore recommend first tuning ReRoll-specific schedule and noise-distribution parameters rather than basic learning hyperparameters.

\subsubsection{Necessity of Non-causal Training}
\label{subsubsec:NonCausal}

\begin{wrapfigure}{r}{0.45\textwidth}
  \vspace{-1.0em}
  \centering
  \includegraphics[width=0.44\textwidth]{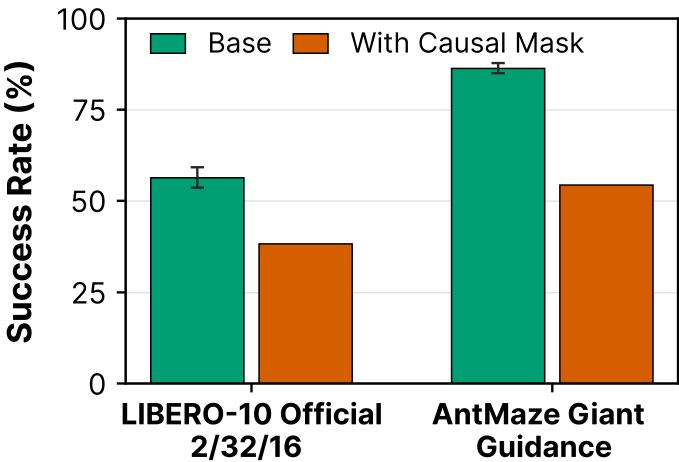}
    \caption{
    Effect of causal masking.
    Causal masking restricts cross-horizon information flow and degrades ReRoll performance.
    }
        \label{fig:CausalMask}
          \vspace{-2.0em}

\end{wrapfigure}

Many sequence models use causal masking to impose left-to-right sequential structure. 
However, DR is designed to use cross-horizon context during denoising. Selected regions are re-noised and revised again using information from other parts of the horizon. 
If causal masking is applied, this information flow is restricted, making it harder for the model to exploit ReRoll's revisable denoising process.

Fig.~\ref{fig:CausalMask} shows that adding a causal mask degrades performance on both LIBERO-10 and AntMaze Giant. 
This supports that DR's gains depend on non-causal cross-horizon revision, rather than only on token-wise noise conditioning.


\subsubsection{SNR Reweighting}
\label{subsubsec:SNR}

Because ReRoll trains with heterogeneous token-wise noise levels, different tokens can have very different signal-to-noise ratios (SNRs). 
We do not introduce a new loss weighting rule. Instead, we adopt the Min-SNR weighting strategy~\cite{minSNR}, following the Diffusion Forcing implementation~\cite{diffusion_forcing}. 
For a token at noise level $k$, we define
\begin{equation}
    \mathrm{SNR}(k)
    =
    \frac{\bar{\alpha}_k}{1-\bar{\alpha}_k}.
\end{equation}

All DR and DF experiments use $x_0$-prediction. 
Under this parameterization, we use the clipped SNR weight
\begin{equation}
    w_t
    =
    \min\left(\mathrm{SNR}(k_t), \gamma\right),
\end{equation}
where $k_t$ is the noise level assigned to token $t$ and $\gamma$ is the Min-SNR clipping threshold. 
The resulting weighted objective is
\begin{equation}
    \mathcal{L}
    =
    \mathbb{E}
    \left[
        \sum_{t=1}^{T}
        w_t
        \left\|
            x_t^0 - \hat{x}_{\theta,t}^{0}
        \right\|^2
    \right].
\end{equation}

In our main experiments, we set $\gamma=2.5$ for both Diffusion Forcing and Diffusion ReRoll, except for the video-generation part of unified video-action generation with larger $\gamma=5.0$. 
We found this relatively low clipping threshold to work robustly across tasks, but we did not exhaustively tune SNR weighting. 
Designing loss weights specifically for structured ReRoll schedule matrices may further improve performance.

\subsection{Action Representation for Manipulation}
\label{subsec:action_representation}

For manipulation tasks, we predict action chunks as in prior
policies~\cite{UMI,pi0}, using an anchor-relative cumulative
representation for DF and DR.
Each token represents cumulative position and rotation commands
anchored at the start of the action chunk.
Unlike accumulating predicted per-step deltas, this makes each token's
cumulative target explicit, so revising an early token does not
automatically shift later targets.
This is useful for ReRoll, which revisits tokens at different horizon
indices during denoising.

Let $a_i=(\Delta p_i,\Delta r_i,g_i)$ denote the position, axis-angle
rotation, and gripper commands at step $i$.
We accumulate normalized position commands and compose rotations:
\begin{equation}
    p_i^c = \sum_{k=1}^{i}\mathcal{N}_p(\Delta p_k),
    \qquad
    R_i^c = R_{i-1}^c\,\operatorname{Rot}(\Delta r_i),
    \qquad R_0^c=I,
\end{equation}
where $\mathcal{N}_p$ is the standard per-step position-action
normalizer and $\operatorname{Rot}$ converts an axis-angle vector
into a rotation matrix.
Thus, $p_i^c$ accumulates normalized action coordinates, not metric
displacements.
Rotation commands are composed before normalization rather than
accumulated by summing axis-angle vectors.

\paragraph{Horizon-dependent scaling.}
Later cumulative targets typically have larger magnitudes.
To compensate, we use the following position and rotation targets
for diffusion:
\begin{equation}
    z_i^p = \frac{p_i^c}{i^{\alpha_p}},
    \qquad
    z_i^r =
    \mathcal{N}_r\!\left(
        \frac{\operatorname{rotvec}(R_i^c)}
             {c_r i^{\alpha_r}}
    \right),
\end{equation}
where $\operatorname{rotvec}$ converts a rotation matrix to its
axis-angle vector, and $\mathcal{N}_r$ is a per-channel affine
normalizer fitted to the rescaled cumulative rotations.
The scaling parameters are fitted once from horizon-wise standard
deviations in the training demonstrations.
Position uses a power law with unit prefactor, while rotation also
includes the fitted scale factor $c_r$ because its commands are
composed before normalization.
We apply no additional cumulative-position normalizer.
Gripper commands use standard per-step normalization and are not
accumulated.

\paragraph{Conversion to executable actions.}
At inference, we undo position scaling and invert rotation
normalization and scaling to reconstruct $p_i^c$ and $R_i^c$.
We then recover per-step commands as
\begin{equation}
    \Delta p_i =
    \mathcal{N}_p^{-1}\!\left(p_i^c-p_{i-1}^c\right),
    \qquad
    \Delta r_i =
    \operatorname{rotvec}\!\left((R_{i-1}^c)^{-1}R_i^c\right),
\end{equation}
with $p_0^c=0$ and $R_0^c=I$.
Together with the gripper commands restored from their normalization,
these form the executable action sequence.

\paragraph{Diffusion Policy with the same representation.}

\begin{wrapfigure}{r}{0.44\textwidth}
  \vspace{-0.5em}       

  \centering
  \includegraphics[width=0.41\textwidth]{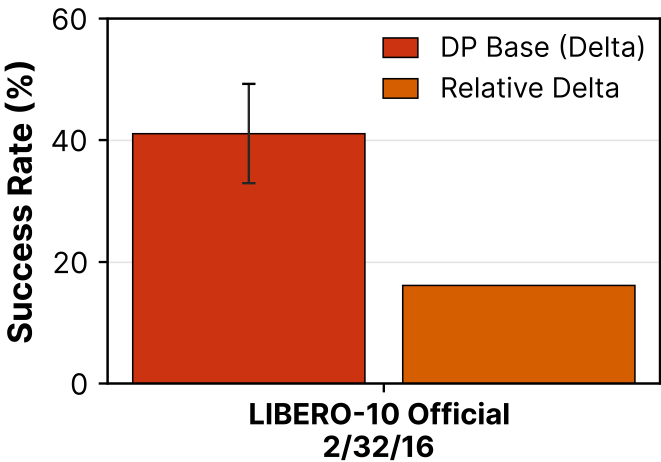}
    \caption{
    Effect of action representation on Diffusion Policy.
    The original delta-action representation performs better than the relative cumulative representation used by DF/DR.
    }
       \label{fig:Representation}
  \vspace{-1.0em}       
\end{wrapfigure}

To check whether DR's gains come mainly from the action representation, we train an additional Diffusion Policy baseline using the same representation used by DF and DR. 
This ablation separates the effect of the action representation from the effect of ReRoll's schedule matrix and linear-chunk training.

As shown in Fig.~\ref{fig:Representation}, applying the relative cumulative representation to vanilla Diffusion Policy does not improve performance; it substantially reduces success compared with the original delta-action representation. 
Thus, the gains of DR are not explained by the action representation alone. 
Since DF and DR use the same action representation, the performance gap between them is more directly attributed to ReRoll's structured schedule matrix and linear-chunk training.

\subsection{Model Architecture}
\label{subsubsec:model_architecture}

\paragraph{Long-horizon planning.}
For OGBench maze planning, Diffuser uses its U-Net trajectory denoiser~\cite{diffuser}, while Diffusion Forcing (DF) and Diffusion ReRoll (DR) share the Transformer sequence denoiser from Diffusion Forcing~\cite{diffusion_forcing}.
ReRoll modifies the training noise distribution and deployment schedule, not the backbone architecture.

\paragraph{Policy Learning.}
For manipulation policy learning, the full-sequence baseline uses the Diffusion Policy Chi-Transformer~\cite{diffusion_policy}, which conditions on a global diffusion timestep token alongside observation tokens and uses learnable action positional embeddings.

For DF and DR, we modify this backbone to support per-token noise levels.
We replace the global timestep token with a sinusoidal noise-level embedding $\phi(k_i)$, concatenated with the noisy action token $\hat{a}_i$ at horizon index $i$ before projection:
\begin{equation}
    z_i
    =
    W_2 \sigma\!\left(
    W_1 [\hat{a}_i;\phi(k_i)]
    \right).
\end{equation}
This allows the denoiser to operate under arbitrary token-wise schedules, including ReRoll.

We also replace learnable action positional embeddings with fixed sinusoidal embeddings indexed by horizon position, avoiding horizon-specific learned parameters.
The observation encoder, observation-side positional embeddings, conditioning encoder, decoder architecture, and output head are retained; attention masking and dropout follow Table~\ref{tab:hparams_policy}.
Thus, DF and DR share the modified architecture and differ only in their training noise distributions and deployment schedules.

This modification adds few parameters.
For $T=16$, the Diffusion Policy denoiser contains $8{,}997{,}895$ parameters, compared with $9{,}096{,}199$ for DF/DR.
The increase of $98{,}304$ parameters ($1.1\%$) comes from the action-token input interface.

\paragraph{Unified Video-Action Modeling.}
For video-action modeling, the full-sequence baseline UWM~\cite{uwm} uses a DiT-style Transformer to denoise action tokens and future image latents.
UWM uses modality-level timestep conditioning through AdaLN, with one global timestep for actions and another for future images.

For UWM + DF and UWM + DR, we use the same per-token-noise-conditioned architecture.
We define a joint action-image horizon with separate noise levels for each action token and future latent frame.
The timestep embedding is removed from AdaLN, leaving it conditioned only on the encoded current observation.
Instead, sinusoidal noise-level embeddings are concatenated with each noisy action vector before the action encoder or with patch-embedded image features before projection to the Transformer width.

We also replace UWM's learnable joint positional embeddings with fixed sinusoidal embeddings over a continuous action-image axis.
Image-token positions follow action-token positions without restarting, while register tokens retain learnable positional embeddings.

Finally, temporal patch stride one allows each future latent frame to contribute its own patch-token row and carry an independent noise level.
Other components, including the observation encoder, DiT trunk, register tokens, AdaLN observation conditioning, and output decoders, remain matched to UWM.
Thus, UWM + DF and UWM + DR share the modified architecture, differing from UWM mainly in noise conditioning, positional embeddings, and temporal patching.

\paragraph{Real-World Navigation.}
For real-world navigation, we adapt the policy-learning Chi-Transformer with a pretrained ResNet-50 visual encoder and robot-relative waypoint outputs.
DP, DF, and DR share this architecture and token-wise noise-conditioning
interface; DP assigns the same noise level to all waypoint tokens.
All use sinusoidal waypoint positional embeddings and full self-attention.
A shared inverse-dynamics MLP converts the first eight of 16 predicted waypoints into linear and angular velocity commands.

\section{Experiment Settings}
\label{sec:ExperimentSettings}

\subsection{Relative gain calculation}
The percentage improvements reported in the abstract are relative gains in
average success rate, not absolute percentage-point gains. For a baseline
average success rate $S_{\mathrm{base}}$ and a ReRoll average success rate
$S_{\mathrm{DR}}$, we compute
\[
\mathrm{Gain} =
\frac{S_{\mathrm{DR}} - S_{\mathrm{base}}}{S_{\mathrm{base}}}
\times 100.
\]
In guidance-based planning, Diffusion Forcing achieves an average success
rate of $72.78\%$, while Diffusion ReRoll achieves $88.07\%$, giving
\[
\frac{88.07 - 72.78}{72.78} \times 100 = 21.0\%.
\]
In goal-inpainting planning, Diffuser achieves $73.78\%$, while Diffusion
ReRoll achieves $90.76\%$, giving
\[
\frac{90.76 - 73.78}{73.78} \times 100 = 23.0\%.
\]
For LIBERO-10 under the official evaluation protocol, Diffusion Policy
achieves an average success rate of $40.06\%$, while Diffusion ReRoll
achieves $62.70\%$, giving
\[
\frac{62.70 - 40.07}{40.07} \times 100 = 56.5\%.
\]
Thus, the reported gains summarize relative improvements over the
corresponding matched baselines.

\subsection{Long-Horizon Planning}
\label{subsec:Maze}

\subsubsection{Evaluation Settings}
\label{subsubsec:MazeEval}

We evaluate long-horizon planning on the OGBench PointMaze and AntMaze navigation datasets. 
The dataset horizon is 500 for Medium and Large mazes and 1000 for Giant mazes. At Diffusion Forcing and Diffusion ReRoll training, we have used 5 frame stack per token for Large and Medium maze and 10 frame stack per token for Giant maze. All tasks use a 100-token prediction horizon.
We mainly follow the MCTD evaluation protocol~\cite{mctd}, but re-enable the default OGBench start-goal randomization that is partially removed in MCTD.
We additionally evaluated full-sequence diffusion using the Transformer
backbone shared by DF and DR.
Under guidance-based replanning, this variant achieved 4.0\% success
on AntMaze Giant and 34.4\% on AntMaze Large, below the corresponding
Diffuser U-Net results.
We therefore retain Diffuser's standard U-Net in the main planning
comparisons, while DF and DR share the same Transformer architecture.
These exploratory results concern guidance-based replanning, not
goal inpainting.
We compare Diffuser, Diffusion Forcing, and Diffusion ReRoll under two planning protocols: \emph{guidance-based replanning} and \emph{goal inpainting}. 
For all methods, the guidance and inpainting results use separately trained models. 
We train three seeds per configuration, select the best-performing checkpoint for each seed, and report the mean and standard deviation across the three training seeds.

\paragraph{Guidance-based replanning.}
In the guidance-based protocol, trajectories are sampled with test-time guidance and replanning during execution~\cite{classiferGuidance,guidance2}. 
For both maze types, we use distance-based guidance that minimizes the distance between planned positions and the goal:
\begin{equation}
    \mathcal{G}(x_{1:H}, g)
    =
    \sum_{i=1}^{H}
    \|x_i - g\|_2^2 .
\end{equation}
Although Diffuser is often evaluated with goal inpainting, we use the same distance-based guidance protocol for all methods. 
For Diffuser guidance training, only the start state is inpainted, while the start and goal state are inpainted for inpainting training. 
During execution, we use receding-horizon replanning every 50 environment steps. 
The maximum episode length is 1000 steps for Medium/Large mazes and 1500 steps for Giant mazes.

\paragraph{Goal inpainting.}
In the inpainting protocol, the start and goal states are fixed in the generated trajectory, and the model predicts the intermediate plan. 
Unlike guidance-based replanning, the model generates one full-horizon plan at the beginning of the episode and does not replan during execution. 
The plan horizon is set to the dataset horizon: 500 for Medium/Large and 1000 for Giant. 
Because the final inpainted token is the goal state, we restrict the execution budget to avoid giving the low-level controller unlimited extra time to solve the task by tracking the goal alone. 
For the same reason, after the planned trajectory is consumed, the controller tracks the penultimate predicted waypoint rather than the final goal token.

For controlled comparison, Diffusion ReRoll uses a bidirectional schedule matrix in the goal-inpainting setting. 
We also train an inpainting-compatible Diffusion Forcing variant using the same masking scheme and similar bidirectional schedule matrix as DR, but without ReRoll events.

\paragraph{Low-level execution.}
For PointMaze, we use the heuristic low-level controller from Diffuser~\cite{diffuser}. 
Given current position $x_t$, velocity $v_t$, planned position $p_t$, and planned velocity $\dot{p}_t$, the executed action is
\begin{equation}
    a_t
    =
    12.5 (p_t - x_t)
    +
    1.2 (\dot{p}_t - v_t),
\end{equation}
clipped to the valid action range. 
Because OGBench PointMaze does not provide velocity in the original observation, we augment the state with velocity to apply this controller.

For AntMaze, we use the learned low-level controller provided by the MCTD codebase~\cite{chen2024plandq,mctd}. 
The high-level diffusion planner outputs desired positions. 
During execution, the low-level policy tracks a subgoal 10 planned steps ahead. 
When the ant comes within distance 1.0 in the raw AntMaze $(x,y)$ coordinate space of the current subgoal, the subgoal is advanced by another 10 planned steps.
If the planned trajectory is exhausted, the controller tracks the
penultimate predicted waypoint for goal inpainting and the final
planned waypoint otherwise.

\subsubsection{ReRoll Parameters for Long-Horizon Planning}
\label{subsubsec:MazeReRollParams}

\begin{figure}[h!]
    \centering
    \includegraphics[width=0.99\linewidth]{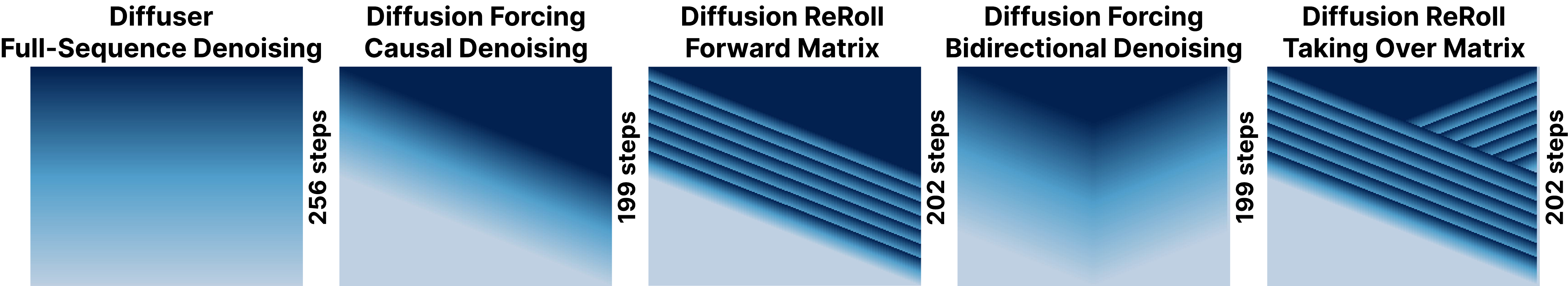}
    \caption{
    Used schedule matrices at OGBench planning tasks.
    }
    \label{fig:maze_matrix}
\end{figure}

For maze planning, we use a prediction horizon of $T=100$ for the ReRoll schedule matrix. 
The deployment matrix uses denoising slope $s_{\mathrm{dns}}=4$ and $6$ ReRoll events. The schedule matrices are shown in Fig.~\ref{fig:maze_matrix}.
This setting creates denoising waves that move across the horizon with moderate temporal span, while allowing repeated re-noising over most of the planned trajectory. Specific parameters are shown in Table~\ref{tab:maze_reroll_params}.



\begin{table}[h!]
\centering
\tiny
\setlength{\tabcolsep}{3.5pt}
\renewcommand{\arraystretch}{1.02}
\resizebox{0.55\linewidth}{!}{
\begin{tabular}{lc}
\toprule
Parameter & Value \\
\midrule
Prediction horizon $T$ & 100 \\
Denoising slope $s_{\mathrm{dns}}$ & 4 \\
ReRoll events $N_{\mathrm{roll}}$ & 6 \\
Reset noise level $k_{\mathrm{reset}}$ & $0.52K$ \\
Training chunk count $N_{\mathrm{chunk}}$ & $1$--$8$ \\
Chunk slope range & $[2,10]$ \\
Starting-level range & $[0,0.5K]$ \\
Forward branch probability $p_{\mathrm{forward}}$ & 0.475 \\
Bidirectional branch probability $p_{\mathrm{bi}}$ & 0.475 \\
Full-horizon branch probability $p_{\mathrm{full}}$ & 0.05 \\
Inpainting-mask probability $p_{\mathrm{inpaint}}$ & 0.1 \\
\bottomrule
\end{tabular}
}
\caption{
ReRoll hyperparameters for maze planning.
}
\label{tab:maze_reroll_params}
\end{table}

\subsubsection{Training and Sampling Hyperparameters}
\label{subsubsec:MazeTrain}

Table~\ref{tab:maze_train_params} summarizes the main training and sampling hyperparameters for the maze-planning experiments. 
Diffuser follows the full-sequence diffusion setup, while Diffusion Forcing and Diffusion ReRoll share the same transformer backbone and differ only in the training noise distribution and deployment schedule matrix. 
For each training seed, we evaluate 7 checkpoints and select the best-performing checkpoint.

\begin{table}[h!]
\centering
\scriptsize
\setlength{\tabcolsep}{4.0pt}
\renewcommand{\arraystretch}{1.06}
\resizebox{0.88\linewidth}{!}{
\begin{tabular}{lcc}
\toprule
Hyperparameter & Diffuser & DF / DR \\
\midrule
Optimizer & AdamW & AdamW \\
$(\beta_1,\beta_2)$ & $(0.9,0.999)$ & $(0.9,0.999)$ \\
Learning rate & $2\times10^{-4}$ & $5\times10^{-4}$ \\
Weight decay & $10^{-4}$ & $10^{-4}$ \\
Warmup steps & $10{,}000$ & $10{,}000$ \\
LR schedule after warmup & constant & constant \\
Batch size & 32 & 2048 \\
Max training steps & $1.2$M & $130{,}005$ \\
Evaluated checkpoints & $500$k--$1.1$M, every $100$k & $70$k--$130$k, every $10$k \\
EMA & decay $0.995$ & not used \\
EMA start / update & $2000$ / every $10$ steps & -- \\
Training diffusion steps & 256 & 1000 \\
Noise clip & 20.0 & 20.0 \\
Beta schedule & cosine & linear \\
Prediction objective & $x_0$ prediction & $x_0$ prediction \\
Loss weighting & uniform elementwise MSE (no SNR weighting) & min-SNR, clip $2.5$ \\
Guidance scale & 0.1 & 2.0 \\
Training window & task horizon & 100 tokens \\
Backbone & Temporal U-Net & Transformer \\
Backbone size & channels $[32,128,256]$, kernel 5 & width 128, 12 layers, 4 heads, FFN 512 \\
\bottomrule
\end{tabular}
}
\caption{
Training and sampling hyperparameters for OGBench PointMaze and AntMaze planning.
}
\label{tab:maze_train_params}
\end{table}

\subsection{Policy Learning}
\label{subsec:DP}

\subsubsection{Evaluation Environment Settings}
\label{subsubsec:EvaluationDP}
We evaluate diffusion-policy-style action prediction on LIBERO-10 multi-task learning and RoboCasa single-task learning. 
LIBERO-10 contains ten long-horizon manipulation tasks using a shared 7-DoF Franka end-effector action space.
Among various RoboCasa benchmark tasks, we have chosen four single-task kitchen manipulation settings: CloseSingleDoor, CoffeePressButton, CoffeeServeMug, and PnPCounterToSink. 
RoboCasa actions are 12-dimensional, including arm motion, gripper command, base motion, and control mode. However, for the selected tasks, only seven action dimensions are used, so we train with the same 7-D action format as LIBERO-10.

\paragraph{Observations.}
LIBERO-10 uses two RGB cameras together with proprioceptive state consisting of end-effector position, end-effector quaternion, and gripper state. 
We additionally provide a scalar \texttt{task\_uid} as the only task identifier. No language instruction or pretrained language encoder is used. 
RoboCasa uses three cameras with the same proprioceptive channels except for the task identifier.

\paragraph{Encoders and data splits.}
For LIBERO-10, all policies are trained from scratch. 
The image streams are encoded by a Robomimic-style convolutional visual encoder with random crop augmentation~\cite{imageRandom}, and no vision encoder or language pretraining is used. 
Each LIBERO task provides 50 demonstrations. For each training seed, we split them into 45 training and 5 validation trajectories per task, yielding 450 training trajectories in total. 
The split is regenerated across the three training seeds.

For RoboCasa, fully from-scratch visual encoders did not reliably converge at the demonstration counts used in our experiments. 
We therefore initialize the per-camera image encoder with the standard ImageNet-pretrained ResNet-18. 
This is the only pretrained component in the RoboCasa setup. 
CloseSingleDoor, CoffeePressButton, and CoffeeServeMug use 50 demonstrations each for training, 4 demonstrations for validation. PnPCounterToSink uses 300 demonstrations for training, including 50 human demonstrations and 250 MimicGen-generated demonstrations, with 24 demonstrations used for validation~\cite{mimicgen}. 

Both benchmarks process $128{\times}128$ camera inputs through the Robomimic image encoder, but use different random-crop sizes consistent with their encoder initialization. LIBERO-10 uses an aggressive $76{\times}76$ random crop with from-scratch weights, whereas RoboCasa uses a milder $116{\times}116$ random crop together with an ImageNet-pretrained ResNet-18 backbone ~\cite{imageRandom}. The crop size and pretraining choice are therefore coupled and are held fixed across all three compared methods within each benchmark.

\paragraph{Checkpoint selection}
\label{subsubsec:PolicyCheckpointSelection}

For policy-learning experiments, we select the best checkpoint for each training seed using test rollouts. 
After selecting the best checkpoint for each seed and protocol, we report test performance as the mean and standard deviation across three training seeds.

For LIBERO-10, we evaluate late-stage checkpoints every 10 epochs. 
All three methods are evaluated at 25 checkpoints from 210 epochs to 450. 
Each candidate checkpoint is evaluated with 50 initializations per task, and the checkpoint with the highest success rate is selected.

For RoboCasa, checkpoint selection is performed similarly. 
For 50-demonstration single-task experiments, we evaluate 31 candidate checkpoints from epochs 500 to 2000 at intervals of 50 epochs. 
For the 300-demonstration Pick-and-Place Counter-to-Sink setting, we evaluate 15 candidate checkpoints from epochs 220 to 500 at intervals of 20 epochs. 
Each candidate checkpoint is evaluated with 50 rollouts.


\subsubsection{LIBERO-10 Reset Protocol}
\label{subsubsec:liberoIssue}
In the main experiments, we report LIBERO-10 results under two reset protocols: the official initialization and an open-gripper initialization. 
This distinction is important because the initial gripper state has a large effect on evaluation difficulty. 
In the training demonstrations, the robot usually starts with an opened gripper. 
However, under the official evaluation reset, the robot waits after object spawning without actively opening the gripper, which produces an initial gripper state farther from the training distribution. 
The open-gripper protocol instead opens the gripper during this waiting period, making the evaluation state closer to the demonstrations.

\begin{figure}[h!]
    \centering
    \includegraphics[width=0.42\linewidth]{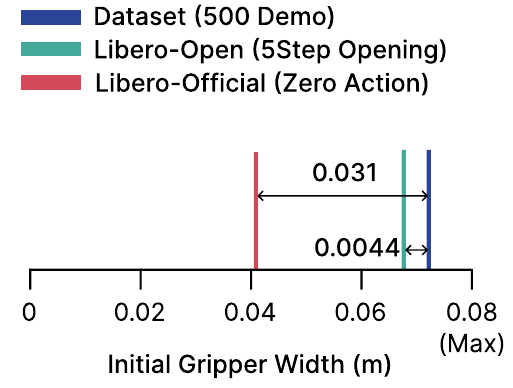}
    \caption{
    Initial gripper-state gap in LIBERO-10.
    The open-gripper reset is closer to the training distribution, while the official reset produces a more closed initial gripper state.
    }
    \label{fig:libero_gripper_gap}
\end{figure}

Fig.~\ref{fig:libero_gripper_gap} shows this distribution gap. 
The average initial gripper value is $0.07246\,\mathrm{m}$ in the training data and $0.06811\,\mathrm{m}$ under the open-gripper reset, but only $0.04135\,\mathrm{m}$ under the official reset. 
Thus, the official reset creates a stronger train-test mismatch in the initial robot state.

This protocol difference raises two concerns. 
First, reported LIBERO-10 results are difficult to compare unless the reset behavior is specified. 
LIBERO-10 is widely used in VLA and manipulation-policy benchmarks, but some comparisons reuse success rates from prior papers rather than re-running all methods under a unified evaluation script. 
In inspecting public codebases, we found that some implementations use different gripper-reset behavior during object spawning, which can unintentionally mix evaluation protocols across reported results. 
We do not claim that all discrepancies in prior results come from this issue, since we did not re-evaluate prior works under both protocols. 
However, our results show that this small implementation detail can noticeably affect success rates, so LIBERO-10 evaluations should explicitly report the initialization procedure.

Second, the open-gripper reset can make LIBERO-10 substantially easier. 
For example, in the $2/16/8$ setting, the standard Diffusion Policy baseline reaches $76.2\%$ success under the open-gripper reset but only $28.1\%$ under the official reset. This gap suggests that a major part of the task difficulty comes from recovering from an initial gripper state that is out of distribution relative to the demonstrations.
Therefore, we report both protocols in the main paper, the official reset for benchmark fidelity, and the open-gripper reset to show performance under an initialization closer to the training distribution.

\subsubsection{Applied ReRoll Parameters}
\label{subsubsec:PolicyReRollParams}

\begin{figure}[h!]
    \centering
    \includegraphics[width=0.99\linewidth]{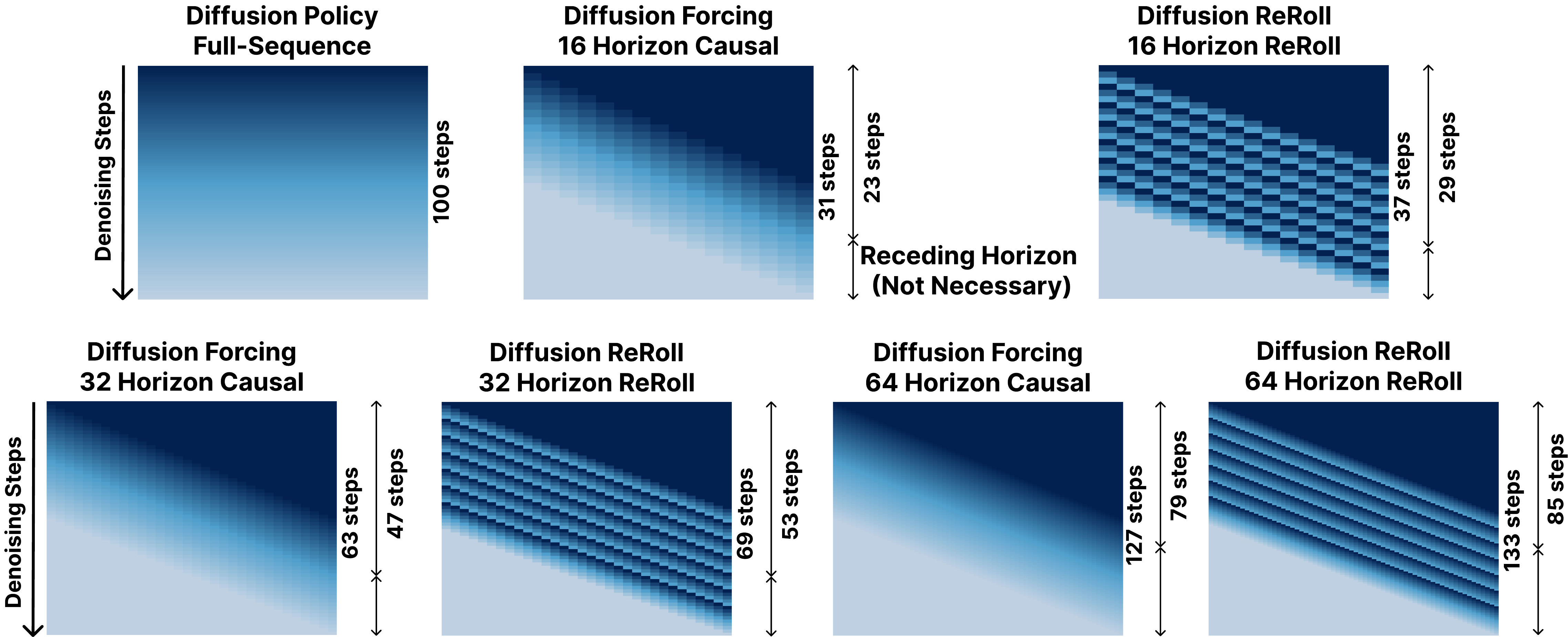}
    \caption{
    Schedule matrices used for diffusion-policy-style action prediction.
    Under receding-horizon execution, only the executable prefix of the predicted action chunk is applied before the next inference call.
    }
    \label{fig:policy_matrix}
\end{figure}

We use nearly shared ReRoll parameters across policy-learning experiments to avoid per-task schedule search. For the main evaluation, 100 Denoising Diffusion Probabilistic Model (DDPM) steps are used for basic diffusion policy. 
Since policies are executed in a receding-horizon manner, the entire predicted horizon does not need to be fully resolved at every inference call. 
The default parameters are summarized in Table~\ref{tab:policy_reroll_parameters}.

\begin{table}[h!]
\centering
\tiny
\setlength{\tabcolsep}{3.5pt}
\renewcommand{\arraystretch}{1.02}
\resizebox{0.48\linewidth}{!}{
\begin{tabular}{lc}
\toprule
Parameter & Value \\
\midrule
Prediction horizon $T$ & $16 / 32 / 64$ \\
Denoising slope $s_{\mathrm{dns}}$ & 4 \\
ReRoll events $N_{\mathrm{roll}}$ & 6 \\
Reset noise level $k_{\mathrm{reset}}$ & $0.5K$ \\
Training chunk count $N_{\mathrm{chunk}}$ & $1$--$8$ \\
Chunk slope range & $[2,10]$ \\
Starting-level range & $[0,0.5K]$ \\
\bottomrule
\end{tabular}
}
\caption{
ReRoll hyperparameters for policy learning.
}
\label{tab:policy_reroll_parameters}
\end{table}

\subsubsection{Basic Learning Hyperparameters}
\label{subsubsec:PolicyHparams}

All three policy-learning methods use the same Chi-Transformer backbone and the same optimizer settings. 
Table~\ref{tab:hparams_policy} summarizes the main hyperparameters.

\begin{table}[h!]
\centering
\scriptsize
\setlength{\tabcolsep}{4.0pt}
\renewcommand{\arraystretch}{1.06}
\resizebox{0.75\linewidth}{!}{
\begin{tabular}{lcc}
\toprule
Hyperparameter & Diffusion Policy & DF / DR \\
\midrule
Backbone & Chi-Transformer & Chi-Transformer \\
Action positional embedding & learnable & sinusoidal \\
Noise conditioning & global timestep token & per-token noise embedding \\
Causal attention mask & used & not used \\
Dropout & 0.1 & 0.01 \\
\midrule
Optimizer & AdamW & AdamW \\
$(\beta_1,\beta_2)$ & $(0.9,0.95)$ & $(0.9,0.95)$ \\
Policy learning rate & $5\times10^{-5}$ & $5\times10^{-5}$ \\
Visual encoder learning rate & $1\times10^{-5}$ & $1\times10^{-5}$ \\
Weight decay & 0 & 0 \\
Warmup steps & 200 & 200 \\
LR schedule after warmup & constant & constant \\
Batch size & 256 & 256 \\
EMA & power EMA & power EMA \\
\midrule
Training diffusion steps & 100 & 100 \\
Inference steps & 100 DDPM & schedule-matrix sampler \\
Prediction target & $\epsilon$ & $x_0$ \\
Beta schedule & squared cosine & squared cosine \\
Loss weighting & none & min-SNR, clip $2.5$ \\
\bottomrule
\end{tabular}
}
\caption{
Training and inference hyperparameters for policy-learning experiments.
DF denotes Diffusion Forcing and DR denotes Diffusion ReRoll.
}
\label{tab:hparams_policy}
\end{table}

The basic learning hyperparameters, including optimizer settings, learning rates, and batch size are inherited from the OAT codebase~\cite{OAT}.
The final reported test success is averaged over three training seeds.

\subsection{Unified Video-Action Modeling}
\label{subsec:appendix-uwm}
\subsubsection{Evaluation Environment Settings}
\label{subsubsec:uwmeval}

\begin{wrapfigure}{r}{0.45\textwidth}
  \vspace{-2.0em}
  \centering
  \includegraphics[width=0.44\textwidth]{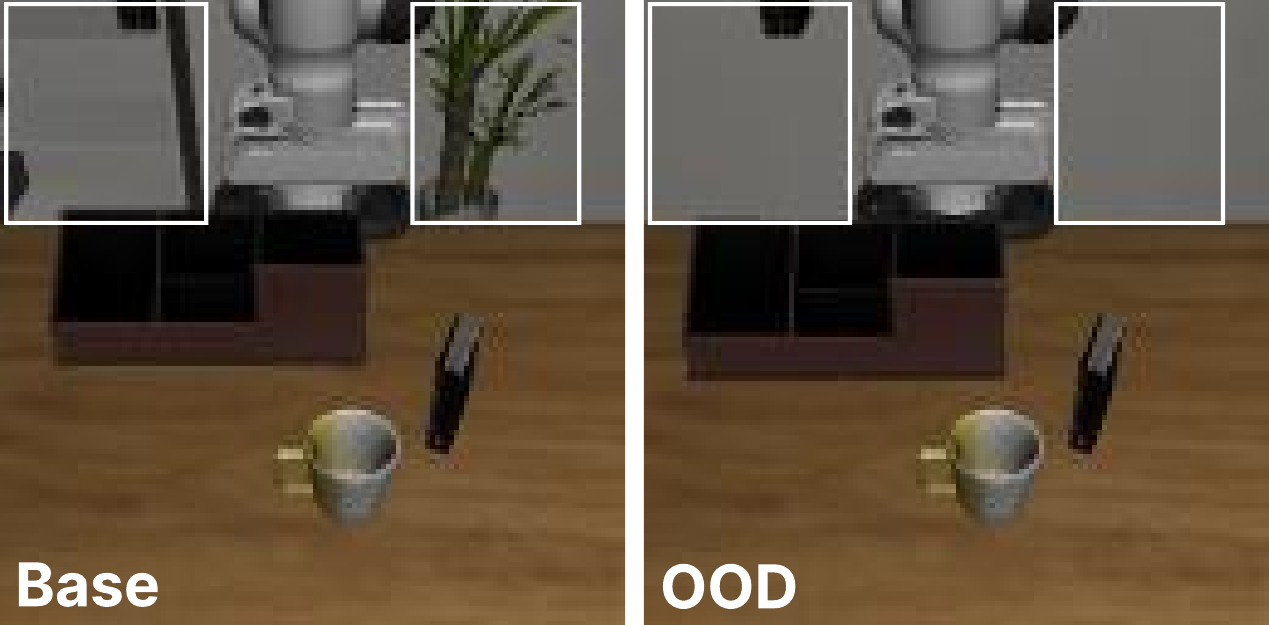}
    \caption{
    Observation differences between the base and OOD settings. Task-irrelevant background objects are removed in the OOD setting.
    }
  \vspace{-1.0em}
    
       \label{fig:OODImage}
\end{wrapfigure}
We evaluate UWM, UWM + DF, and UWM + DR on LIBERO-10.
For each method, a fixed LIBERO-90 checkpoint initializes three
fine-tuning runs per task.
Reported means and standard deviations reflect fine-tuning-seed
variation, not pretraining variability.
For OOD evaluation, we enlarge the object initialization range by
$0.03\,\mathrm{m}$ and hide task-irrelevant background objects.

\paragraph{Joint horizon and action-execution protocol.}
The original UWM uses a $16$-action chunk with $2$ future observation latents. 
We instead use $32$ action tokens with the same $2$ future image latents for all three variants, giving the joint sequence layout
\begin{equation}
    T_{\mathrm{joint}}
    =
    \underbrace{T_a}_{32 \text{ action tokens}}
    +
    \underbrace{T_v}_{2 \text{ future image latents}}
    =
    34 .
\end{equation}
For policy evaluation, every predicted action is executed before the next inference call, corresponding to the $2/32/32$ configuration: 2 observation frames, 32 predicted actions, and 32 executed actions. 
We evaluate two policy-generation modes. 
In \emph{action-only generation}, only the action tokens are generated, while the future image tokens are kept at full noise. 
In \emph{joint generation}, action tokens and future image tokens are denoised together under the joint schedule matrix. 
After environment reset, the policy is queried immediately from the post-spawn observation without additional gripper-opening or object-settling actions, roughly consistent with the official LIBERO-style reset described in Appendix~\ref{subsubsec:liberoIssue}.

\paragraph{Inverse dynamics setting.}
Evaluation starts from each demonstration's initial state.
At each prediction, the model receives current simulator observations
and fixed future demonstration images, whose latent tokens are
inpainted as clean constraints.
The target image pairs advance through zero-based indices
$(32,33),(64,65),(96,97),\ldots$, with 32 actions executed per prediction.

Evaluation stops when the next target pair is unavailable, or earlier
on success or termination.
For a demonstration containing $N$ frames, the available action budget is
\begin{equation}
    B_{\mathrm{inv}}(N)
    =
    32\left\lfloor\frac{N-2}{32}\right\rfloor .
\end{equation}
These budgets range from 128 to 512 steps across the evaluated
demonstrations, while policy evaluation permits up to 1,000 steps.

\paragraph{Asymmetric min-SNR clipping.}
We use min-SNR loss weighting~\cite{minSNR} for both action and image losses, but with different clipping thresholds. 
The action loss uses $\gamma_{\mathrm{act}}=2.5$, while the image-latent loss uses $\gamma_{\mathrm{img}}=5.0$. 
This gives the image branch a broader effective SNR range during training.

\paragraph{Image-loss weight.}
We also increase the image dynamics loss weight:
\begin{equation}
    \mathcal{L}
    =
    \mathcal{L}_{\mathrm{action}}
    +
    w_{\mathrm{img}}\mathcal{L}_{\mathrm{image}},
    \qquad
    w_{\mathrm{img}}=1.5 .
\end{equation}
This adjustment improves joint video-action generation by preventing the image-latent objective from being underweighted relative to the action objective.  
All other diffusion-side settings, including the prediction target, beta schedule, sample clipping, and train/inference step counts, are kept the same for the two streams.

\subsubsection{Applied ReRoll Parameters}
\label{subsubsec:uwm-rerollparam}
For UWM~+~DR, we use the bidirectional \emph{meeting} schedule matrix. 
The schedule is defined over the joint action-image horizon with $T_a=32$ action tokens and $T_v=2$ future image-latent tokens, giving $T_{\mathrm{joint}}=34$. 
Nearly the same schedule parameters are used for marginal action-policy generation, joint action-image generation, and inverse dynamics. 
In marginal action-policy generation, the image-latent columns are kept at full noise, so only the 32 action tokens are used for control. 
In inverse dynamics, the future image-latent columns are fixed as clean inpainting constraints, so the free denoising variables are again the 32 action tokens.

\begin{figure}[h]
    \centering
    \includegraphics[width=0.75\linewidth]{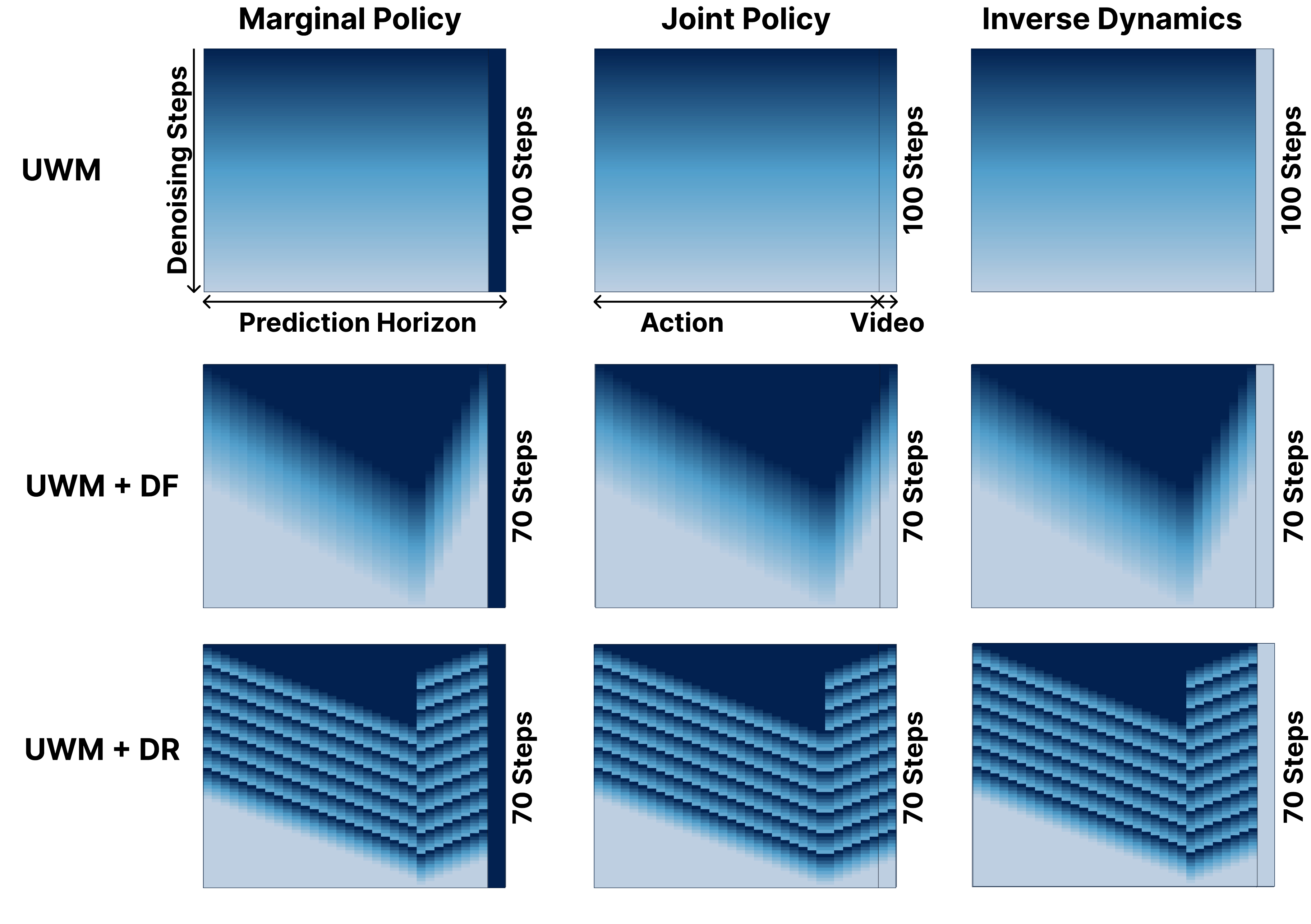}
  \vspace{-0.4em}

    \caption{
    Unified Video-Action generation schedule matrices.
    The schedule is defined over 32 action tokens and 2 future image-latent tokens.
    For inverse dynamics, the future image columns are clamped as clean inpainting constraints.
    }
    \label{fig:uwm-meet-matrices}
\end{figure}

  \vspace{-0.4em}

\begin{table}[h!]
\centering
\tiny
\setlength{\tabcolsep}{3.5pt}
\renewcommand{\arraystretch}{1.02}
\resizebox{0.58\linewidth}{!}{
\begin{tabular}{lc}
\toprule
Parameter & Value \\
\midrule
Joint horizon $T_{\mathrm{joint}}$ & 34 \\
Action horizon $T_a$ & 32 \\
Meeting point & $0.765T_{\mathrm{joint}}$ / $0.75T_a$ \\
Denoising slope $s_{\mathrm{dns}}$ & $3.78(T_{\mathrm{joint}})$ / $3.56(T_{\mathrm{a}})$ \\
ReRoll events, forward $N_{\mathrm{roll}}$ & 6 \\
ReRoll events, backward $N_{\mathrm{roll}}$ & 9 \\
Reset noise level $k_{\mathrm{reset}}$ & $0.412K$ \\
Training chunk count $N_{\mathrm{chunk}}$ & $1$--$8$ \\
Chunk slope range & $[2,10]$ \\
Starting-level range & $[0,0.5K]$ \\
Bidirectional branch probability $p_{\mathrm{bi}}$ & 0.30 \\
Full-horizon branch probability $p_{\mathrm{full}}$ & 0.05 \\
Image-inpainting probability $p_{\mathrm{inpaint}}$ & 0.1 \\
\bottomrule
\end{tabular}
}
\caption{
ReRoll hyperparameters for unified video-action modeling.
For entries with two values, the first corresponds to the joint action-image schedule and the second to the action-only schedule.
}
\label{tab:uwm-reroll-params}
\end{table}

\subsubsection{Basic Learning Hyperparameters}
\label{subsubsec:uwm-basic-hp}

All three methods use the same DiT-style joint transformer backbone and optimizer settings whenever possible. 
UWM~+~DF and UWM~+~DR differ from UWM mainly in the per-token noise-conditioning interface. Parameters are shown at Table~\ref{tab:uwm-hyperparams}.

\begin{table}[h!]
\centering
\scriptsize
\setlength{\tabcolsep}{4.0pt}
\renewcommand{\arraystretch}{1.06}
\resizebox{0.7\linewidth}{!}{
\begin{tabular}{lcc}
\toprule
Hyperparameter & UWM & UWM + DF / DR \\
\midrule
Backbone & DiT-style joint transformer & DiT-style joint transformer \\
Embedding dimension & 768 & 768 \\
Depth / heads / MLP ratio & 12 / 12 / 4 & 12 / 12 / 4 \\
Register tokens & 24 & 24 \\
Image latent patch shape & $2\times4\times4$ & $1\times4\times4$ \\
Positional embedding & learnable & sinusoidal \\
Noise conditioning & global AdaLN & per-token input concat. \\
\midrule
Optimizer & AdamW & AdamW \\
$(\beta_1,\beta_2)$, $\epsilon$ & $(0.9,0.999)$, $10^{-8}$ & $(0.9,0.999)$, $10^{-8}$ \\
Learning rate, pretrain & $1\times10^{-4}$ & $1\times10^{-4}$ \\
Learning rate, finetune & $1\times10^{-4}$, cosine, 1k warmup & $1\times10^{-4}$, cosine, 1k warmup \\
Weight decay & $10^{-6}$ & $10^{-6}$ \\
Pretraining steps & $100{,}000$ & $100{,}000$ \\
Fine-tuning steps & $20{,}000$ & $20{,}000$ \\
Batch size, train / eval & 16 / 4 & 16 / 4 \\
Observation history & 2 & 2 \\
Visual backbone & ResNet & ResNet \\
Image resize / crop & $240{\times}240$ / $224{\times}224$ & $240{\times}240$ / $224{\times}224$ \\
\bottomrule
\end{tabular}
}
\caption{
Training and inference hyperparameters for unified video-action modeling.
}
\label{tab:uwm-hyperparams}
\end{table}

\clearpage

\subsection{Real-World Navigation}
\label{subsec:RealNavigation}

\subsubsection{Evaluation Environment Settings}

\begin{wrapfigure}{r}{0.7\textwidth}
  \vspace{-1.0em}
  \centering
  \includegraphics[width=0.69\textwidth]{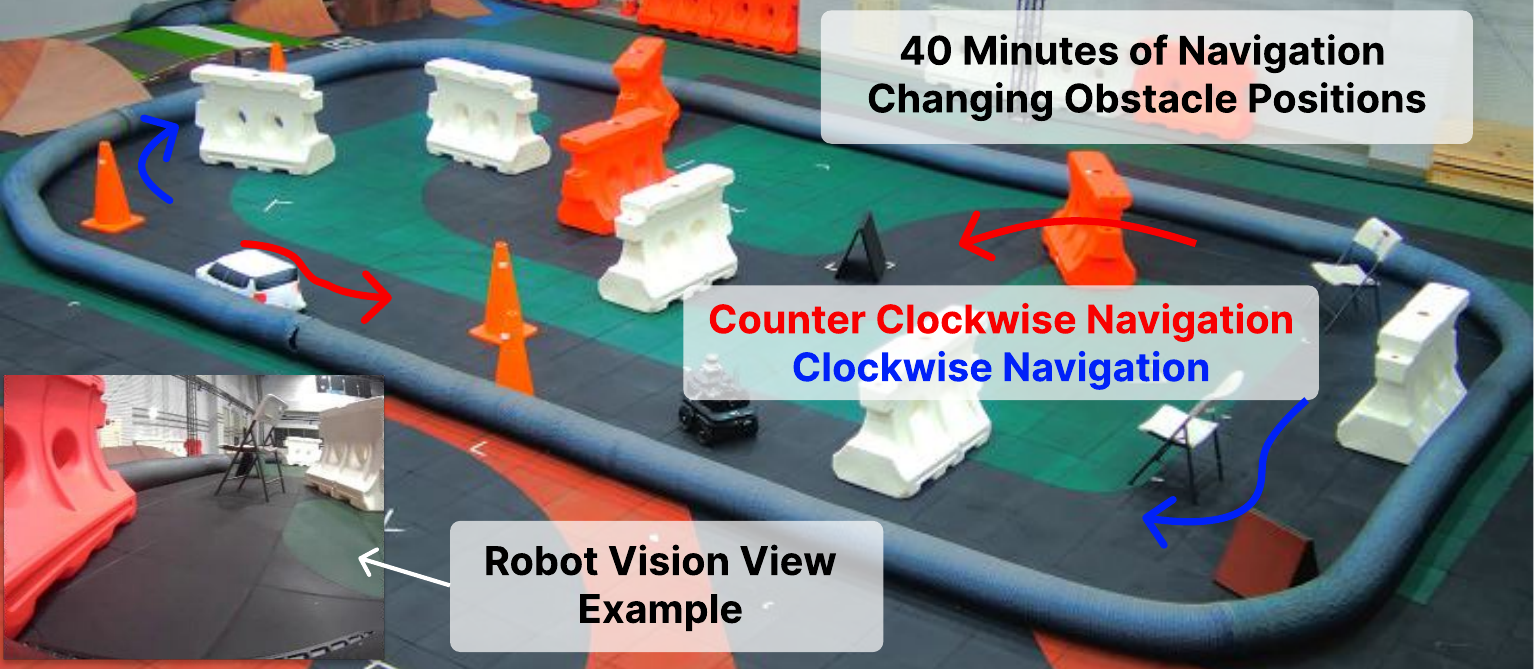}
    \caption{Real-world navigation demonstrations. Approximately 40 minutes
    of navigation cover clockwise and counterclockwise traversal with changing
    obstacle positions. The inset shows an example observation from the robot
    camera.}

  \vspace{-1.0em}
  \label{fig:real_nav_training_data}
\end{wrapfigure}

We evaluate DP, DF, and DR on a Clearpath Jackal mobile robot with onboard
inference using NVIDIA RTX3060. The robot navigates a physical
circuit consisting of black drivable regions, green restricted regions, and
obstacles. Successful navigation requires completing the designated circuit or
subtask while remaining within the black regions and avoiding collisions.
We evaluate full-circuit completion in both clockwise and counterclockwise
directions under OOD obstacle conditions, together with four
subtasks.

\paragraph{Evaluation protocol.}

We evaluate three independently trained policies per method. Each policy is
tested in three trials per circuit direction and five trials per subtask,
giving nine trials per direction and fifteen trials per subtask for each
method. We report the mean success rate across the three training seeds and
show each seed separately in the main-paper results.

\paragraph{Failure criteria.}
A collision failure occurs when contact prevents further motion or
displaces an obstacle by more than 3\,cm; minor contact without either
effect is allowed.
Leaving the drivable region is counted as failure when more than half of the
robot footprint lies outside the black region.
In practice, most boundary failures involved clear departures from the
drivable region rather than borderline cases.


\paragraph{Demonstrations.}

The demonstration data contain approximately 40 minutes of navigation in both
directions with changing obstacle positions, as shown in
Fig.~\ref{fig:real_nav_training_data}. The navigation policies use ten recorded
sequences, comprising five recordings in each direction. We assigned 90\% of data to training and
10\% to validation, using the same split for all methods and training seeds.
Future waypoint targets are
sampled at 5\,Hz.

\begin{figure}[htbp]
    \centering
    \includegraphics[width=0.8\linewidth]{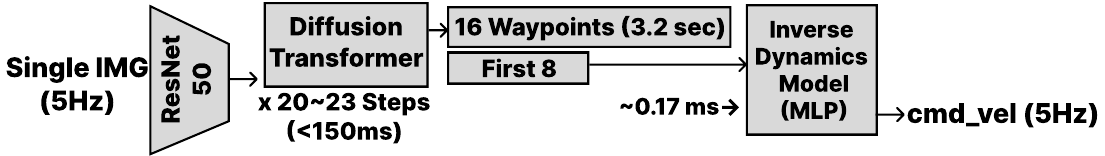}
    \caption{Onboard navigation pipeline. A single camera image conditions a
    diffusion model that predicts sixteen future waypoints.
    The first eight waypoints are mapped by an inverse-dynamics MLP to one
    linear and angular velocity command. The pipeline is configured to update
    at 5\,Hz.}
    \label{fig:real_nav_pipeline}
\end{figure}

\paragraph{Observations and control.}
At each control update, the policy receives a single RGB image resized to
$320\times240$ pixels. A pretrained ResNet-50 visual encoder conditions a
modified Chi-Transformer that predicts $T=16$ future waypoints spanning
3.2\,s. Each waypoint is represented in the current robot frame as
$(\Delta x,\Delta y,\cos\Delta\psi,\sin\Delta\psi)$, where $\Delta\psi$
is the relative yaw angle.
The first eight waypoints, corresponding to a 1.6-second planning prefix,
are passed to a learned inverse-dynamics MLP, which produces one linear and
angular velocity command. The pipeline is configured to update at 5\,Hz
using the latest image, as illustrated in Fig.~\ref{fig:real_nav_pipeline}.
Thus, the $1/16/8$ setting denotes one observation, sixteen predicted
waypoints, and eight waypoints supplied to the controller at each update.

\vspace{-0.4em}

\subsubsection{Applied ReRoll Parameters}

DR uses the forward ReRoll schedule with $N_{\mathrm{roll}}=3$ events to
reduce deployment latency. We retain the policy-learning setting's denoising
slope $s_{\mathrm{dns}}=4$ and reset trigger level
$k_{\mathrm{reset}}=0.5K$. The ReRoll parameters are
summarized in Table~\ref{tab:real_nav_reroll}.

\begin{figure}[htbp]
    \centering
    \includegraphics[width=\linewidth]{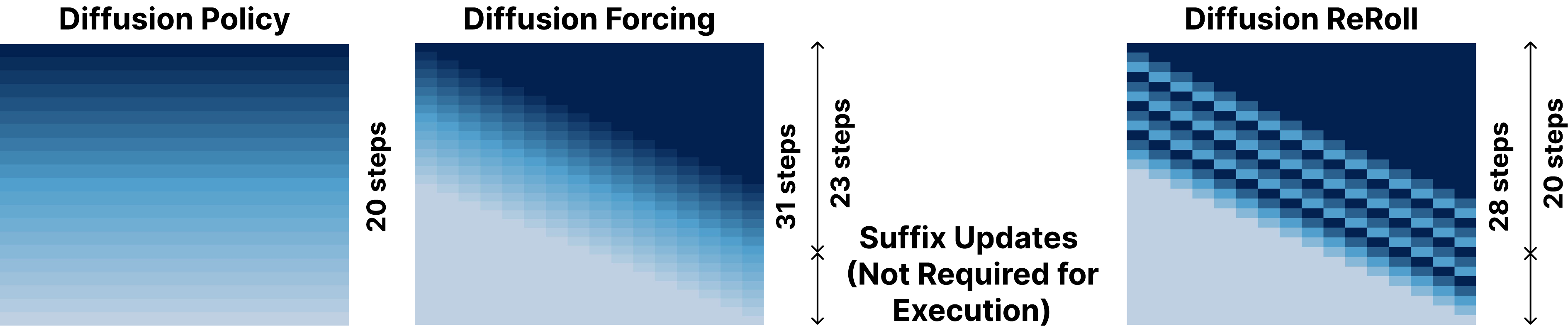}
    \caption{Denoising schedules for real-world navigation. DP, DF, and DR
    use 20, 23, and 20 deployed denoising updates, respectively. DF and DR stop
    once the first eight waypoints required by the inverse-dynamics controller
    are resolved; the remaining suffix need not be fully denoised.}
    \label{fig:real_nav_schedules}
\end{figure}

Only the first eight waypoints are required by the inverse-dynamics
controller. DF and DR therefore stop denoising once this prefix is clean,
while later waypoints may remain unresolved. DP uses 20 DDIM updates, DF uses
23 schedule-matrix updates, and DR uses 20 schedule-matrix updates per
prediction, as shown in Fig.~\ref{fig:real_nav_schedules}.

\begin{table}[htbp]
    \centering
    \small
    \begin{tabular}{@{}ll@{}}
        \toprule
        Parameter & Value \\
        \midrule
        Prediction horizon $T$ & 16 waypoints \\
        Prefix supplied to inverse dynamics & 8 waypoints \\
        Denoising slope $s_{\mathrm{dns}}$ & 4 \\
        ReRoll events $N_{\mathrm{roll}}$ & 3 \\
        Reset trigger level $k_{\mathrm{reset}}$ & $0.5K$ \\
        Training chunk count $N_{\mathrm{chunk}}$ & 1--8 \\
        Training chunk slope range & $[2,10]$ \\
        Noninitial chunk starting-level range & $[0,0.5K]$ \\
        Deployed updates, DP / DF / DR & 20 / 23 / 20 \\
        \bottomrule
    \end{tabular}
    \caption{ReRoll training and deployment parameters for real-world
    navigation.}
    \label{tab:real_nav_reroll}
\end{table}

\subsubsection{Basic Learning Hyperparameters}

Table~\ref{tab:real_nav_learning} summarizes the architecture and
optimization settings. All three navigation policies use sinusoidal
waypoint positional embeddings, per-token noise conditioning, and full
self-attention. The visual encoder is jointly fine-tuned with the
trajectory model using the same learning rate.

\paragraph{Diffusion training.}
All methods use a squared-cosine noise schedule.
DP shares one timestep across waypoint tokens and uses uniform loss
weighting. DF samples noise levels independently per token, while DR
uses the linear-chunk distribution in Table~\ref{tab:real_nav_reroll}.
DF and DR use min-SNR weighting with clipping threshold 2.5.

\paragraph{Data preprocessing.}
We apply brightness, contrast, and saturation jitter with strength 0.2
and ImageNet normalization. Waypoint positions are scaled to $[-1,1]$
using training-set minima and maxima, then restored to metric
coordinates for control.

\paragraph{Inverse dynamics.}
The same inverse-dynamics controller is used for all methods.
It receives the flattened eight-waypoint trajectory, uses ReLU hidden
activations, and has no current-velocity or yaw-rate inputs.
Training pairs demonstrated future trajectories with recorded velocity
commands, using ten demonstration sequences for fitting and two
separately recorded sequences for validation.

\begin{table}[h!]
\centering
\scriptsize
\setlength{\tabcolsep}{4.0pt}
\renewcommand{\arraystretch}{1.04}
\resizebox{0.83\linewidth}{!}{
\begin{tabular}{lcc}
\toprule
Hyperparameter & Navigation policies & Inverse dynamics \\
\midrule
Backbone & Modified Chi-Transformer & MLP \\
Visual encoder & Pretrained ResNet-50 & --- \\
Hidden dimension & 256 & 256 \\
Decoder / hidden layers & 8 & 3 \\
Attention heads & 4 & --- \\
Feed-forward dimension & 1024 & --- \\
Noise embedding dimension & 128 & --- \\
Dropout & 0.01 & 0.1 \\
\midrule
Optimizer & AdamW & AdamW \\
$(\beta_1,\beta_2)$ & $(0.9,0.999)$ & $(0.9,0.999)$ \\
Peak learning rate & $10^{-4}$ & $10^{-3}$ \\
Weight decay & $10^{-4}$ & $10^{-4}$ \\
Warmup & 5 epochs & None \\
LR schedule after warmup & Cosine & Cosine \\
Minimum learning rate & $10^{-6}$ & $10^{-6}$ \\
Batch size & 64 & 32 \\
Training epochs & 300 & 100 \\
Gradient clipping norm & 1.0 & 1.0 \\
Training diffusion steps & 100 & --- \\
Prediction target & Clean trajectory $x_0$ & Linear/angular velocity \\
Loss & MSE; method-specific weighting & Smooth L1, $\beta=0.5$ \\
Checkpoint selection & Full-horizon validation ADE & Validation loss \\
\bottomrule
\end{tabular}
}
\caption{
Architecture and learning hyperparameters for real-world navigation.
Navigation-policy settings are shared by DP, DF, and DR.
}
\label{tab:real_nav_learning}
\end{table}

\section{Additional Results}
\label{sec:ADDRE}

\subsection{Best-Checkpoint Results}
\label{subsec:best_checkpoint_results}

The main results report mean performance over three training seeds after selecting the best checkpoint for each seed. 
For completeness, we also report the best single-seed performance observed for each method and setting. 
These results are not used as the primary comparison, but show the maximum performance reached by each method under the evaluated checkpoint set. 
Table~\ref{tab:planning_results_max} reports best-checkpoint results for long-horizon planning, while Tables~\ref{tab:libero_policy_max} and~\ref{tab:robocasa_policy_max} report best-checkpoint results for LIBERO-10 and RoboCasa policy learning.

\begin{table}[h!]
\centering
\scriptsize
\setlength{\tabcolsep}{4.2pt}
\renewcommand{\arraystretch}{1.03}
\resizebox{0.87\linewidth}{!}{
\begin{tabular}{@{}llccc ccc@{}}
\toprule
\multirow{2}{*}{Mode}
& \multirow{2}{*}{Method}
& \multicolumn{3}{c}{AntMaze}
& \multicolumn{3}{c}{PointMaze} \\
\cmidrule(lr){3-5}
\cmidrule(lr){6-8}
& & Giant & Large & Medium
& Giant & Large & Medium \\
\midrule

\multirow{3}{*}{\shortstack[c]{Guidance\\Replan}}
& Diffuser
& 17.2
& 46.8
& 70.0
& 32.0
& 31.6
& 47.2 \\

& Forcing
& \underline{56.0}
& \underline{61.6}
& \underline{96.0}
& \underline{92.8}
& \underline{62.0}
& \underline{94.0} \\

& ReRoll
& \textbf{87.6}
& \textbf{71.2}
& \textbf{98.4}
& \textbf{100.0}
& \textbf{78.4}
& \textbf{99.2} \\

\midrule

\multirow{3}{*}{Inpainting}
& Diffuser
& 34.4
& \underline{72.8}
& 89.2
& 66.8
& 96.4
& \underline{95.2} \\

& Forcing
& \underline{75.6}
& 68.0
& \underline{91.2}
& \underline{96.4}
& \underline{97.6}
& 94.4 \\

& ReRoll
& \textbf{84.8}
& \textbf{74.8}
& \textbf{94.4}
& \textbf{99.2}
& \textbf{100.0}
& \textbf{99.2} \\

\bottomrule
\end{tabular}
}
\caption{
Maximum long-horizon planning success rates (\%) on OGBench PointMaze and AntMaze.
Values report the best observed success rate for each method and task.
Best results are bolded and second-best results are underlined.
}
\label{tab:planning_results_max}
\end{table}

\clearpage

\begin{table}[h!]
\centering

\scriptsize
\setlength{\tabcolsep}{2.9pt}
\renewcommand{\arraystretch}{1.08}
\resizebox{\linewidth}{!}{
\begin{tabular}{lccccc ccccc}
\toprule
\multirow{2}{*}{Method}
& \multicolumn{5}{c}{LIBERO-10 Official \;{\scriptsize (Obs./Hor./Act.)}}
& \multicolumn{5}{c}{LIBERO-10 Open \;{\scriptsize (Obs./Hor./Act.)}} \\
\cmidrule(lr){2-6}
\cmidrule(lr){7-11}
& 2/16/8 & 2/32/16 & 2/64/16 & 4/32/16 & 8/32/16
& 2/16/8 & 2/32/16 & 2/64/16 & 4/32/16 & 8/32/16 \\
\midrule

Diff. Policy
& 31.6
& 49.6
& 38.2
& 52.2
& 54.8
& \underline{76.6}
& \underline{67.6}
& 60.2
& 70.6
& 66.0 \\

Forcing
& \underline{47.2}
& \underline{50.8}
& \underline{46.0}
& \underline{65.4}
& \underline{65.0}
& 69.4
& 61.4
& \underline{62.0}
& \underline{71.6}
& \underline{77.8} \\

ReRoll
& \textbf{68.8}
& \textbf{59.4}
& \textbf{67.8}
& \textbf{76.6}
& \textbf{73.8}
& \textbf{85.0}
& \textbf{77.8}
& \textbf{71.8}
& \textbf{83.8}
& \textbf{83.8} \\

\bottomrule
\end{tabular}
}
\caption{
Maximum LIBERO-10 multi-task policy learning success rates under two initialization protocols.
Values report the best observed success rate (\%) for each method and setting.
Best results are bolded and second-best results are underlined.
}
\label{tab:libero_policy_max}
\end{table}

\begin{table}[h!]
\centering

\scriptsize
\setlength{\tabcolsep}{3.2pt}
\renewcommand{\arraystretch}{1.06}
\resizebox{\linewidth}{!}{
\begin{tabular}{lccc ccc ccc ccc}
\toprule
\multirow{2}{*}{Method}
& \multicolumn{3}{c}{CloseSingleDoor}
& \multicolumn{3}{c}{CoffeePressButton}
& \multicolumn{3}{c}{CoffeeServeMug}
& \multicolumn{3}{c}{PnPCounterToSink (300 Demos)} \\
\cmidrule(lr){2-4}
\cmidrule(lr){5-7}
\cmidrule(lr){8-10}
\cmidrule(lr){11-13}
& 2/32/16 & 2/64/16 & 4/32/16
& 2/32/16 & 2/64/16 & 4/32/16
& 2/32/16 & 2/64/16 & 4/32/16
& 2/32/16 & 2/64/16 & 4/32/16 \\
\midrule

Diff. Policy
& 54.0
& 54.0
& 66.0
& 82.0
& 64.0
& 84.0
& \textbf{46.0}
& 36.0
& \textbf{46.0}
& \textbf{34.0}
& 26.0
& 26.0 \\

Forcing
& \underline{66.0}
& \textbf{70.0}
& \underline{68.0}
& \underline{88.0}
& \underline{72.0}
& \textbf{94.0}
& 42.0
& \textbf{46.0}
& \underline{40.0}
& \textbf{34.0}
& \underline{30.0}
& \textbf{38.0} \\

ReRoll
& \textbf{74.0}
& \underline{66.0}
& \textbf{70.0}
& \textbf{90.0}
& \textbf{88.0}
& \underline{90.0}
& \textbf{46.0}
& \underline{40.0}
& 38.0
& 32.0
& \textbf{36.0}
& \underline{32.0} \\

\bottomrule
\end{tabular}
}
\caption{
Maximum RoboCasa single-task policy learning success rates (\%).
Values report the best observed success rate for each method and setting.
Best results are bolded and second-best results are underlined.
}
\label{tab:robocasa_policy_max}
\end{table}

\subsection{Details for the Zero-Action Intervention}
\label{subsec:zero_action_details}

We report zero-action error as a diagnostic of action-video consistency.
We fix future action tokens to zero and generate future image tokens
using forward dynamics. We then apply inverse dynamics to the generated
images and measure the terminal end-effector displacement implied by
the recovered actions. Lower displacement indicates stronger consistency
with the imposed zero-action condition.

Successful demonstration trajectories have an average terminal
end-effector displacement of $d_{\mathrm{demo}}=0.351\,\mathrm{m}$
over the same horizon. We also report the reduction relative to this
reference:
\begin{equation}
    \Delta_{\mathrm{demo}}
    =
    d_{\mathrm{demo}} - d,
\end{equation}
where $d$ is the mean zero-action displacement across three training
seeds. Larger $\Delta_{\mathrm{demo}}$ indicates less implied motion
than the successful-demonstration reference.

\begin{table}[h!]
\centering
\scriptsize
\setlength{\tabcolsep}{4.0pt}
\renewcommand{\arraystretch}{1.08}

\resizebox{\linewidth}{!}{
\begin{tabular}{lcccc}
\toprule
Evaluation setting
& Dataset ref.
& UWM
& UWM + DF
& UWM + DR \\
\midrule

Successful demonstrations
& \shortstack{
    $0.3510$ \\
    {\scriptsize $(\Delta=0.0000)$}
  }
& --
& --
& -- \\

Gen. zero-act video $\rightarrow$ Inv. Dyn.
& --
& \shortstack{
    $0.3283 \pm 0.0038$ \\
    {\scriptsize $(\Delta=0.0227)$}
  }
& \shortstack{
    $0.3170 \pm 0.0121$ \\
    {\scriptsize $(\Delta=0.0340)$}
  }
& \shortstack{
    $\mathbf{0.3006} \pm 0.0029$ \\
    {\scriptsize $(\Delta=0.0504)$}
  } \\

GT zero-act video $\rightarrow$ Inv. Dyn.
& --
& \shortstack{
    $0.3104 \pm 0.0006$ \\
    {\scriptsize $(\Delta=0.0406)$}
  }
& \shortstack{
    $0.3156 \pm 0.0142$ \\
    {\scriptsize $(\Delta=0.0354)$}
  }
& \shortstack{
    $\mathbf{0.2915} \pm 0.0040$ \\
    {\scriptsize $(\Delta=0.0595)$}
  } \\

\bottomrule
\end{tabular}
}
\caption{
Zero-action consistency diagnostics.
Model entries report terminal end-effector displacement in meters
(mean$\pm$std over three training seeds), with
$\Delta_{\mathrm{demo}}=0.351-d$ in parentheses.
Lower displacement and larger reduction are better.
The lowest mean in each diagnostic is bolded.
}
\label{tab:zero_action_details}
\end{table}

All models yield lower mean implied displacements than the
successful-demonstration reference. In the generated-video setting,
UWM~+~DR achieves the lowest displacement
($0.3006\,\mathrm{m}$) and the largest reduction
($0.0504\,\mathrm{m}$), indicating the strongest consistency
under this diagnostic.

The GT zero-action video setting isolates inverse-dynamics behavior
by providing images from true zero-action rollouts, eliminating
video-generation errors. Mean displacement decreases for all three
methods, although UWM~+~DF improves only slightly
($0.3170 \rightarrow 0.3156\,\mathrm{m}$).
UWM~+~DR again achieves the lowest displacement
($0.2915\,\mathrm{m}$).
Since UWM~+~DF and UWM~+~DR share the same architecture,
these results are consistent with improved alignment from DR's
structured training and revisable sampling, without isolating
their individual contributions.

\subsection{High Variance in LIBERO-10 Official 2/16/8}
\label{subsec:high_variance_short_horizon}

In Table 2 of the main paper, DR shows high variance in the LIBERO-10 Official $2/16/8$ setting. 
We do not interpret this as a general instability of ReRoll. 
As discussed in Appendix~\ref{subsubsec:liberoIssue}, the official reset creates an out-of-distribution initial gripper state relative to the demonstrations. 
The short-horizon $2/16/8$ setting appears especially sensitive to this mismatch. If the policy fails to recover from the initial gripper state, official success can drop sharply.

Table~\ref{tab:libero_2168_seed_open_official} shows the per-seed results under both the official and open-gripper reset protocols. 
For each seed, DP, DF, and DR use the same 10\% validation split.
Under the official reset, all three methods show at least one weak seed, and DR has one particularly low seed. 
Under the open-gripper reset, where the initial state is closer to the training distribution, all DR seeds perform strongly and consistently. 
These results are consistent with sensitivity to the reset condition
and training-seed variation, rather than by short-horizon ReRoll behavior alone. 
This effect may also be amplified because our LIBERO-10 policies are trained from scratch without pretrained vision encoders or text encoders, making recovery from the OOD initial gripper state more difficult.

\begin{table}[h!]
\centering

\scriptsize
\setlength{\tabcolsep}{5pt}
\renewcommand{\arraystretch}{1.05}
\resizebox{0.72\linewidth}{!}{
\begin{tabular}{lccc ccc}
\toprule
\multirow{2}{*}{Seed}
& \multicolumn{3}{c}{Official}
& \multicolumn{3}{c}{Open} \\
\cmidrule(lr){2-4}
\cmidrule(lr){5-7}
& DP & DF & DR
& DP & DF & DR \\
\midrule
17 & 28.6 & 24.0 & 23.6 & 76.4 & 66.4 & 78.4 \\
23 & 31.6 & 40.8 & 60.6 & 76.6 & 69.4 & 85.0 \\
42 & 24.0 & 47.2 & 68.8 & 75.6 & 69.2 & 84.2 \\
\midrule
Mean
& 28.1\sdev{3.8}
& 37.3\sdev{12.0}
& 51.0\sdev{24.1}
& 76.2\sdev{0.5}
& 68.3\sdev{1.7}
& 82.5\sdev{3.6} \\
\bottomrule
\end{tabular}
}
\caption{
Per-seed LIBERO-10 $2/16/8$ success rates under official and open-gripper resets.
The open-gripper reset reduces the initial-state mismatch and produces more stable performance.
}
\label{tab:libero_2168_seed_open_official}
\end{table}

\subsection{Execution-Length and Delay Ablations}
\label{subsec:execution_delay}

We examine execution length and delayed-action effects on LIBERO-10
Official, using \textsc{Obs./Hor./Act.}=2/32/16 as the reference setting.

\begin{wrapfigure}{r}{0.65\textwidth}
  \vspace{-1.0em}
  \centering
  \includegraphics[width=0.64\textwidth]{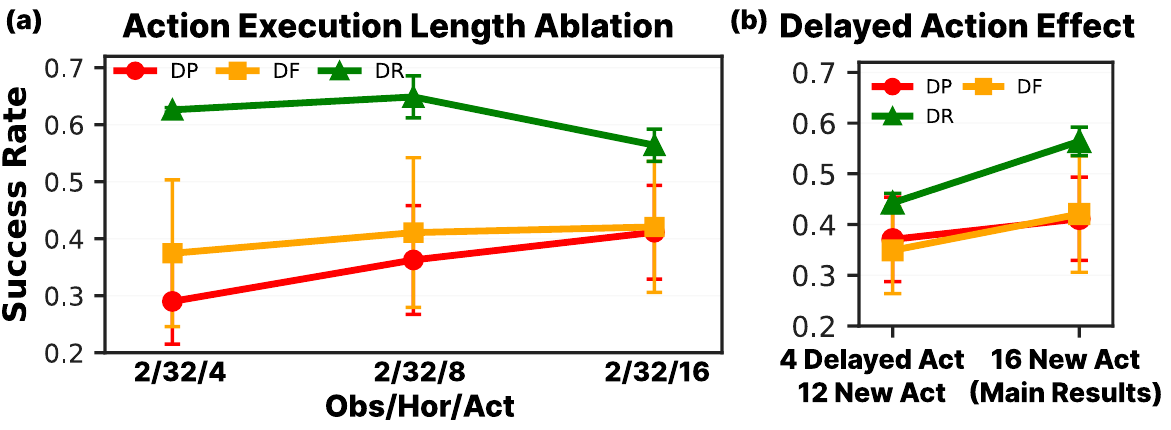}
  \caption{
  Execution-length and delay ablations on LIBERO-10 Official with
  two observation steps and a 32-step prediction horizon.
  (a) Executing 4, 8, or 16 actions before replanning.
  (b) Executing four retained actions followed by twelve newly predicted
  actions, compared with sixteen newly predicted actions.
  Points show means; error bars indicate $\pm$ one standard deviation
  over three training seeds.
  }
  \label{fig:Horizon_Action}
  \vspace{-1.0em}
\end{wrapfigure}

\paragraph{Execution length.}
We keep the observation history and prediction horizon fixed and vary
the executed prefix from 16 to 8 or 4 actions.
Fig.~\ref{fig:Horizon_Action}(a) shows that DR achieves the highest mean
success at all three execution lengths, with higher success at 4 and 8
than at 16.
Thus, DR's advantage persists even when only a short prefix is executed
before replanning, suggesting that ReRoll-refined future context benefits
earlier action predictions.

\paragraph{Delayed actions.}
At each replanning step, we acquire the current observation and inpaint
the first four action positions with values from the previous prediction.
Conditioned on this observation and fixed prefix, we predict the next
twelve executable actions within the 32-step horizon.
After inference, we execute the four retained actions followed by the
twelve new actions.
Thus, the new actions are predicted from observations acquired before
the retained prefix is executed.
The fixed actions can condition later predictions but cannot themselves
be revised using updated future context.
Fig.~\ref{fig:Horizon_Action}(b) shows that DR experiences the largest
performance drop, although it retains the highest mean success.

Together, these results are consistent with cross-horizon revision
benefiting near-future predictions, while revealing sensitivity to
previously committed actions.
This delayed-action setting approximates asynchronous deployment, where
the robot continues executing actions from the previous prediction while
a revised plan is computed from the latest observation.
Longer blocking latency therefore increases the number of stale actions
that must be executed before the revised plan can take effect, motivating
the latency reductions in Fig.~\ref{fig:inference_latency}.

\subsection{Long-Horizon Planning Visualization}
\label{subsec:planning_visualization}

Fig.~\ref{fig:Explore} shows qualitative denoising behavior on AntMaze Giant. 
We provide additional visualizations for Diffuser, Diffusion Forcing, and Diffusion ReRoll in Figs.~\ref{fig:diffuser_visualization}--\ref{fig:DR_visualization}. 
These examples illustrate different failure and recovery patterns: full-sequence diffusion can commit to early denoising errors, causal denoising can lock in incorrect prefixes, while Diffusion ReRoll keeps plans revisable through selective re-noising.

\subsection{Generated Video Visualization}
\label{VideoVisualization}
\begin{figure}[h!]
    \centering
    \includegraphics[width=0.99\linewidth]{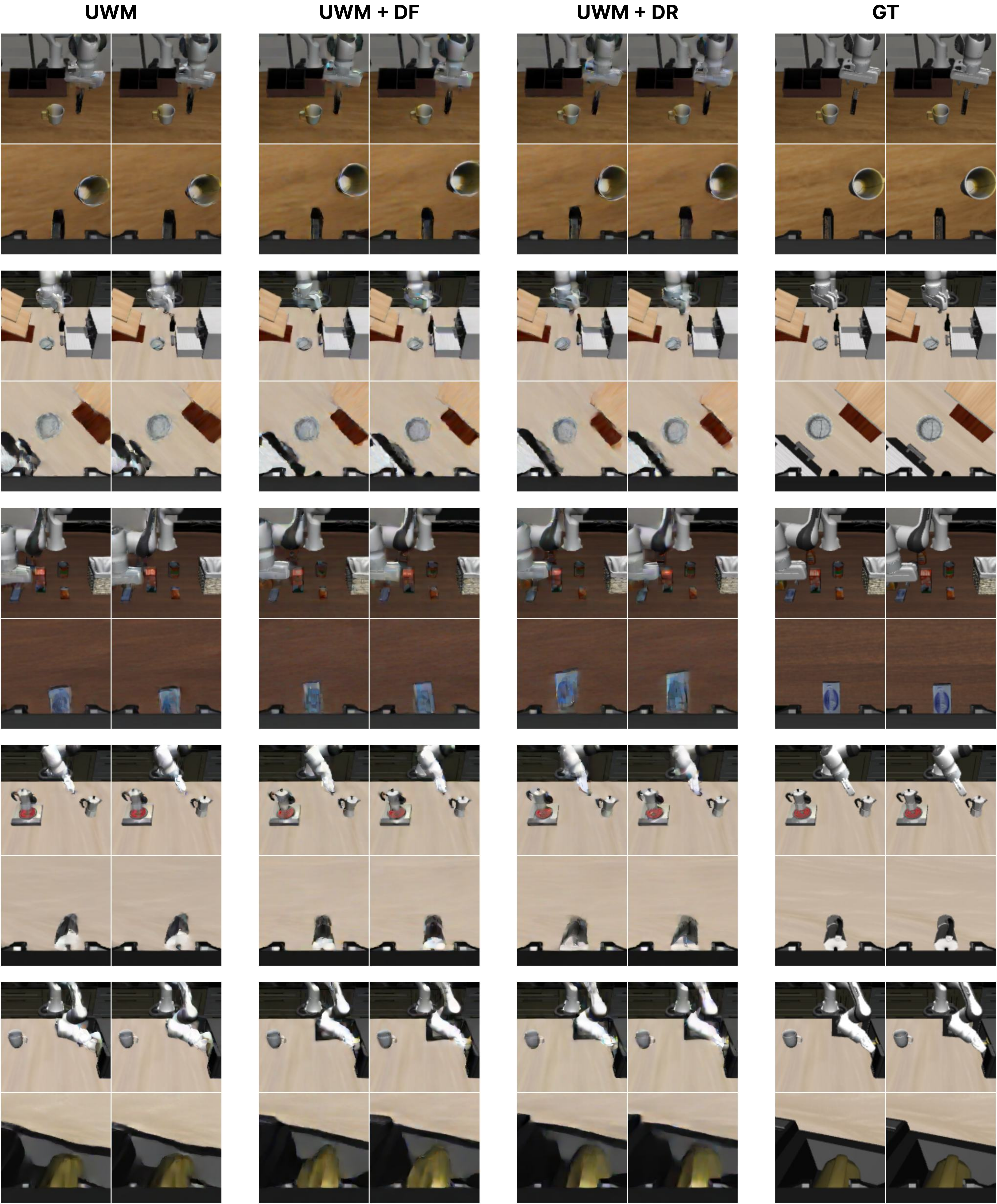}
    \caption{
    Generated future-video examples from UWM, UWM~+~DF, and UWM~+~DR.
    }
    \label{fig:Visualization1}
\end{figure}

\begin{figure}[h!]
    \centering
    \includegraphics[width=0.99\linewidth]{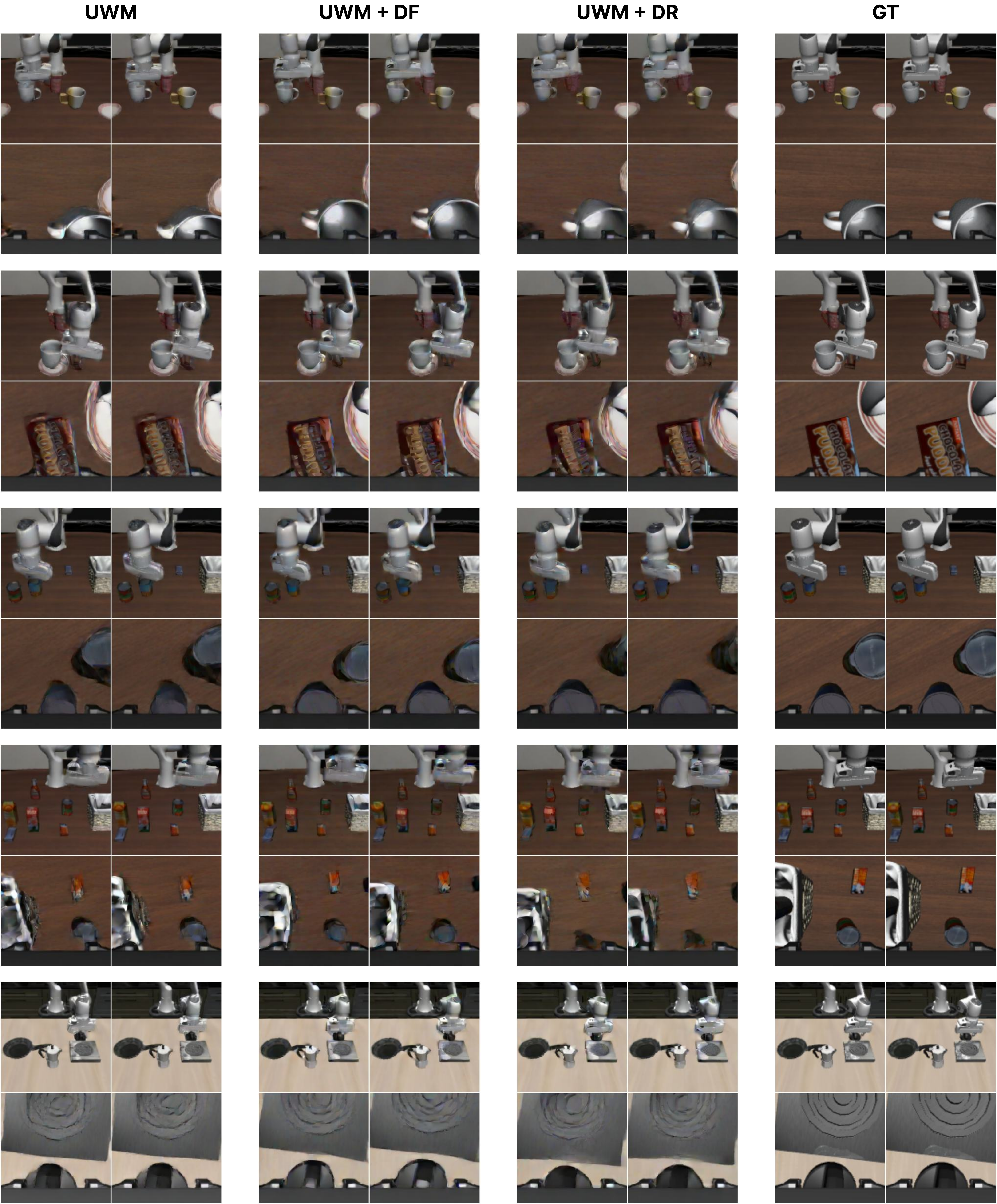}
    \caption{
    Additional generated future-video examples from UWM, UWM~+~DF, and UWM~+~DR.
    }
    \label{fig:Visualization2}
\end{figure}

We provide generated-video examples in Figs.~\ref{fig:Visualization1} and~\ref{fig:Visualization2}. UWM, UWM~+~DF, and UWM~+~DR produce comparable visual generation quality in these examples. The original UWM uses 100 DDIM steps, while UWM~+~DF and UWM~+~DR use 70 schedule-matrix updates. These results suggest that the downstream performance differences are not caused by a clear degradation in generated-video quality.

\begin{figure}[h!]
    \centering
    \includegraphics[width=0.99\linewidth]{images/Appendix/mazevisualization/Full3.pdf}
    \caption{
    Denoising visualization for Diffuser.
    We visualize predicted clean trajectories $\hat{x}_0$ at selected denoising steps.
    Full-sequence diffusion can commit to an early incorrect plan.
    }
    \label{fig:diffuser_visualization}
\end{figure}

\begin{figure}[h!]
    \centering
    \includegraphics[width=0.99\linewidth]{images/Appendix/mazevisualization/Forcing_Total_0916.pdf}
    \caption{
    Denoising visualization for Diffusion Forcing.
    We visualize predicted clean trajectories $\hat{x}_0$ at selected denoising steps.
    Causal denoising can resolve early prefixes before later horizon regions are sufficiently revised.
    }
    \label{fig:DF_visualization}
\end{figure}

\begin{figure}[h!]
    \centering
    \includegraphics[width=0.99\linewidth]{images/Appendix/mazevisualization/ReRoll3.pdf}
    \caption{
    Denoising visualization for Diffusion ReRoll.
    We visualize predicted clean trajectories $\hat{x}_0$ at selected denoising steps.
    ReRoll keeps trajectory regions revisable and can recover a valid long-horizon plan.
    }
    \label{fig:DR_visualization}
\end{figure}

\end{document}